\RequirePackage{fix-cm}
\documentclass[smallextended]{svjour3}       
\smartqed  
\usepackage{graphicx}
\usepackage{listings}
\usepackage{comment}
\usepackage{graphicx}
\usepackage{epsfig}
\usepackage{array, colortbl}
\newcolumntype{P}[1]{>{\raggedright\arraybackslash}p{#1}}
\usepackage{listings}
\usepackage{epstopdf}
\usepackage{multirow}
\usepackage{rotating}
\usepackage{subfigure}
\usepackage{float}
\usepackage{balance}
\usepackage{scalefnt}
\usepackage[normalem]{ulem}
\usepackage[hang]{footmisc}
\usepackage{booktabs}
\usepackage{pbox}
\usepackage{ragged2e}
\usepackage{fancybox}
\usepackage{guehene}
\usepackage{chngpage} 
\usepackage[numbers]{natbib}
\usepackage{soul}
\usepackage[hyphens]{url}
\usepackage{hyperref}
\usepackage[T1]{fontenc} 
\usepackage{tcolorbox}
\usepackage{amsmath}
\usepackage[english]{babel}
\usepackage[english=american]{csquotes}
\usepackage[edges]{forest}
\usepackage{tikz}
\usepackage{ltablex}
\usepackage{xltabular}
\usetikzlibrary{trees}
\usepackage{algorithm}
\usepackage{algpseudocode}
\usepackage{multicol}

\usepackage[toc,page]{appendix}
\tikzset{parent/.style={align=center,text width=2.5cm,rounded corners=2pt},
    child/.style={align=center,text width=2.5cm,rounded corners=2pt}
    }

\newcommand{\green}[1] {\textcolor{green!60!black}{#1} }
\definecolor{dkgreen}{rgb}{0,0.6,0}
\definecolor{gray}{rgb}{0.5,0.5,0.5}

\renewcommand{\eg}{\textit{e}.\textit{g}.}
\renewcommand{\ie}{\textit{i}.\textit{e}.}

\makeatletter

\renewcommand\paragraph{\@startsection{paragraph}{4}{\z@}%
                    {-2.5ex\@plus -1ex \@minus -.2ex}
                                    {0.7ex \@plus .2ex}%
                                    {\normalfont\normalsize\bfseries}}
\makeatother

\newcommand{\app}{$^\dag\,$}

\newcommand*{\affmark}[1][*]{\textsuperscript{#1}}

\DeclareMathOperator*{\argmax}{arg\,max}

\journalname{Automated Software Engineering}
\begin{document}
\sloppy

\title{How to Certify Machine Learning Based Safety-critical Systems?\thanks{This work is supported by the DEEL project CRDPJ 537462-18 funded by the National Science and Engineering Research Council of Canada (NSERC) and the Consortium for Research and Innovation in Aerospace in Québec (CRIAQ), together with its industrial partners Thales Canada inc, Bell Textron Canada Limited, CAE inc and Bombardier inc.}
}
\subtitle{A Systematic Literature Review}


\author{Florian Tambon\affmark[1] \and Gabriel Laberge\affmark[1] \and Le An\affmark[1] \and Amin Nikanjam\affmark[1] \and Paulina Stevia Nouwou Mindom\affmark[1] \and Yann Pequignot\affmark[2] \and Foutse Khomh\affmark[1] \and Giulio Antoniol\affmark[1] \and Ettore Merlo\affmark[1] \and François Laviolette\affmark[2]}

\authorrunning{Tambon et al.} 

\institute{
              \affmark[1] Polytechnique Montréal, Québec, Canada \\
              \email{\{florian-2.tambon, gabriel.laberge, le.an, amin.nikanjam, paulina-stevia.nouwou-mindom, foutse.khomh, giuliano.antoniol, ettore.merlo\}@polymtl.ca}  \\
               \affmark[2] Laval University, Québec, Canada \\
               \email{francois.laviolette@ift.ulaval.ca, yann.pequignot@iid.ulaval.ca}
}

\date{Received: date / Accepted: date}

\maketitle

\begin{abstract}

\textit{Context:} Machine Learning (ML) has been at the heart of many innovations over the past years. However, including it in so-called \enquote{safety-critical} systems such as automotive or aeronautic has proven to be very challenging, since the shift in paradigm that ML brings completely changes traditional certification approaches. 

\textit{Objective:} This paper aims to elucidate challenges related to the certification of ML-based safety-critical systems, as well as the solutions that are proposed in the literature to tackle them, answering the question \enquote{How to Certify Machine Learning Based Safety-critical Systems?}. 

\textit{Method:} We conduct a Systematic Literature Review (SLR) of research papers published between 
2015 and 2020, covering topics related to the certification of ML systems. 
In total, we identified 217 papers covering topics considered to be the main pillars of ML certification: \textit{Robustness}, \textit{Uncertainty}, \textit{Explainability}, \textit{Verification}, \textit{Safe Reinforcement Learning}, and \textit{Direct Certification}. We analyzed the main trends and problems of each sub-field and provided summaries of the papers extracted.

\textit{Results:} The SLR results highlighted the enthusiasm of the community for this subject, as well as the lack of diversity in term of datasets and type of ML models. It also emphasized the need to further develop connections between academia and industries to deepen the domain study. Finally, it also illustrated the necessity to build connections between the above mention main pillars that are for now mainly studied separately.

\textit{Conclusion} We highlighted current efforts deployed to enable the certification of ML based software systems, and discuss some future research directions.
\keywords{Machine Learning, Certification, Safety-critical, Systematic Literature Review}

\end{abstract}

\section{Introduction}\label{sec:introduction}
 
Machine Learning (ML) is drastically changing the way we 
interact with the world. We are now using software applications powered by ML in critical aspects of our daily lives; from finance, energy, to health and transportation. 
Thanks to frequent innovations in domains like Deep Learning (DL) and Reinforcement Learning (RL), the adoption of ML is expected to keep rising and 
the economic benefits of systems powered by ML is forecast to reach 
\$30.6 billion by 2024\footnote{\url{https://www.forbes.com/sites/louiscolumbus/2020/01/19/roundup-of-machine-learning-forecasts-and-market-estimates-2020/}}. 
However, the integration of ML in systems is not without risks, especially in \textit{safety-critical} systems, where any system failure can lead to catastrophic events such as death or serious injury to humans, severe damage to equipment, or environmental harm, as for example in avionic or automotive\footnote{\url{https://www.cbc.ca/news/business/uber-self-driving-car-2018-fatal-crash-software-flaws-1.5349581}}. Therefore, before applying and deploying any ML based components into a safety-critical system, these components need to be certified.

\textcolor{black}{Certification is defined as \enquote{procedure by which a third-party gives written assurance that a product, process, or service conforms to specified requirements} \cite{Rodriguez99}.} This certification aspect is even more important in the case of safety-critical systems such as plane or car. If this certification aspect is preponderant for mechanical systems, it has also been a mandatory step for the inclusion of software or electronic related component, especially with the rise of embedded software. As such, standards were developed in order to tackle this challenge: IEC 61508\footnote{\url{https://webstore.iec.ch/publication/6007}} developed as an international standard to certify electrical, electronic, and programmable electronic safety related systems with ISO 26262\footnote{\url{https://www.iso.org/standard/68383.html}} being the adaptation (and improvement) of this standard applied specifically to road vehicles safety or DO-178C\footnote{\url{https://my.rtca.org/NC__Product?id=a1B36000001IcmqEAC}}\cite{do178c} created for the certification of airborne systems and equipment, introducing for instance Modified Conditions/Decision Coverage (MC/DC) criteria as a requirement, which goal is to test all the possible conditions, contributing to a certain decision. Those widely-used standards generally aim to deal with the functional safety consideration of a safety-critical system. They provide guidelines, risk-level and a range of requirements that need to be enforced in order to reach a certain level of safety based on the criticality of the system and/or components inside the whole architecture.

However, the introduction of ML components in \textcolor{black}{software} systems changes the game completely. While ML can be extremely useful as it conveys the promise of replicating human knowledge with the power of a machine, it also introduces a major shift in software development and certification practices. Traditionally, software systems are constructed deductively, by writing down the rules that govern the behavior of the system as program code. With ML techniques, these rules are generated in an inductive way (i.e., \textit{learned}) from training data. With this paradigm shift, the notion of specification, \textcolor{black}{which used to apply solely to the code itself, must now encompass the data and the learning process as well, therefore making most} of previously defined standards not applicable \textcolor{black}{to those new ML software systems}. In the light of this observation, the scientific community has been working to define new standards specific to the unique nature of ML \textcolor{black}{software} applications (see \textbf{Section \ref{sec:background}}). \textcolor{black}{Yet, as of now, no certification method -- in the previously defined sense -- exists for ML \textcolor{black}{software systems}.}

This paper aims to provide a snapshot of the current answers to the question \enquote{How to certify machine learning based safety-critical systems?} by examining the progress made over the past years with an emphasis on the transportation domain. ML based safety-critical \textcolor{black}{software} systems here designates any safety-critical \textcolor{black}{software} system that is partly or completely composed of ML components. While we want to shed light on recent advances in different fields of ML regarding this question, we also aim at identifying current gaps and promising research avenues that could help to reach the community's goal of certifying ML based safety-critical \textcolor{black}{software} systems. To achieve this objective, we adopt the Systematic Literature Review (SLR) approach, which differs from traditional literature reviews, by its rigour and strict methodology which aim to eliminate biases when gathering and synthesizing information related to a specific research question \cite{kitchenham2004procedures}. 

The paper is structured as follows. \textbf{Section \ref{sec:background}} discusses key concepts related to the content of the paper, \textbf{Section \ref{sec:methodology}} describes in depth the SLR methodology; explaining how we have applied it to our study. \textbf{Section \ref{sec:results:stats}} presents descriptive statistics about our collected data and \textcolor{black}{\textbf{Section \ref{sec:results:tax}} details the taxonomy derived from the studied papers, with explanations about the different categories of the given taxonomy}. In \textbf{Section \ref{sec:review}}, we leverage the collected data and present a broad spectrum of the progresses made and explain how they can help certification. We also discuss current limitations and possible leads. We subdivide this section as follow; \textbf{Section \ref{robustness}} is about \textit{Robustness}, \textbf{Section \ref{uncertainty}} is about \textit{Uncertainty} and \textit{Out-of-Distribution} (OOD), \textbf{Section \ref{explainability}} is about \textit{Explainability} of models, \textbf{Section \ref{verification}} is about \textit{Verification} methods which are divided between \textit{Formal} and \textit{Non-Formal} (Testing) ones, \textbf{Section \ref{safe_rl}} is about \textit{Safety considerations in Reinforcement Learning} (RL), \textbf{Section \ref{direct_certif}} deals with \textit{Direction Certifications} proposals. \textbf{Section \ref{sec:discussion}} summarizes key lessons derived from the review. \textcolor{black}{\textbf{Section \ref{sec:rel}} gives a brief overview of related literature review existing on the subject and how they differ from our work. \textbf{Section \ref{sec:threats}} details the threats to the validity of the paper}. Finally \textbf{Section \ref{sec:conclusion}}, concludes the paper.

\section{Background}\label{sec:background}

ML is a sub-domain of Artificial Intelligence (AI) that makes decisions based on information learned from data \cite{Zhang19-3}. \textcolor{black}{The \enquote{learning} part introduces the major paradigm shift from classical software, where the logic is coded by a human, and is the quintessential reason why ML cannot currently comply with the safety measures of safety-critical systems.}
ML is generally divided into three main categories; \emph{supervised} learning, \emph{unsupervised} learning, and \emph{reinforcement} learning. The difference between those denominations comes from the way the model actually processes the data. In supervised learning, \textcolor{black}{the model is given data along with desirable outputs and the goal is to learn to compute the supplied outputs from the data. For a classification task, this amounts to learning boundaries among the data -- placing all inputs with the same label inside the same group (or class) -- and while for a regression task, this amounts to learning the real valued output for each input data.} For example, a self-driving car must be able to recognise a pedestrian from another car or a wall. In unsupervised learning, the model does not have access to labels and has to learn \textcolor{black}{patterns  (\eg{} clusters) directly from data}. In RL, the model \textcolor{black}{learns to evolve and react in an environment based only on reward feedback}, so there is no fixed dataset as the data is being generated continuously from interaction with the environment. For example, in autonomous driving, a lane-merging component needs \textcolor{black}{to learn how to make safe decisions using only feedback from the environment which provide information on 
the potential consequences of an action leading to learning best actions. Among ML models, Neural Networks (NN) have become very popular recently. NN are an ensemble of connected layers of neurons, where neurons are computing units composed of weights and a threshold of activation to pass information to the next layer. Deep Neural Networks (DNN) are simply a \enquote{deeper} version of NN (\ie{}, with multiple layers/neurons), which allow for tackling the most complex tasks.}

Safety certification of software and/or electronic component is not new: the first version of IEC 61508 was released around 1998 to tackle those considerations and was later upgraded to match the evolution of the state-of-the-art. However, it is only recently that the specificity of ML was acknowledged with the improvement of the techniques and the increased usage of the paradigm as a component in bigger systems, with for instance EASA (European Union Aviation Safety Agency) releasing a Concept Paper \enquote{First usable guidance for Level 1 machine learning applications}\footnote{https://www.easa.europa.eu/newsroom-and-events/news/easa-releases-consultation-its-first-usable-guidance-level-1-machine} following its AI Roadmap. Nonetheless, to this day, there are no released standards that tackle specifically ML certification.

ISO 26262 Road vehicles — Functional safety \cite{iso26262}, tackles issue of safety-critical systems that contain one or more electronic/electrical components in passenger cars. They analyze hazard due to malfunction of such components or their interaction and discuss how to mitigate them. It was designed to take into account the addition of ADAS (Advanced Driver Assistance System), yet it doesn't acknowledge the specificity of ML. In the same train of thoughts but for aeronautics, DO-178C is the primary document used for certification of airborne systems and equipment. Those two standards are based on risk-level and a range of criteria that needs to be matched to reach a certain risk-level that has to be correlated to the impact of the component on the system and the consequences in case of malfunction.\\
ISO/PAS 21448 Safety of the intended functionality (SOTIF)\cite{sotif} is a recent standard that took a different approach: instead of tackling functional safety, this one aims for \textit{intended} functionality, \ie{},  establishing when an unexpected behavior is observed in absence of fault in the program (in the traditional sense). Indeed, a hazard can happen even without a system failure and this standard attempts to take into consideration the impact of an unknown scenario that the system was not exactly built to tackle. It was also designed to be applied for instance to Advanced Driver Assistance System (ADAS) and brings forward an important reflection about ML specification: our algorithm can work as it was trained to but not as we intended to.
Yet, the standard does not offer strict criteria that could be correlated to problems affecting ML algorithms such as robustness against out-of-distribution, adversarial examples, or uncertainty.

Recently, it seems the ISO organization is working towards more ML/AI oriented standards: in particular, ISO Technical Committee dealing with AI (ISO/IEC JTC 1/SC 42)\footnote{\url{https://www.iso.org/committee/6794475/x/catalogue/}} lists to this date 7 published standards dedicated to AI subjects and 22 more are under development. One example is 
ISO/IEC 20546:2019\footnote{\url{https://www.iso.org/standard/68305.html?browse=tc}} which defines a common vocabulary and provides an overview of the Big Data Field. Another example is ISO/IEC TR 24028:2020\footnote{\url{https://www.iso.org/standard/77608.html?browse=tc}} which gives an overview of trustworthiness in AI  or ISO/IEC TR 24029-1:2021\footnote{\url{https://www.iso.org/standard/77609.html?browse=tc}} which provides background about robustness in AI. All the above mentioned papers only offer an overview of possible directions without proposing detailed specification/risk levels for each critical property of 
ML systems. The situation could be improved with the advent of standards such as ISO/IEC AWI TR 5469 Artificial intelligence — Functional safety and AI systems\footnote{\url{https://www.iso.org/standard/81283.html?browse=tc}} which seem to be in direct relation with ISO 26262; however not much information is available at the moment, as they still are under development. 

In parallel to these standards, the scientific community and various organizations have tried to come up with ways to characterize the paradigm shift that ML induces to systems and to tackle these unique issues. In \cite{Gualo21}, the authors aim to certify the level of quality of a data repository with examples of data quality from three organizations using SQL databases: they use as a base, the ISO/IEC 25012 which defines data quality and characteristics and ISO/IEC25024 which bridges those concepts of data quality with \enquote{quality property} in order to evaluate data quality. 
The FAA introduced the notion of Overarching Properties\footnote{\url{https://www.faa.gov/aircraft/air_cert/design_approvals/air_software/media/TC_Overarching.pdf}} as an alternative to DO-178C that could be better adapted to ML components and which is based on three main sufficient properties: intent (\enquote{the defined intended behavior is correct and complete with respect to the desired behavior}), correctness (\enquote{the implementation is correct with respect to its defined intended behavior under foreseeable operating conditions}), and acceptability (\enquote{any part of the implementation that is not required by the defined intended behavior has no unacceptable safety impact}), which can be summed up as follows: be specified correctly (intent), do the right things (correctness) and do no wrong (acceptability). However, as there is no clear checklist on what criteria to comply to, it is more complicated to enforce in practice. As of now, the approach does not seem to have been adopted widely.

To help tackle scientific challenges related to the certification of ML based safety-critical systems, the 
DEEL (DEpendable \& Explainable Learning) project\footnote{\url{https://www.deel.ai}}, which is born from the international collaboration between the Technological Research Institute (IRT) Saint Exup\'{e}ry in Toulouse (France), the Institute for Data Valorisation (IVADO) in Montreal (Canada) and the Consortium for Research and Innovation in Aerospace of Qu\'{e}bec (CRIAQ) in Montreal (Canada),
aims to develop novel theories, techniques, and tools to help ensure the Robustness of ML based systems (i.e., their efficiency even outside usual conditions of operation), their Interpretability (i.e., making their decisions understandable and explainable), Privacy by Design (i.e., ensuring data privacy and confidentiality during design and operation), and finally, their Certifiability.
A white paper \cite{Delseny21} released in 2021 introduces the challenges of certification in ML.

\section{Methodology}\label{sec:methodology}
We planned, conducted, and reported our SLR based on the guidelines provided in \cite{kitchenham2004procedures}. In the rest of this section, we will elaborate on our search strategy, the process of study selection, quality assessment, and data analysis.

\subsection{Search Terms}\label{sec:search_terms}
Our main goal is to identify, review, and summarize the state-of-the-art techniques used to certify ML based applications in safety-critical systems. To achieve this, we select our search terms from the following aspects:
\begin{itemize}
	\item \textbf{Machine learning}: Papers must discuss techniques for ML systems. 
	As described in Section \ref{sec:background}, ML encompasses any algorithm whose function is \enquote{learned} through information the system received. We used the terms \emph{machine learning}, \emph{deep learning}, \emph{neural network}, and \emph{reinforcement learning} to conduct our searches. To include more potential papers, we also used \emph{black box} (DL models are often considered as non-interpretable black boxes~\cite{castelvecchi2016can,shwartz2017opening}, so certification approaches for such black boxes can be useful), \emph{supervised} and \emph{unsupervised} (because people may use the term \enquote{(un)supervised learning} to refer to their ML algorithm).
	\item \textbf{Safety-critical}: As described in Section \ref{sec:introduction}, we call safety-critical any system where failures can lead to catastrophic consequences in term of material damage or victims. In this study, we consider that the systems, which need to be certified, are safety-critical, such as self-driving, avionic, or traffic control systems. We use the terms \emph{safety critical} and \emph{safety assurance} to search for this aspect.
	\item \textbf{Certification}: As described in Section \ref{sec:introduction}, \enquote{certification} refers to any process that aims at ensuring the reliability and safety of a system. The key term we need to search for is \emph{certification} itself. However, through some experiments, we saw that using this term alone will yield a lot of irrelevant results, \eg{} \enquote{Discover learning behavior patterns to predict certification} \cite{zhao2016discover}, where \enquote{certification} denotes an educational certificate. We also observed that two major topics of papers can be returned using the terms of \emph{certification}, \emph{certify}, or \emph{certified}, \ie{} medical related papers and transportation related papers. In this study, we are particularly interested in the certification of the transport systems, therefore we use the following terms to limit the certification scope: \emph{automotive}, \emph{driv*}, \emph{pilot}, \emph{aerospace}, and \emph{avionic}.
\end{itemize}
Figure \ref{fig:search_pattern} shows all the search terms used and their logical relationship.

\begin{figure*}[t]
    \centering
        \begin{lstlisting}
( 
  ("safety critical" OR "safety assurance") 
  OR 
  ( 
    (certification OR certified OR certify) AND (automotive OR drive* OR driving OR pilot OR aerospace OR avionic) 
  )
)
AND 
("machine learning" OR "deep learning" OR "neural network*" OR "black box" OR "reinforcement learning" OR supervised OR unsupervised)
        \end{lstlisting}
    \caption{Search terms used in our study and their logical relationship. The terms in quotes denote the exact match to a phrase.}
	\label{fig:search_pattern}
\end{figure*}

\subsection{Scope}

As the subject of certification in ML can be pretty vast, we chose to restrict ourselves to any methods that tackle certification efforts at the algorithm/system structure level of a ML system from the lens of transportation systems. As such, we will not discuss:
\begin{itemize}
    \item \textbf{Hardware}: If hardware is also one possible source of error for ML applications, we will not consider them in this SLR, as the root cause of such an error is not necessarily the ML system itself.
    \item \textbf{Security/Privacy}: Security threats are an important point of safety-critical systems. However, such considerations are not limited to ML systems as security is a major issue in all systems. As such, we will not develop on security related problems in ML.
    \item \textbf{Performance only}: As the end goal is to certify the safety of ML system, papers only describing an improvement of performance such as accuracy without any safety guarantees will not be considered in this review.
    \item \textbf{Techniques tied to a specific field} (other than transportation): while we aim for general certification in ML, we mainly focus on the transportation field. If a technique cannot be extended to transportation problems, \eg{} entirely revolves around a particular field, it will be excluded.
\end{itemize}

\subsection{Paper search}
Inspired by \cite{SYRIANI201843, WEN201241}, we searched papers from the following academic databases:
\begin{itemize}
	\item Google Scholar\footnote{\url{https://scholar.google.com}}
	\item Engineering Village (including Compendex)\footnote{\url{https://www.engineeringvillage.com}}
	\item Web of Science\footnote{\url{https://webofknowledge.com}}
	\item Science Direct\footnote{\url{https://www.sciencedirect.com}}
	\item Scopus\footnote{\url{https://www.scopus.com}}
	\item ACM Digital Library\footnote{\url{https://dl.acm.org}}
	\item IEEE Xplore\footnote{\url{https://ieeexplore.ieee.org}}
\end{itemize}

Certification of ML systems is a relatively new topic. To review the state-of-the-art techniques, we limited the publication date from January 2015 to September 2020 (the month when we started this work) in our searches. The threshold of 5 years was chosen as it is a decent time period in order to have a good snapshot of the most recent advances in the field, while limiting the potential number of papers that could be returned. Moreover, as can be seen on Figure \ref{fig:yearly_count}, keywords related to certification became prevalent only after 2017. Other SLRs (e.g., by Kitchenham et al. \cite{Kitchenham10}) have used a similar time range. As searching key terms throughout a paper may return a lot of noise. For example, an irrelevant paper entitled \enquote{Automated Architecture Design for Deep Neural Networks} \cite{abreu2019automated} can be retrieved using the terms mentioned in Section \ref{sec:search_terms}, because some terms can appear in the content of the paper with other meanings (where the key terms are highlighted in bold):
\emph{\enquote{... With my signature, I \textbf{certify} that this thesis has been written by me using only the in ... \textbf{Deep learning} is a subfield of \textbf{machine learning} that deals with deep artificial \textbf{neural networks} ... in a network to work with different combinations of other hidden units, essentially \textbf{driving} the units to}}. 
The sentence \emph{\enquote{I \textbf{certify} that this thesis}} appears in the declaration page, which misled our result.
Therefore, we restricted our searches only from papers' title, abstract, and keywords. Note that we adapted the pattern depending on the requirements of the database we searched on.

We leveraged the \enquote{Remove Duplicates} feature of Engineering Village, by keeping Compendex results over Inspec, to reduce our workload in the step of manual paper selection, described in the next section.

Unlike other academic databases, Google Scholar (GS) does not provide any API or any official way to output the search result. It only allows users to search terms from the paper title or from the full paper. Also, GS does not encourage data mining with it and can at most return 1,000 results per search. To tackle these issues, we:
\begin{enumerate}
	\item Used the \enquote{Publish or Perish} tool\footnote{Harzing, A.W. (2007) Publish or Perish, available from \url{https://harzing.com/resources/publish-or-perish}} to automate our search process.
	\item Performed searches year by year. Using the above two expressions, we can perform two searches per year, which in turn increases the number of our candidate papers. As we selected to review papers from 2015 to 2020 (six years in total), GS can provide us with at most 1,000 $\times$ 6 $\times$ 2 = 12,000 results.
\end{enumerate}

\subsection{Paper selection}
Table \ref{tab:raw_papers} shows the number of results returned from each academic database we used. In the rest of this section, we will describe the steps we used to filter out irrelevant papers and to select our final papers for review.
\begin{table}[t]
	\centering
	\small
	\caption{Number of papers retrieved from the academic databases.}
	\begin{tabular}{ | l | r | }
		\hline
		\textbf{Database} & \textbf{Number of papers} \\ \hline	
 		Google Scholar & 11,912 \\ \hline
		Engineering village & 704 \\ \hline
		Web of Science & 146 \\ \hline
		Science Direct & 62 \\ \hline
		Scopus & 195 \\ \hline
		ACM Digital Library & 154 \\ \hline
		IEEE Xplore & 291 \\ \hline
		Total & 13,464 \\ \hline
	\end{tabular}
	\label{tab:raw_papers}
\end{table}

\subsubsection{First Round of Google Scholar Pruning}
Google Scholar (GS) returned a lot of results but many of them are irrelevant because GS searches terms from everywhere. We thus performed another round of term search within the results of GS by only considering their paper title and the part of the abstract returned by GS results. At this step, 1,930 papers remained. We further noticed that some surviving results had incomplete or truncated titles, from which we can hardly trace back to the correct papers. Therefore, we manually examined the paper titles and kept 1,763 papers for further pruning. 

\subsubsection{Pooling and Duplicate Filtering}
We pooled together the surviving papers (1,763 papers from GS and 1,551 papers from other databases). Then, we used \emph{EndNote}\footnote{\url{https://endnote.com}} to filter duplicates. 
To avoid any incorrect identification, EndNote does not remove duplicates but instead it shows the results to users and asks them to remove duplicates. This mechanism prevents false positives (two different papers can be mistakenly identified as duplicates by the system). 
There were 727 papers identified as duplicates, from which we manually removed 398 papers. 
Thus, 2,914 papers (1,741 from GS and 1,173 from other databases) survived at this step.

\subsubsection{Second Round of Google Scholar Pruning}
To further remove irrelevant results from GS, two of the authors manually examined the title and abstract of the GS papers that survived the previous step. They independently identified any potential papers that are related to the topic of certification of ML systems. Knowing that title and abstract alone may not show the full picture of a paper, to avoid missing any valuable paper, the examiners intentionally included all papers \textcolor{black}{that were somehow related to our topics.}

The two examiners then compared their results and resolved every 
conflicts through meetings until they reached agreement on all surviving papers. As a result, 290 papers survived in this step. Therefore, we obtained a total of 
290 + 1,173 = 1,463 papers for further analysis. 

\subsubsection{Inclusion and Exclusion criteria}
To select our final papers for review, two of the authors read title, abstract, and eventually, introduction/conclusion of each surviving paper by considering the following inclusion and exclusion criteria.

\underline{Inclusion criteria:}
\begin{itemize}
    \item Techniques that directly strive to certify a ML based safety-critical system, \ie{}, making ML models comply with a safety-critical standard.
    \item Techniques that indirectly help to certify a ML based safety-critical system, \ie{}, general techniques that improve the robustness, interpretability, or data privacy of such systems.
\end{itemize}

\underline{Exclusion criteria:}
\begin{itemize}
    \item Papers leveraging ML to improve other techniques, such as software testing, distracted driving detection. This is on the contrary of our purpose, which is to discover techniques that improve ML models' certifiability.
    \item Papers that thrive to solely improve a prediction metric such as accuracy, precision, or recall.
    \item Techniques whose scope of application cannot fit into transportation related systems, \ie{}, we do not consider techniques that are strictly limited to other fields, such as medicine.
    \item Survey, review papers, research proposals, and workshop or position papers. 
    \item Papers not written in English. Understanding and reviewing papers in other languages might be helpful but is out of the scope of our study.
    \item Papers which we cannot have access free of charge through our institution subscriptions.
\end{itemize}

Based on the above inclusion/exclusion criteria, two of the authors independently examined the surviving papers. As a result, 1,000 papers were rejected and 163 papers were accepted by both examiners. \textcolor{black}{Meetings were then held} to resolve the 300 conflicts, among which 86 papers were accepted.
Finally, \textbf{249 papers were selected for reading and analysis}.

\subsection{Quality Control Assessment}

The next step is to further control papers' relevance by using quality-control criteria to check that they could potentially answer our research questions, in a clear and scientific manner. As such we devised the following criteria, some of which are inspired from previous systematic literature review studies \cite{DYBA2008833,WEN201241}:\\
\underline{Quality control questions (common to many review papers):}
\begin{enumerate}
    \item Is the objective of the research clearly defined?
    \item Is the context of the research clearly defined?
    \item Does the study bring value to academia or industry?
    \item Are the findings clearly stated and supported by results?
    \item Are limitations explicitly mentioned and analyzed?
    \item Is the methodology clearly defined and justified?
    \item Is the experiment clearly defined and justified?
\end{enumerate}
\underline{Quality control questions (particular to our topic):}
\begin{enumerate}
    \item Does the paper propose a direct approach to comply with a certain certification standard?
    \item[2a)] Does the paper propose a general approach that can help certify a safety-critical system?
    \item[2b)] If it's a general approach, was it experimented on a transport related system?
\end{enumerate}

We proceeded as before, by evaluating each paper for each criterion, listing them with a score of \emph{0}, \emph{1}, or \emph{2}, with \emph{2} denoting a strong agreement with the statement of the control question. To assess the quality of the papers we read every paper entirely.  
Each paper was controlled by two different reviewers. For a paper to get validated, each reviewer had to assign  a minimum of 7 for the common Quality Control Questions and 1 for the particular Quality Control Questions. If reviewers end up not agreeing, a discussion would ensue to reach an agreement, with a third reviewer deciding if the reviewers could not agree after discussion. For those particular Quality Control Questions, note that; first, (1) and (2) are mutually exclusive, so a paper can only score in one of the two and secondly, (2a) and (2b) are complementary, as such their score is averaged to have the result of question (2). 
 
Out of 249 papers, 24 papers were rejected before Control Questions were assessed as they were found not to comply with the scope and requirements of our SLR after further reading. Hence, 225 papers were considered for Control Questions step. Following the above mentioned protocol, we further pruned 12 papers which score was too low. As such, we end up with \textbf{213 papers}.

\subsubsection{Snowballing}
The snowballing process aims to further increase the papers pool by screening relevant references from selected papers. Four interesting papers on ML certification were not discovered by our keywords but were mentioned multiple times in our selected papers as references. We also included these papers in our reading list for analysis after performing similar steps as before on those papers to make sure they would fit our methodology. Hence, in total we collected \textbf{217 papers} for examination in our review.

\subsection{Data Extraction}
For each paper, we extracted the following information: title, URL to the paper, names(s) of author(s), year of publication, and publication venue.
Aside from Quality Control Questions, each reviewer was assigned a set of questions to answer based on the paper under review to help the process of data extraction:
\begin{itemize}
    \item Does the paper aim at directly certify ML based safety-critical systems?
    If yes, what approach(es) is(are) used? Simply describe the idea of the approach(es)
    \item Does the paper propose a general approach?
    If yes, what approach(es) is(are) used? Simply describe the idea of the approach(es)
    \item What are the limitations/weaknesses of the proposed approaches?
    \item What dataset(s) is(are) used?
\end{itemize}

Those questions will help us elaborate into the \textbf{Section \ref{sec:discussion}}. We have prepared a replication package that includes all data collected and processed during our SLR. This package covers information of selected papers, answers to quality control questions, and summary of papers. The package is available online\footnote{\url{https://github.com/FlowSs/How-to-Certify-Machine-Learning-BasedSafety-critical-Systems-A-Systematic-Literature-Review}}.

\subsection{Authors Control}

We further asked multiple authors whose paper we considered in our review, and from which we wanted further precisions, to give us feedback on the concerned section, in order to make sure we had the right understanding of the developed idea. This allowed us to further improve the resilience of our methodology and review.

\section{Statistical results}\label{sec:results:stats}

\subsection{Data Synthesis}

From the extracted data, we present statistical description about the pool of papers.
\begin{figure}[t]
    \centering
    \includegraphics[width=0.55\textwidth]{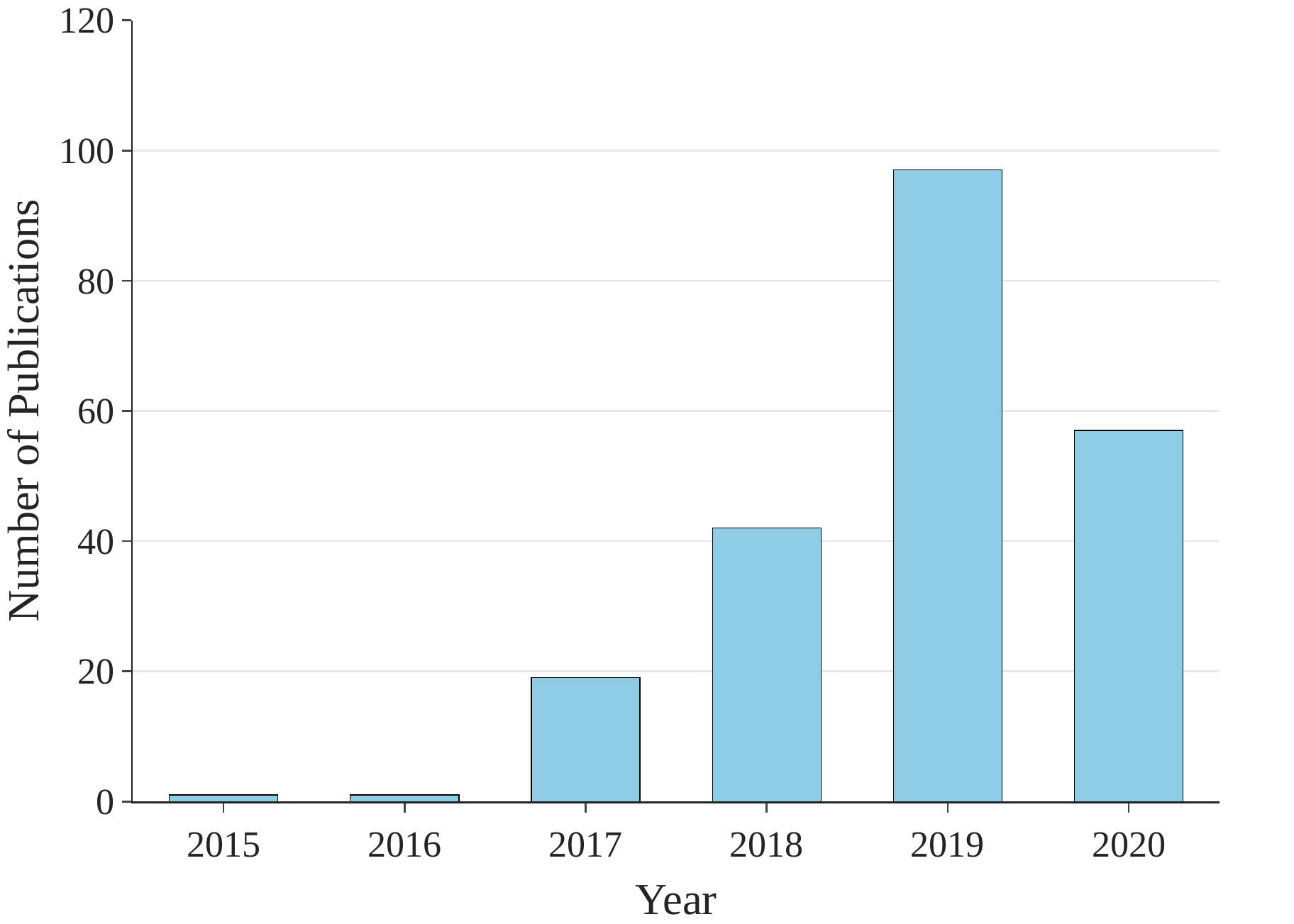}
    \caption{Number of studied papers on certification topics per year from 2015 to 2020.}
    \label{fig:yearly_count}
\end{figure}
\begin{figure}[t]
    \centering
    \begin{minipage}{.5\textwidth}
        \centering
        \includegraphics[width=\linewidth]
        {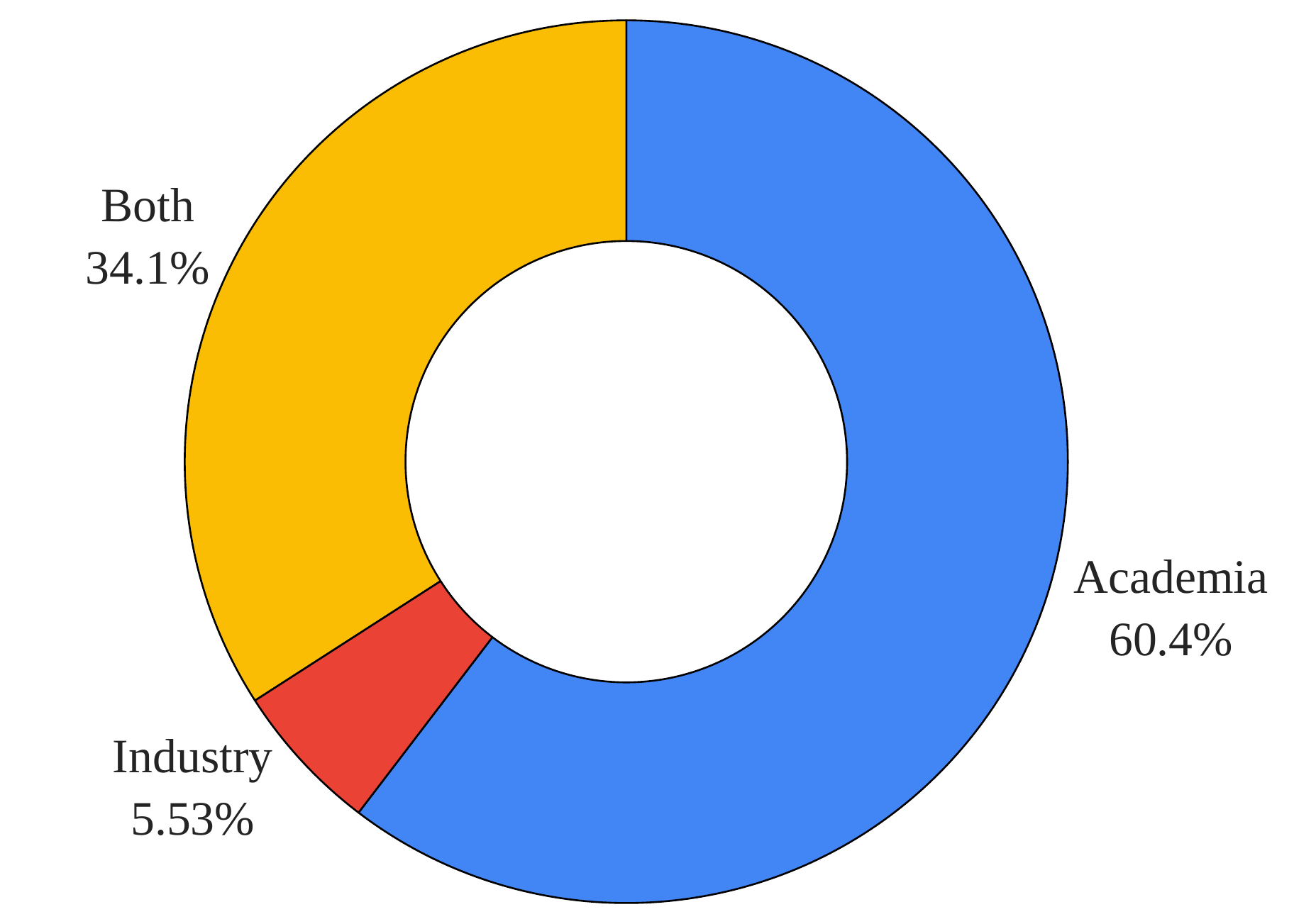}
        \caption{Authors' Affiliation Distribution.}
    \label{fig:university_industry}
    \end{minipage}%
    \begin{minipage}{0.5\textwidth}
        \centering
        \includegraphics[width=\linewidth]
        {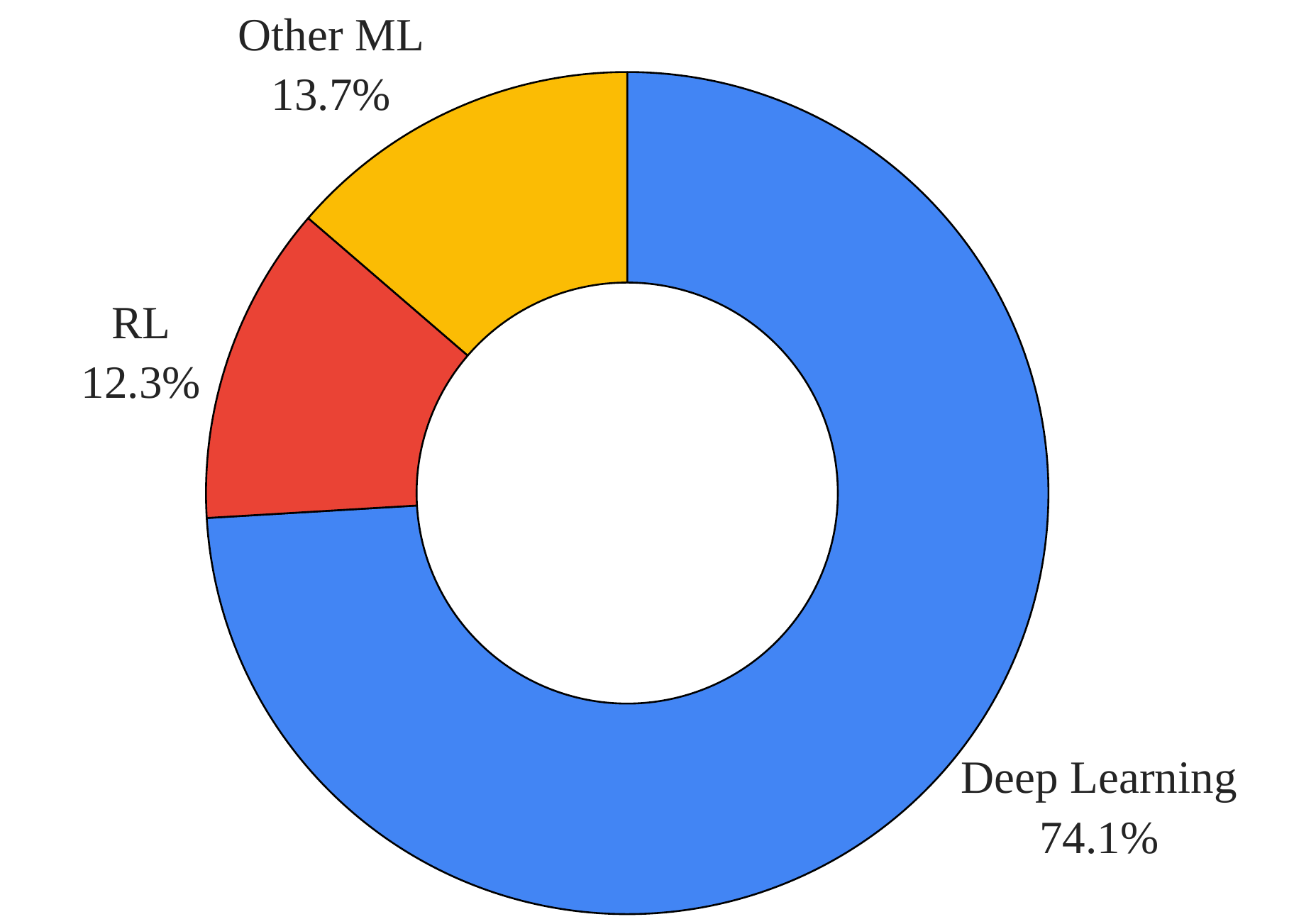}
        \caption{Categories of the studied papers.}
        \label{fig:ml_algo}
    \end{minipage}
\end{figure}
Figure \ref{fig:yearly_count} shows the number of selected papers published in each year from 2015 to 2020. We observed a general increasing trend of the papers related to the certification topics. The number of papers in 2020 is less than those in 2019 because our paper search was conducted until September 2020. This increasing trend shows that the topic of using ML technique for safety-critical systems attracts more attention over the years. As displayed in Figure \ref{fig:university_industry}, we found the topics were well investigated by industrial practitioners or researchers because 40\% of the papers were either from companies alone or from a collaboration between universities and companies. This information was deduced by screening papers for author's affiliation and potential explicit industrial grants mentioned.

In terms of studied models in the reviewed papers, Figure \ref{fig:ml_algo} shows that nearly three-quarters of the papers employed (deep) Neural Networks (NN). These models received great attention recently and have been applied successfully to a wide range of problems. Moreover, they benefit from a huge success on some popular classification and regression tasks, such as image recognition, image segmentation, obstacle trajectory prediction or collision avoidance. More than 12\% of papers studied RL since it has been well investigated for its transportation-related usage. In particular, researchers intended to apply RL to make real-time safe decisions for autonomous driving vehicles \cite{hart2019lane,lutjens2019safe,baheri2019deep}.

Figure \ref{fig:overview} illustrates the categorization of the studied paper. \textcolor{black}{Categories will be developed and defined in the next Section.} Robustness and Verification are the most popularly studied problems. Some of the categories can be further split into sub-categories. For example, ML robustness includes the problems of Robust training and Post-training analysis. Similarly, ML verification covers testing techniques and other verification methods \eg{} formal methods. However, we only extracted 14 papers proposing a direct certification technique for ML based safety-critical systems. In other words, although researchers are paying more attention to ML's certification (illustrated by Figure \ref{fig:yearly_count}), most of our reviewed approaches can only solve a specific problem under a general context (in contrast with meeting all safety requirements of a standard under a particular use scenario, such as autonomous driving or piloting). 

\begin{figure*}[t]
    \centering
    \includegraphics[width=1.0\textwidth]
    {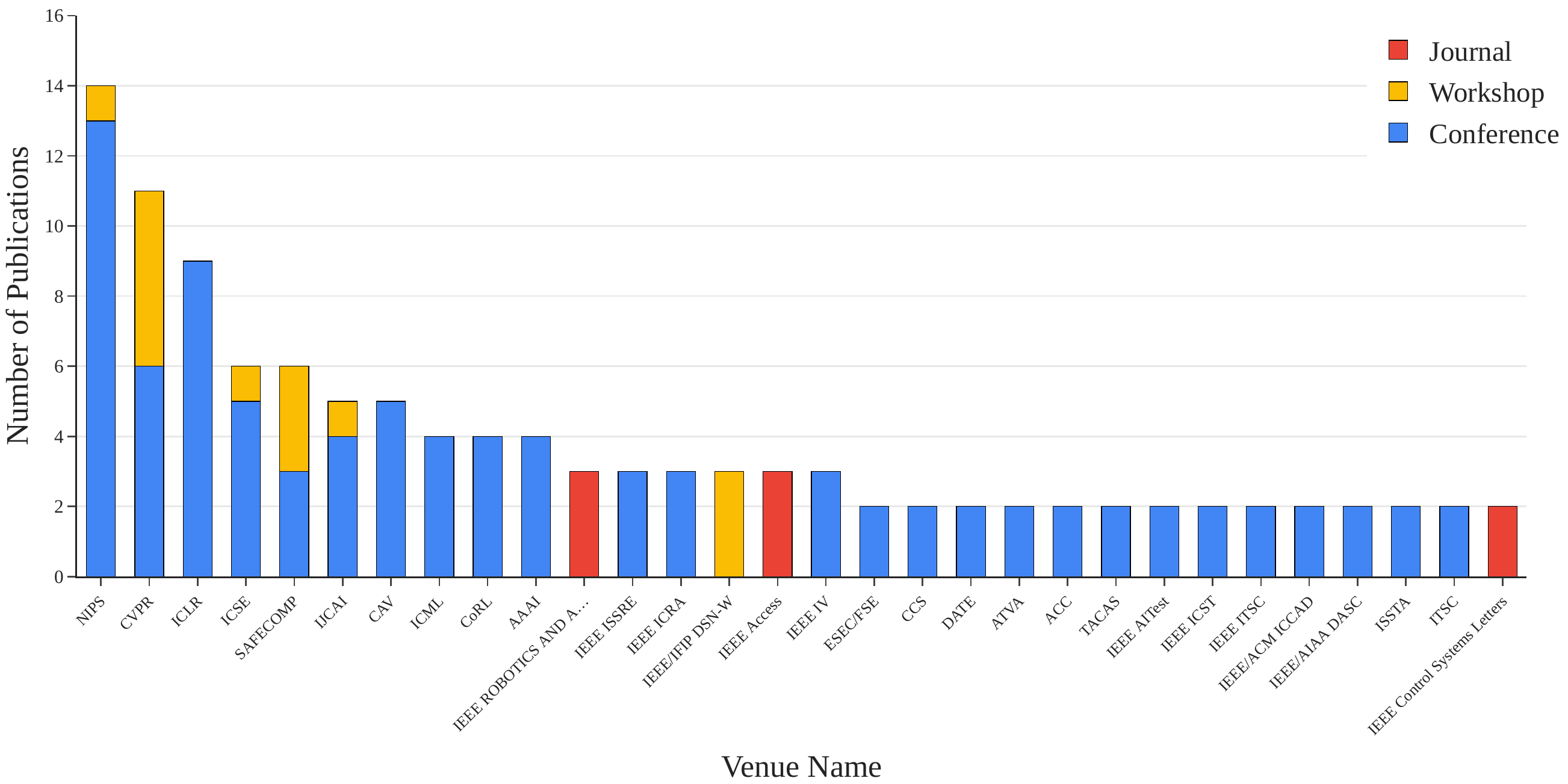}
    \caption{Papers venue representation in our selected papers. Only venues being represented more than 2 times are shown for readability purposes.}
    \label{fig:paper_venues}
\end{figure*}

Figure \ref{fig:paper_venues} shows the papers distribution across venues. If ML related ones are vastly represented with NIPS, CVPR and ICLR being the most important ones, papers also come from a wide range of journals/conferences across the board, mainly in computer science/engineering related fields. 

\begin{figure*}[t]
    \centering
    \includegraphics[width=1.0\textwidth]
    {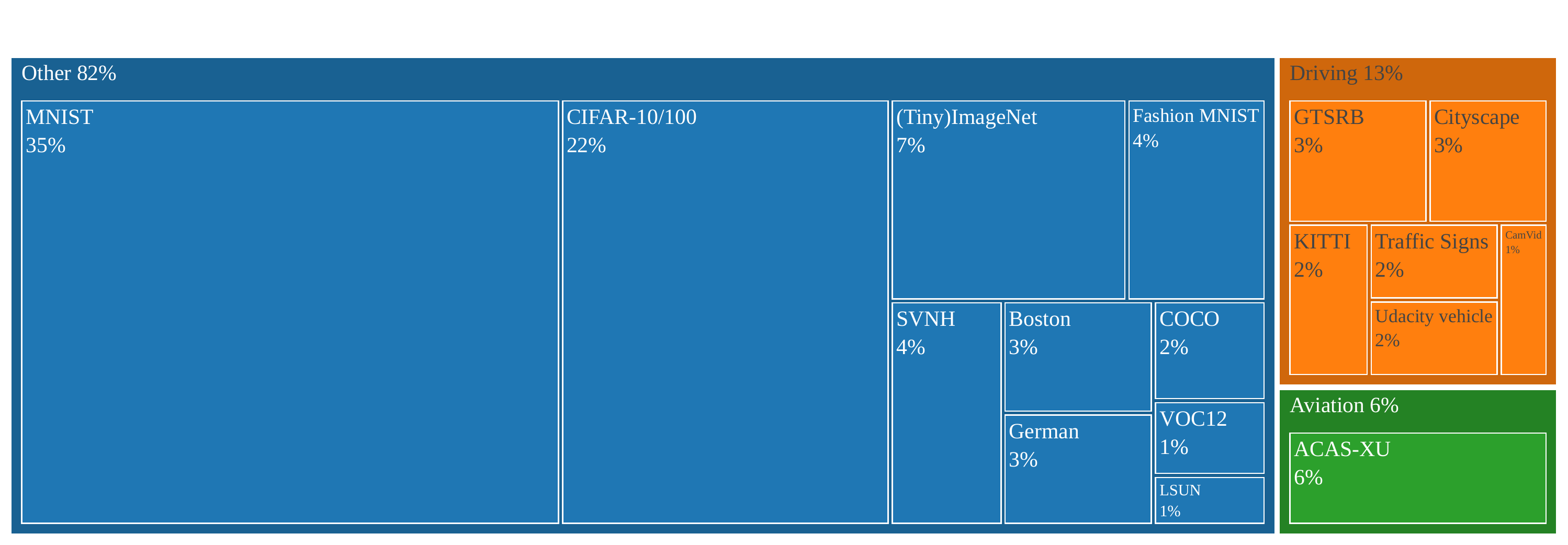}
    \caption{Dataset repartition by field (Autonomous Driving, Aviation, and Other). The area of each rectangle is proportional to the number of distinct papers that used the associated dataset for experiments. Percentages show the ratio between the area of each rectangle and the whole area, which helps making comparisons between datasets and across fields.} 
    \label{fig:dataset_names}
\end{figure*}

Figure \ref{fig:dataset_names} represents the distribution of the datasets used across the papers we collected. Note that we only showed the datasets that appeared in more than three papers for readability purposes. Classical computer vision datasets such as MNIST/CIFAR/ImageNet are the most represented, probably because they are easy for humans to interpret and straightforward to train/test a model on, which make them perfect candidates for most experimentation. \textcolor{black}{Moreover, safety-critical aspects of vision-based systems (\eg{} self-driving cars for instance) are generally more explored by the community so far. In particular, MNIST (black and white digits image from 0 to 9) and  CIFAR (colored images of animals and vehicle, either 10 classes or 100 classes) are widely used because of their low resolution (28x28x1/32x32x1 and 32x32x3) which allow for not too time consuming experiments and inexpensive networks. They therefore constitute the \enquote{default} datasets to report a study. ImageNet dataset is in higher resolution (generally 256x256x3) with thousands of categories and millions of images, many with bounded boxes, making it a traditional benchmarking dataset for computer vision algorithms.} We also note the presence of many driving related datasets such as Cityscape, Traffic Signs etc. illustrating the interest of the research toward the development of systems for automotive. \textcolor{black}{The resolution of the image is generally the same as ImageNet and encompasses driving scenes in different city/conditions, many annotated and with bounding boxes. Cityscape also contains semantic segmantation as annotations.} In non-computer vision, ACAS-XU is the most represented, especially as it is used in formal verification techniques to assess effectiveness and performance of the technique as the system can process the information from the flight of an unmanned drone, which makes it interesting especially from a safety-critical point of view. \textcolor{black}{Strictly speaking, ACAS-XU is not a dataset like the previously mentioned ones, but a system for collision avoidance/detection and avoidance unmanned aerial systems. Yet, it is used as a tool to assess effectiveness of verification or control methods, as it provides concrete applications while still remaining simple enough to run experimentation in a relatively acceptable time. We note the lack of datasets related to other domains, such as Natural Language Processing (NLP), where ML is getting more and more used. This absence can probably be explained by the scarce safety-critical NLP application, which is the focus of this paper.}

\section{\textcolor{black}{Selected Taxonomy}}\label{sec:results:tax}
\textcolor{black}{As mentioned in the introduction, certification is defined as a \enquote{procedure by which a third-party gives written assurance that a product, process, or service conforms to specified requirements} \cite{Rodriguez99}. Applying this definition to the case of systems with ML components raises several issues. Specifically, ML introduces new procedures: data collection, pre-processing, model training etc. for which precise requirements are currently missing. Moreover, even in the presence of specific requirements, because the logic employed by ML models is learned instead of coded, assessing that said requirements are met is challenging.
For this reason, in this paper we consider that the question "How to certify ML based systems?" amounts to asking 1) What requirements should be specified for ML based systems? 2) How to meet these requirements? 3) what are the procedures a third-party should use to provide assurance that the ML systems conform to these requirements? Hence, pursuing certification for ML software systems refers to any research related to one of the three above questions subsumed by the question "How to certify ML based systems?"}

\textcolor{black}{Our mapping process (see \textbf{Section \ref{sec:methodology}}) allowed us to identify two main trends in the pursuit of certification for ML software systems; the first trend regroups techniques proposed to
solve challenges introduced by ML that currently hinder certification w.r.t traditional requirements and the second trend aims at developing certification processes based on traditional software considerations adapted to ML.
Similar observations were made by Delseny et al. in their work detailing current challenges faced by ML \cite{Delseny21}.}

\textcolor{black}{Our review of the first trend that addresses ML challenges w.r.t traditional software engineering requirements allowed us to identify the following categories:}
\begin{itemize}
    \item \textbf{Robustness}:
    \textcolor{black}{Models are usually obtained using finite datasets, and therefore can have unpredictable behaviors when being fed inputs that considerably differ from the training distribution. Those inputs can be even crafted to make the model fail on purpose (Adversarial Example). Robustness deals with methods that aim to tackle this \textit{distributional shift}, that is when the model is out of its normal operational condition.
    In terms of certification, robustness characterizes the resilience of the model.}
    \item \textbf{Uncertainty}: 
    \textcolor{black}{All ML models outputs do not necessarily consist in correctly calibrated statistical inference from the data and the model knowledge or confidence is hard to estimate and certify properly. In simple words, most current ML models fail to properly assess when \enquote{they do not know}}. Uncertainty deals with such ability of a model to acknowledge its own limitations and with techniques related to the calibration of the margin of error a model can make because of its specificity and available data. It also covers the concept of Out-of-distribution (OOD) which broadly refers to inputs that are, in some sense, too far from the training data.
    Uncertainty characterizes the capacity of the model of saying it does not recognize a given input, \textcolor{black}{by providing uncertainty measures in tandem with its predictions}, to allow a backup functionality to act instead of leading to an error for certification purposes. 
    \item \textbf{Explainability}:
    \textcolor{black}{Traceability and interpretability are crucial properties for the certification of software programs, not only because they allow to backtrack to the root cause of any error or wrong decision made by the program, but also because they help understanding the inner logic of a system. However, ML models are usually regarded as "black-box" because of the difficulty to provide high-level descriptions of the numerous computations carried out on data to yield an output}. Explainability covers all techniques that can shed some light on the decision a model makes in a human interpretable manner, that is to remove this traditional \enquote{black-box} property of some models.
    \item \textbf{Verification}: \textcolor{black}{Verifying the exact functionality of a software algorithm is part of the traditional certification. Nevertheless, because the burden of developing the process is passed from the developer to the model for ML, verifying the exact inner process is non trivial and quite complex}. Verification encompasses both Formal/Non-formal methods of verification, that is methods that aim to prove mathematically the safety of a model in an operation range, as well as all testing techniques that are more empirical and based on criteria of evaluations. This aspect highlights testing mechanisms that allow probing the system for eventual error for potential certification, by exploring potential unpredicted corner cases and to see how the model would react. 
    \item \textbf{Safe Reinforcement Learning}: By its very nature, RL differs from supervised/unsupervised learning and, as such, deserves to have its own dedicated category. This covers all techniques that are specifically tailored to improve the resilience and safety of a RL agent. This covers all the aspects of certifications when dealing with RL, from excluding hazardous actions and estimating uncertainty of decisions to theoretical guarantees for RL-based controllers.
\end{itemize}

\textcolor{black}{Table \ref{tab:comp_soft_ml} summarizes parallels we established between Software Engineering and ML w.r.t the challenges ML introduced.
We refer interested readers to the complementary material where these ML challenges are shown to impede the satisfaction of basic requirements on a simplified task example\app.}

\begin{tabularx}{\textwidth}{|>{\centering\arraybackslash}X|>{\centering\arraybackslash}X|>{\centering\arraybackslash}X|}
\caption{\textcolor{black}{Comparing traditional Software Engineering and Machine Learning on a simplified task of computing the sine function. The ML challenges are emphasized and contrasted with how they would typically be tackled in Software Engineering.}
    \label{tab:comp_soft_ml}}\\
        \hline
        & \textbf{Software Engineering} & \textbf{Machine Learning}  \\
        \toprule
        \hline
        \hline
        \textbf{Logic} & Coded & \enquote{Learned}\\
        \hline
        \textbf{System} &
        \begin{minipage}{3in}
        \begin{algorithmic}
        \If{$|x|>\pi$}
            \State $\texttt{Translate}(x)$
        \EndIf
        \State $y\gets x$
        \For{$k=1, 2, \ldots$}
            \State $n\gets 2k+1$
            \State $y \gets y+(-1)^{k} \frac{x^n}{n!}$
        \EndFor
        \State \Return $y$
        \end{algorithmic}
        \end{minipage}
        &
        \raisebox{-0.4\height}{
        \begin{tikzpicture}
        \tikzset{>=latex}
        \def\x{-1}
        \def\xx{1}
        \def\y{-1}
        \def\yy{1}
        \def\middle{\x/2+\xx/2}
        
        \draw[fill=black]  (\x,\yy) rectangle (\xx,\y);
        \node[color=white] at (\middle,\y/2+\yy/2) {ML};
        
        \draw (\xx/3,\yy) -- (\xx/3,\yy+0.15) -- (\x/3,\yy+0.15) -- (\x/3,\yy);
        \draw (\xx/3,\y) -- (\xx/3,\y-0.15) -- (\x/3,\y-0.15) -- (\x/3,\y);
        
        \node at (\middle,\yy+1) {$x$};
        \node at (\middle,\y-1) {$y$};
        \draw[<-] (\middle,\yy+0.25) -- (\middle,\yy+0.8);
        \draw[->] (\middle,\y-0.25) -- (\middle,\y-0.8);
        \end{tikzpicture}}
        \\
         \hline
         \hline
         \textbf{Explainability:} Interpret model's decision & Logs, Debug, Trace. & Lack of interpretability on inner mechanisms \\
         \hline
         \textbf{Uncertainty:} Consider model's margin of error & Theoretical bounds, Fail safe (default) mechanism, account for sensor uncertainty. & Aleatoric \& Epistemic uncertainty estimates as proxies of model error and OOD. \\
         \hline
         \textbf{Robustness:} Boost model's resilience in deployment & Clipping, property checking. & Distributional shift \& Adversarial Examples \\
         \hline
         \textbf{Verification:} Check model's correctness & Unittest, Symbolic, Coverage test & Lack of property formalization, high input dimensionality, complexity explosion\\
        \bottomrule
\end{tabularx}

\textcolor{black}{The second research trend tackling the lack of established process and specified ML requirements, discusses the possibility of developing a certification procedure specific to ML, and analogous to the ones from traditional software (mentioned in \textbf{Section \ref{sec:background}}). In order to account for this direction, we identify a last category:}\\
\begin{itemize}
    \item \textcolor{black}{\textbf{Direction Certification:} contrasts sharply with previous categories as it deals with higher-level considerations of obtaining a full certification process for ML models, specifically by taking into account all the challenges raised by the others categories. This category would cover drafted or partially/fully defined certification standards specific to ML, following similar standards defined in other fields.}
\end{itemize}

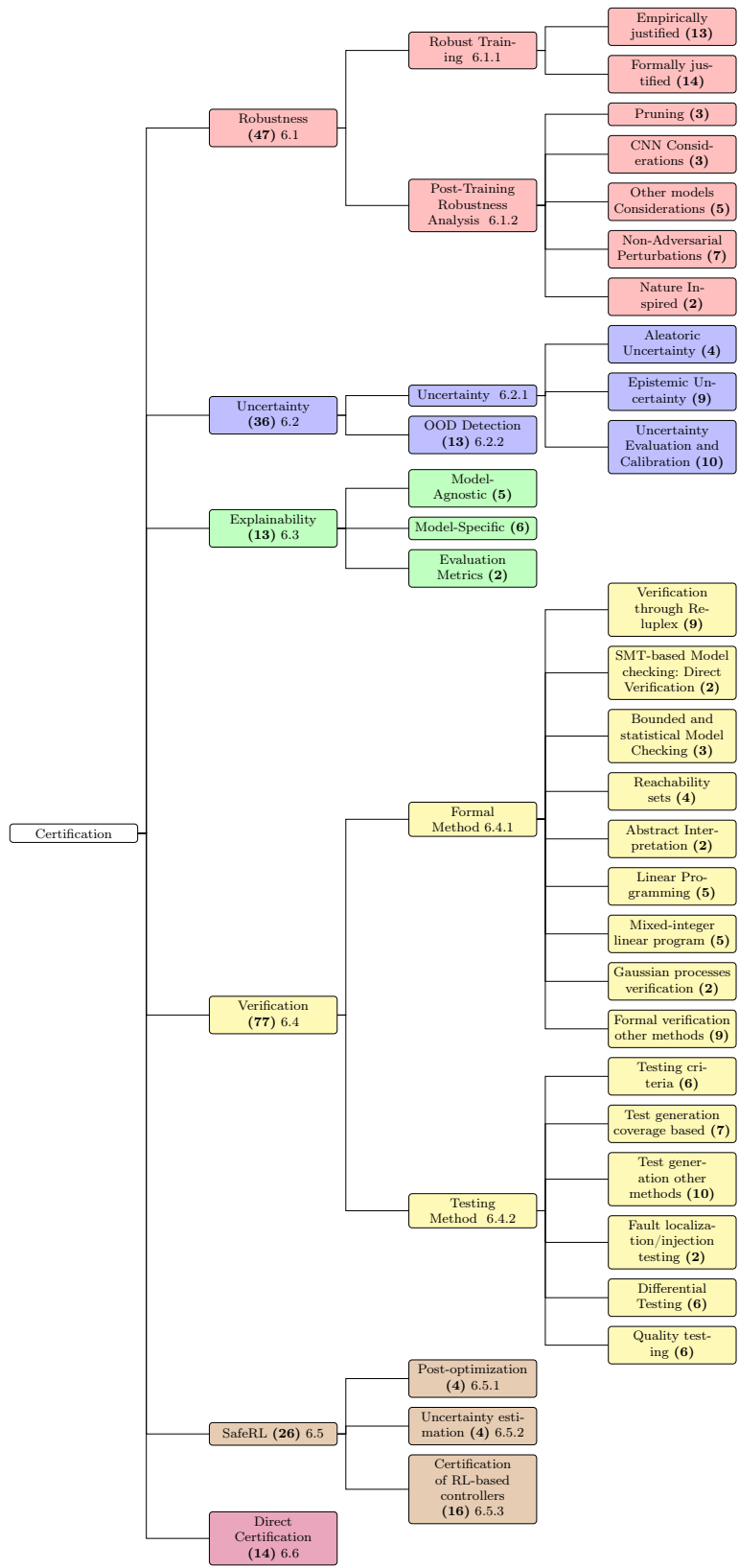
\begin{figure}
       \scalebox{0.65}{
       \begin{forest}
                for tree={
                    grow'=east,
                    forked edges,
                    draw,
                    rounded corners,
                    node options={
                        align=center },
                    text width=3cm,
                    anchor=west,
                    l sep=1.5cm,
                }
                [Certification, fill=white!25, parent,
                [\hyperlink{robustness}{Robustness \textbf{(47)}} \ref{robustness}, for tree={fill=red!25, child},
                [\hyperlink{robustness:robust_training}{Robust Training } \ref{robustness:robust_training}, for tree={fill=red!25, child},
                [\hyperlink{robustness:robust_training:empirical}{Empirically justified  \textbf{(13)}}]
                [\hyperlink{robustness:robust_training:theory}{Formally justified  \textbf{(14)}}]
                ]
                [\hyperlink{robustness:post_training}{Post-Training Robustness Analysis } \ref{robustness:post_training},  for tree={fill=red!25, child}
                [\hyperlink{robustness:post_training:pruning}{Pruning  \textbf{(3)}}]
                [\hyperlink{robustness:post_training:cnn}{CNN Considerations  \textbf{(3)}}]
                [\hyperlink{robustness:post_training:other_model}{Other models Considerations  \textbf{(5)}}]
                [\hyperlink{robustness:post_training:non_adv}{Non-Adversarial Perturbations  \textbf{(7)}}]
                [\hyperlink{robustness:post_training:nature}{Nature Inspired  \textbf{(2)}}]
                ]
                ]
                [\hyperlink{uncertainty}{Uncertainty  \textbf{(36)}} \ref{uncertainty},  for tree={fill=blue!25,child}
                [\hyperlink{uncertainty:uncertainty}{Uncertainty } \ref{uncertainty:uncertainty},  for tree={fill=blue!25,child}
                [\hyperlink{uncertainty:uncertainty:aleatoric}{Aleatoric \\ Uncertainty \textbf{(4)}}]
                [\hyperlink{uncertainty:uncertainty:epistemic}{Epistemic Uncertainty \textbf{(9)}}]
                [\hyperlink{uncertainty:uncertainty:eval}{Uncertainty Evaluation and Calibration \textbf{(10)}}]
                ]
                [\hyperlink{uncertainty:ood}{OOD Detection \textbf{(13)}} \ref{uncertainty:ood}]
                ]
                [\hyperlink{explainability}{Explainability \textbf{(13)}} \ref{explainability},  for tree={fill=green!25, child}
                [\hyperlink{explainability:model_agnostic}{Model-Agnostic \textbf{(5)}},  for tree={fill=green!25, child}
                [,phantom]
                ]
                [\hyperlink{explainability:model_specific}{Model-Specific \textbf{(6)}},  for tree={fill=green!25, child}
                [,phantom]
                ]
                [\hyperlink{explainability:metrics}{Evaluation Metrics \textbf{(2)}},  for tree={fill=green!25, child}
                [,phantom]
                ]
                ]
                [\hyperlink{verification}{Verification \textbf{(77)}} \ref{verification},  for tree={fill=yellow!35, child}
                [\hyperlink{verification:formal_methods}{Formal Method} \ref{verification:formal_methods},  for tree={fill=yellow!35, child}
                [\hyperlink{verification:formal_methods:reluplex}{Verification through Reluplex \textbf{(9)}}]
                [\hyperlink{verification:formal_methods:direct_verification}{SMT-based Model checking: Direct Verification \textbf{(2)}}]
                [\hyperlink{verification:formal_methods:bounded}{Bounded and statistical Model Checking \textbf{(3)}}]
                [\hyperlink{verification:formal_methods:sets}{Reachability sets \textbf{(4)}}]
                [\hyperlink{verification:formal_methods:abstract}{Abstract Interpretation \textbf{(2)}}]
                [\hyperlink{verification:formal_methods:lp}{Linear Programming \textbf{(5)}}]
                [\hyperlink{verification:formal_methods:milp}{Mixed-integer linear program \textbf{(5)}}]
                [\hyperlink{verification:formal_methods:GaussianP}{Gaussian processes verification \textbf{(2)}}]
                [\hyperlink{verification:formal_methods:other}{Formal verification other methods \textbf{(9)}}]
                ]
                [\hyperlink{verification:testing}{Testing Method } \ref{verification:testing},  for tree={fill=yellow!35, child}
                [\hyperlink{verification:testing:criteria}{Testing criteria \textbf{(6)}}]
                [\hyperlink{verification:testing:coverage_based}{Test generation coverage based \textbf{(7)}}]
                [\hyperlink{verification:testing:other_generation}{Test generation other methods \textbf{(10)}}]
                [\hyperlink{verification:testing:fault}{Fault localization/injection testing \textbf{(2)}}]
                [\hyperlink{verification:testing:differential}{Differential Testing \textbf{(6)}}]
                [\hyperlink{verification:testing:quality}{Quality testing \textbf{(6)}}]
                ]
                ]
                [\hyperlink{safe_rl}{SafeRL \textbf{(26)}} \ref{safe_rl},  for tree={fill=brown!40, child}
                [\hyperlink{safe_rl:post_opt}{Post-optimization \textbf{(4)}} \ref{safe_rl:post_opt},  for tree={fill=brown!40, child}
                [,phantom]
                ]
                [\hyperlink{safe_rl:uncertainty}{Uncertainty estimation \textbf{(4)}} \ref{safe_rl:uncertainty},  for tree={fill=brown!40, child}
                [,phantom]
                ]
                [\hyperlink{safe_rl:stability}{\textcolor{black}{Certification of RL-based controllers} \textbf{(16)}} \ref{safe_rl:stability}, for tree={fill=brown!40, child}
                [,phantom]
                ]
                ]
                [\hyperlink{direct_certif}{Direct \\ Certification \textbf{(14)}} \ref{direct_certif}, for tree={fill=purple!35, child},
                [,phantom]
                ]
                ]     
                ]
            \end{forest}} 
\caption{Overview of techniques discussed in the paper (clickable). \textcolor{black}{The numbers in bold designates the amount of papers in each part (some are presents in multiple sections and some are present only in the Complementary Material).}}
\label{fig:overview}
\end{figure}

\textbf{Section \ref{sec:review}} will elaborate on the different categories, and for each individual one, we shall present the state-of-the-art techniques gathered from our SLR, explain their core principles, discuss their advantages/disadvantages, and shed light on how they differ from one another. Figure \ref{fig:overview} provides an overview of how the main categories will be tackled. Afterward, \textbf{Section \ref{sec:discussion}} will develop on the global insight extracted from the SLR. We shall highlight the existing gaps that could serve as a starting point for future work in the research community.

\section{Review}\label{sec:review}
In this section, we summarize the state-of-the-art techniques that can be used for ML certification. Each of the categories is handled separately and hence do not need to be read in a specific order. \textcolor{black}{To keep the paper within manageable length, papers presenting method not fitting any of our current categories or details/remarks not necessary for the comprehension were moved to the Complementary Material of the paper. Parts for which more details can be found in the Complementary Material are marked with a \app.}

\hypertarget{robustness}{\subsection{Robustness}}\label{robustness}

\textcolor{black}{Generally speaking, robustness is the property of a system to
remain effective even outside its usual conditions of operation. In the specific context of ML, robust models are required to be resilient to unknown inputs. Although modern ML models have been able to achieve/surpass human performance on tasks such as image recognition, they are facing many robustness challenges hampering their integration in satefy-critical systems.} 
In most cases, robustness issues arise because of the \emph{distributional shift} problem, \ie{} when the training distribution the model was trained on is different from the deployment distribution. The most notorious examples of such phenomena are the \emph{adversarial attacks}, where carefully crafted perturbations can deceive a ML model \cite{goodfellow2014explaining}. Formally, given a sound input $x$, an adversarial example is defined as a crafted input $\Tilde{x}$ such as $\| x - \Tilde{x}\| \leq \delta$ and $f(x) \neq f(\Tilde{x})$ where $f(x)$ is the ML model prediction for a given input and $\delta$ is small enough to ensure that $\Tilde{x}$ is indistinguishable from $x$ from a human perspective. One of the first techniques used to generate Adversarial Examples is called the Fast-Gradient Sign Method (FGSM)  \cite{goodfellow2014explaining} and is defined as:
\begin{equation}
\Tilde{x} = x + \epsilon \times \text{sign}(\,\nabla_x \mathcal{L}(\theta, x, y)\,),
\label{eq:FGSM}
\end{equation}
where $\mathcal{L}$ is the loss function, $\theta$ are the weights of the model and $\epsilon$ is a small perturbation magnitude barely distinguishable from a human point of view, but strong enough to fool a model. This attack slightly perturbs the input following the gradient so as to increase the loss of the model. 
\textcolor{black}{Further improvements of FGSM include the Projected Gradient Descent (PGD) \cite{madry2017towards} and C\&W attacks \cite{Carlini17}.}
As such, this whole problem has raised much attention \cite{ren2020adversarial} as it shows the existence of countless inputs on which ML models make wrong predictions, but on which we would expect them to perform accurately. Adversarial examples are not the only examples of distributional shift; random perturbations/transformations, without any crafted behaviour, such as Gaussian noises or Out-of-distribution (OOD) examples (see \textbf{Section \ref{uncertainty:ood}}) can also lead to such a problem.

\hypertarget{robustness:robust_training}{\subsubsection{Robust Training}}\label{robustness:robust_training} 
\textcolor{black}{Robust training regroups techniques that introduce new objectives and/or processes during model training in order to increase robustness.}
When only considering adversarial perturbations of the input, the terminology becomes more specific:
adversarial training. Some of the recent techniques in the literature were found to be justified purely with empirical evidence while others were derived from formal robustness guarantees.

\hypertarget{robustness:robust_training:empirical}{\paragraph{Empirically Justified}}
The first formalization of adversarial training takes the form of a saddle point optimization problem \cite{madry2017towards}
\begin{equation}
\min_{\theta} \rho(\theta), ~\textrm{where}~~ \rho(\theta)=\mathbb{E}_{(x,y)\sim{D}} \bigg[ \max_{\delta\in{S}} \mathcal{L}(\theta, x+\delta,y) \bigg],
\label{eq:madry}
\end{equation}
where the inner maximisation problem searches for the most adversarial perturbation of a given data point $(x, y)$. The outer minimization problem attempts to fit the model such that it attributes a low loss on the most adversarial examples. The authors provide empirical evidence that the saddle point problem can be efficiently solved with gradient descent using the following procedure: generate adversarial examples (with FGSM or PGD) during the training phase and encourage the model to classify them correctly. Two recent modifications of this training procedure add a new term to the training objective that encourage adversarial and clean examples to have similar logits \cite{summers2019improved} or similar normalized activations per layer \cite{li2018learning}. Both modifications led to an increase of robustness to FGSM and PGD attacks.

An alternative robust training procedure uses a redundant Teacher-Student framework involving three networks: the static teacher, the static student, and the adaptive student \cite{bar2019robustness,bar2020robust}. The two students apply model distillation of the teacher by learning to predict its output, while having a considerably simpler architecture. Contrary to both static networks, the adaptive student is trained online and is encouraged to have different layer-wise features from the static student through an additional loss-term called the inverse feature mapper. The role of the adaptive student is to act as a watchdog to point out which NN is being attacked online through a threshold computation. Similar ideas are used in \cite{suri2018hardening}.

Other methods that increase robustness to attacks include employing an auto-encoder before feeding inputs to the model \cite{bakhti2019ddsa}, adding noisy layers
\cite{lecuyer2018connection,rakin2018parametric}, GAN-like  (Generative Adversarial Networks) procedures
\cite{duddu2019adversarial,liu2019affine}, adding an \enquote{abstain} option
\cite{laidlaw2019playing}, and using Abstract Interpretations,
\ie{} approximating an infinite set of behaviours with a finite representation \cite{mirman2018differentiable}.
Techniques for dealing with distributional shift over time, also known as concept shift, have also recently been developed for K-Nearest-Neighbors and Random Forest classifiers \cite{gopfert2018mitigating}.

\hypertarget{robustness:robust_training:theory}{\paragraph{Formally Justified\app}}\label{robustness:robust_training:theory}
Now, we describe training procedures conceived by first deriving a formal robustness guarantee, and modifying the standard training process so that the guarantee is met.
\textcolor{black}{A first theoretical guarantee for robustness of NNs relies on Lipschitz continuity. Formally, a network $f:\mathcal{X}\to \mathcal{Y}$ is $L$-Lipschitz (w.r.t. some choice of distances on $\mathcal{X}$ and $\mathcal{Y}$, in practice the $L_2$ euclidean distance is used) for some constant $L\geq 0$ if the following condition holds for all inputs $x$ and $z$ in $\mathcal{X}$:
\begin{equation}
    \Vert f(x) - f(z)\Vert \leq L \Vert x - z\Vert.
\label{eq:lipchitz}
\end{equation}
The smallest constant $L^\star$ is referred to as the Lipschitz constant of $f$. Intuitively, knowing that a network is Lipschitz allows to bound the distance between two outputs in terms of the distance between the corresponding inputs, thereof attaining guarantees of robustness to adversarial attacks.
The exact calculation of $L^\star$ is NP-hard for NNs and one is usually limited to up bounding it.}
Recently, tight bounds on the Lipschitz constant were obtained by solving a Semi-Definite Problem and including a Linear Matrix Inequality (LMI) constraint during training, allowing for a bound that can be enforced and minimized  \cite{pauli2020training}. 

\textcolor{black}{Alternative robustness guarantees provide lower bounds on the minimal $L_p$ perturbation required to change prediction of the model. Maximizing said bound leads to novel robust training procedures. For instance, \cite{croce2019provable,croce2019provable2} demonstrate a lower bound specific to NNs with ReLU activations, which are known to be piecewise linear over a set of polytopes. The theoretical bound is indirectly maximized by increasing, for each training point, the distance to the border of the polytope it lies in, as well as its signed distance to the decision boundary. As a complementary result, \cite{hein2017formal} provides a lower bound applicable to any NNs with differentiable activations, therefore excluding ReLUs. Increasing this lower bound, which is done by enlarging the difference between the logits of the predicted class and other classes as well as decreasing the difference between the logit gradients, is shown to increase the robustness of the network.}

When considering robustness to any distributional shift and not just adversarial examples, a common framework is Wasserstein Distributionally Robust Learning (WDRL)\textcolor{black}{, whose general principle is to consider the worst possible generalisation performance of the model under all data-generating distributions \enquote{close} to the original training distribution w.r.t the Wasserstein distance} \cite{sinha2017certifying}.
\textcolor{black}{In most practical settings, the formulation of WDRL is intractable and must be approximated/reformulated.} 
\textcolor{black}{For example, in the context of Model Predictive Control (MPC), WDRL is reformulated using a convexity-preserving approximation with minimal additional computation \cite{kandel2020safe}.
For general supervised learning applications, \cite{sinha2017certifying} relaxed WDRL into a Lagrangian form optimizable with stochastic gradient methods. More recently, it was proposed to modify the
Lagrangian form to consider differences in stability of the correlations between features and the target \cite{liu2020invariant}.}

\textcolor{black}{Alternative theoretically grounded robust training methods include
optimizing PAC-Bayes bounds for Gaussian Processes \cite{reeb2018learning} and employing robust Q-values for Reinforcement Learning \cite{everett2020certified}.}

\textcolor{black}{
Most of the theory of ML robustness discussed above studies the effects of distributional shift \ie{}, when the inputs observed in deployment differ from the training data. However, we also found papers that defined robustness in a manner specific to dynamic systems modeling \cite{richards2018lyapunov,revay2020convex,dean2020robust,varghese2020unsupervised}. Please see the supplementary materials for more details\app.}

\hypertarget{robustness:post_training}{\subsubsection{Post-Training Robustness Analysis}}\label{robustness:post_training}

This subsection encompasses considerations and analysis made post-training, in order to study extensively robustness behavior of the model and how it can be further improved.
 
\hypertarget{robustness:post_training:pruning}{\paragraph{Pruning methods}}\label{robustness:post_training:pruning}
Pruning techniques remove weights connections from pre-trained NNs, which was mainly used previously for reducing models sizes so they could fit on smaller systems. However, recent studies have surprisingly shown that prunning improves robustness. Pruning is generally done following certain metrics, generally the weights with the lowest magnitude are pruned, however \cite{Sehwag20} showed that its effectiveness against adversarial attacks is reduced and it is therefore proposed to use pruning based on importance score that are scaled proportionally to pre-trained weights to fasten computation. An optimization step can then be realized in order to further prune the model. They showed that the reduced model is almost as robust as the original model. A central point in pruning is the \enquote{Lottery Ticket Hypothesis}, which roughly states that any randomly initialized NN contains a subnetwork that can match the original network accuracy when trained in isolation. \cite{Consentino19} tested this hypothesis and confirms that the pruned model tends to be more robust to adversarial attacks while being faster to train. \cite{Yushuang19} is an example of pruning improving both robustness and compression.

\hypertarget{robustness:post_training:cnn}{\paragraph{Considerations in Convolutional NNs}}\label{robustness:post_training:cnn}
Convolutional Neural Networks (CNN) were studied extensively as they are the base for a lot of applications; \cite{Arnab2018} studied adversarial robustness in CNNs, showing that residual models such as ResNet are more resilient than chained model such as VGG. They also pointed out that multi-scale processing makes the model more robust. \cite{zhang2019neuron} observed that neurons of a CNN activated by normal examples follow the rule of \enquote{vital few and trivial many}. In other words, in a convolutional layer, only a few neurons (related to useful features) should be activated and many other ones (related to adversarial pertubations) should not. \cite{liu2018analyzing} also discovered the differences between clean and adversarial examples in terms of neuron activation and data execution paths. They introduced a tool to generate data path graphs, which visually shows how a pre-trained CNN model processes input data. Note that this technique, while focusing on robustness also pertains to explainability, a challenge which is discussed in \textbf{Section \ref{explainability}}.
 
\hypertarget{robustness:post_training:other_model}{\paragraph{Other models considerations}}
Aside from CNN, robustness considerations have been tackled in multiple other models, with for instance $k$-nearest neighbors \cite{Wang19-2} or Graph Neural Networks (GNN) \cite{Wei20}. Robustness also was studied from the point of view of ensemble learning (that is, training multiple models independently on the same datasets and averaging predictions to have a global one), that is known to be more resilient than a single model \cite{Grefenstette18}, as redundancy can provide extra security. \cite{Mani19-2} pushes the step further, by training each model of an ensemble to be resilient to a different adversarial attack by injecting a small subset of adversarial examples, which profit to the ensemble globally, even though it comes at the cost of training more models. As a last note, class distribution itself can also be a vector of robustness. Indeed, \cite{Pan19} showed that some classes can be more likely to flip to another when under adversarial perturbation. By analyzing the nearest neighbors map, it's possible to identify such classes and to increase the number of examples belonging to those to retrain the model. In a sense, it shows that class unbalance can also affect not only the model predictions, but also its robustness against adversarial examples.

\hypertarget{robustness:post_training:non_adv}{\paragraph{Non-adversarial Perturbations}}\label{robustness:post_training:non_adv}
Distributional shifts were also studied more empirically from the point of view of image transformations and noise injection. \cite{Hendrycks19} designed an ImageNet benchmark by adding corruption and/or perturbations in order to evaluate the resistance to corrupted images of different models. They notably show empirically that even small non-adversarial perturbations can lead a model to error. \cite{Arcaini20}\cite{Muller15} showed similar observations and indicate that using such images as data augmentation can boost robustness against adversarial examples. In fact, similar perturbations were shown to be useful in defending against adversarial examples \cite{Colangelo19}, by pre-processing potentially adversarial examples with those transformations or by using adversarial re-training \cite{Jeddi20}. In particular, \cite{Jeddi20} used noise injection technique to strengthen model's robustness; the network is updated in the presence of feature perturbation injection to improve adversarial robustness while the parameters of the perturbation injection modules are updated to strengthen perturbation capabilities against the improved network. The idea behind this is that, when both the network parameters and the perturbed data are optimized, it is harder to craft successful adversarial attacks and it seems empirically to be more effective than some adversarial training methods such as \cite{rakin2018parametric}\cite{lecuyer2018connection}.
 
\hypertarget{robustness:post_training:nature}{\paragraph{Nature inspired techniques}}
Although these techniques also apply to the training stage, they differ from the aforementioned ones in that they tackle the adversarial attacks from a completely different angle. \cite{dapello2020simulating} observed that CNN models with hidden layers are somehow closer to the primate primary visual cortex (V1) and are more robust to adversarial attacks. This inspiration leads to a novel CNN architecture where the early layers simulate primate V1 (the VOneBlock), followed by a NN back-end adapted from the existing CNN models (\eg{} ResNet). The paper showed a number of interesting findings\footnote{It is worth noting, the method can outperform other defense based on adversarial training such as \cite{rusak2020increasing} ($L_\infty$ constraint and adversarial noise with Stylized ImageNet training) when considering a wide range of attack constraints and common image corruptions.} such as evidence that the new architecture can improve the robustness against white-box attacks. Similarly \cite{Ye19}, inspired by the association and attention mechanisms of the human brain, introduced a \enquote{caliber} module on the side of a NN to replicate such a mechanism. Traditionally, we assume the training data distribution and operation data are independently and identically distributed which is not always the case in practice (\eg{} image with sun in training vs rainy day in operation), which leads to retraining and/or fine-tuning in order to account for that, which can potentially be very expensive. They instead propose to retrain only the light weight caliber module on those new data, to help the model \enquote{understand} those.
 
\begin{tcolorbox}[colback=blue!5,colframe=blue!40!black]
\begin{itemize}
    \item Robustness deals with resilience of the model against potential unexpected or corner cases examples, which represent potential shortcomings a system needs to deal with in a safety-critical scenario, as such robustness is part of the ML certification process. 
    \item Robust (Adversarial) training encapsulates all methods that modify the training procedure of a model in order to increase its robustness. \textcolor{black}{(Section \ref{robustness:robust_training})}
    \item There exists a wide range of robust training procedures that are based on formal theoretical guarantees, which should be extensively investigated. \textcolor{black}{(Section \ref{robustness:robust_training:theory})}
    \item Post-training \textcolor{black}{(Section \ref{robustness:post_training})} empirical observations or model considerations can also serve as a base for Robustness improvement, with for instance analysis of non-adversarial perturbations \textcolor{black}{(Section \ref{robustness:post_training:non_adv})} or interactions of neurons inside of a model \textcolor{black}{(Section \ref{robustness:post_training:pruning})}.
    \item However, a more thorough understanding of how such observations can indeed strengthen a model's robustness is needed for those methods to be effectively applied to certification of safety-critical systems \textcolor{black}{(Section \ref{robustness:post_training:cnn})}.
\end{itemize}
\end{tcolorbox}
 
\hypertarget{uncertainty}{\subsection{Uncertainty estimation and OOD detection}}\label{uncertainty}
The concepts of uncertainty and OOD detection are closely related. In fact, the former is often used as a proxy for the latter. For this reason, this section starts by delving deeply into uncertainty quantification, before discussing the topic of OOD detection.
 
\hypertarget{uncertainty:uncertainty}{\subsubsection{Uncertainty}}\label{uncertainty:uncertainty}
In ML, uncertainty generally refers to the lack of knowledge about a given state, and can be categorised as either aleatoric or
epistemic \cite{gruber2018uncertainties}. On the one hand, aleatoric uncertainty measures the stochasticity that is inherent to the data, and can be induced by noisy sensors or a lack of meaningful features. On the other hand, epistemic uncertainty refers to the under-specification of the model given the finite amount of data used in the ML pipeline. This uncertainty is assumed to be reducible as more and more data is available, while aleatoric uncertainty is irreducible.
 
\hypertarget{uncertainty:uncertainty:aleatoric}{\paragraph{Aleatoric Uncertainty}}
 
The aleatoric uncertainty quantifies the amount of noise in the data and is 
formalized by treating both inputs $x$ and output $y$ as random variables which follow
a joint probability distribution $(x, y)\sim D$. The stochasticity of the 
distribution $D$ is inherent to the task at hand and cannot be reduced by 
considering larger datasets.
 
In a regression setting, the most common technique to estimate the aleatoric 
uncertainty of the data is to fit a neural-network using the loss attenuation objective \cite{gruber2018uncertainties}. The main idea behind this objective
function is to assume a Gaussian conditional distribution of
$y$ given the input $x$ \ie{} 
\begin{equation}
    y\,|\,x\sim \mathcal{N}(f_\theta(x),
\sigma^2_\theta(x)),
\label{eq:conditionnal_y_x}
\end{equation}
 
where $f_\theta(x)$ and $\sigma^2_\theta(x)$ are both outputs
of the network, and represent the prediction, and the aleatoric uncertainty
respectively. The choice of Gaussian conditionals is often made for computational convenience but can also be justified by the central limit theorem.
Now, the loss attenuation objective is defined as the logarithm of the conditional distribution, averaged across the training set
\begin{equation}
    \mathcal{L}(\theta) = \mathbb{E}_{(x, y) \sim D}\bigg[
    \frac{(y - f_\theta(x))^2}{2\sigma_\theta^2(x)} + \log \sigma_\theta(x)\bigg].
\end{equation}
 
Although this loss function is derived from a regression setting, we have observed some efforts to adapt it to other tasks such as classification \cite{le2018uncertainty}, and space embeddings for visual retrieval systems \cite{taha2019unsupervised}.
 
Loss attenuation has the drawback of requiring a modification
of the training objective as well as the network architecture since the network
must output both the mean and variance of the conditional distribution.
A model-agnostic alternative 
is to propagate the sensor noise through-out the network 
using a technique called Assumed Density Filtering (ADF) \cite{segu2019general}.

\hypertarget{uncertainty:uncertainty:epistemic}{\paragraph{Epistemic Uncertainty}}
Epistemic uncertainty encapsulates our lack of knowledge about the
task induced by the finite amount of data.
Because of this lack of data, models are under-specified, meaning that
there exists a wide diversity of models that perform equivalently well 
on any given
task. The quintessential technique used to estimate this type of uncertainty
in deep learning are Bayesian Neural Networks (BNN), \textcolor{black}{that employ a probability distribution over good parameters, called the
posterior distribution, rather than a single set of optimal parameters.
For non-linear neural networks, the exact posterior distribution must be approximated, often by applying dropout at train and test time, a scheme called
MC Dropout \cite{gruber2018uncertainties,sheikholeslami2020minimum} which has successfully been applied to automonous driving \cite{lee2019ensemble}, robot control \cite{toubeh2019risk}, and health pronostics \cite{peng2019bayesian}.
To compute the epistemic uncertainty at a specific input $x$, MC Dropout measures the variance of the model's predictions from different forward passes with random shutdown of neural activations.}
The bottleneck of this approach is the requirement of several forward-passes to estimate the epistemic uncertainty at any input. However, some efforts are being done to estimate the variance of the network output with a single forward-pass, using first-degree Taylor expansions \cite{postels2019sampling}.
 
\textcolor{black}{Alternatives to MCDropout either suggest to approximate the posterior by collecting parameter samples using Stochastic Gradient Descent with
noise injection when reaching a local optima \cite{park2018sampling}, to use 
Deep Gaussian Processes as a non-parametric alternative to
BNNs \cite{jain2020decision}, or to employ
the so-called \textit{Deep Evidential Regression}
\cite{amini2019deep} which allows for simultaneous estimation of aleatoric and epistemic uncertainties with a single forward pass.}
 

\hypertarget{uncertainty:uncertainty:eval}{\paragraph{Uncertainty Evaluation and Calibration\app}}
\textcolor{black}{A major challenge in uncertainty quantification is that, given our ignorance of the true data-generating distributions $D$, there are no ground-truth values to which uncertainty estimates can be compared. Nonetheless, the quality of uncertainty estimates can still be assessed based on their ability to increase model safety at prediction-time.
Indeed, uncertainties are generally used as measures of confidence that the model has in any given prediction.} In safety critical 
systems, confidence measures are useful as accurate proxies for 
prediction error, \ie{} the confidence should be high for correct classifications and low for incorrect ones. This is such an important property that in some work, uncertainty is directly defined in terms of the ability to predict erroneous decisions \cite{klas2019uncertainty}. \textcolor{black}{Proposed uncertainty evaluations that go along these lines are the Remaining Error Rate (RER) and the Remaining Accuracy Rate (RAR), namely the ratio of all confident miss-classifications to all samples and confident correct-classifications to all samples respectively. A curve of the two quantities for different confidence thresholds allows comparisons of uncertainties estimates \cite{henne2020benchmarking}.} More specific uncertainty evaluations can be employed depending on the task. Notably, in autonomous driving control, uncertainty can be evaluated as a proxy of the risk of crashing in the next $n$ seconds \cite{michelmore2018evaluating}.

An alternative way to assess the quality of uncertainty estimates in active learning settings, where the model is evolving over time, is to measure how  much uncertainty helps guide exploration of the model
\cite{turchetta2016safe,fisac2018general,Fan20,zhan2017safe}\app.

Moreover, in safety-critical systems, it is crucial to ensure that uncertainty estimations are calibrated. For example, in the context of classification, uncertainties
are usually within the $[0, 1]$ interval which makes it tempting to interpret them as probabilities. However, these \enquote{probabilities} do not necessarily have a frequentist interpretation, \ie{} it is not because a system makes 100 predictions each with 0.9 certainty that about 90 of those predictions will indeed be correct. When the uncertainties of classifications are close to their frequency of errors, the uncertainty method is said to be well-calibrated.
 
Common techniques for calibration such as isotonic regression and temperature scaling are post-hoc meaning they can modify any uncertainty estimation after the model is trained. Said approaches however require a held-out dataset called the calibration set in order to be applied. It is also possible to calibrate uncertainties by modifying the training objective so it directly takes calibration into account \cite{feng2019can}.

\textcolor{black}{
The notion of calibration is far less intuitive in the context of regression, seeing that multiple definitions have been advanced
\cite{feng2019can, levi2019evaluating}\app. Nonetheless, calibration of regressors is crucial in multiple applications, especially object detectors where bounding boxes are obtained via regression
\cite{feng2019can,levi2019evaluating,kuppers2020multivariate}.}


\hypertarget{uncertainty:ood}{\subsubsection{OOD detection}}\label{uncertainty:ood}
\textit{Out-of-distribution} (OOD) refers to input values that differ drastically from the data used in the ML-pipeline for both training and validation. Unsafe predictions on these OOD inputs can easily occur when a model is put \enquote{in-the-wild}. This unwanted behavior of a ML model can be explained on the one hand, by the lack of control on the model performance on data that differ from the training/testing data, \ie{} OOD data; and on the other hand, by the difficulty to specify which inputs are safe for the trained model, \ie{} in-distribution data. Traditionally, OOD instances are obtained by sampling data from a dataset that differs from the training one. In the context of certification for safety-critical systems, it is primordial to either provide theoretical guarantees on how models would perform in production (robustness), or develop techniques to detect OOD inputs, so that the prediction made by the model can be safely ignored or replaced by decisions made by a human in-the-loop or a more standard program. This subsection focuses on various techniques used to detect OOD inputs.

For classification, the simplest approach to this detection task is to rely on the maximum softmax probability of the last layer of the network \cite{Hendrycks18} as a proxy for the uncertainty on the prediction. If this method sometimes succeeds in detecting OOD datasets sufficiently different from the in-distribution, it's not necessarily the case for more closely related datasets. Moreover, softmax was shown not to be a good measure of model confidence \cite{Gal16,Hendrycks18}, as it is known to yield overly-confident predictions on some OOD samples. In fact, this has recently been mathematically proven for ReLU networks \cite{Hein19}. Therefore, recent methods attempt to find other proxies for OOD detection; GLOD\footnote{Author's remark: A new improvement of GLOD, FOOD, was released earlier this year. Following our methodology, we kept only GLOD reference our methodology extracted, but we invite readers to check the new instalment of the method: \url{https://arxiv.org/abs/2008.06856}} \cite{Amit20} replaced this softmax layer with a Gaussian likelihood layer to fit multivariate Gaussians on hidden representations of the inputs. \cite{Hendrycks20} tackled the problem of OOD detection for multi-class by using the negative of the max unnormalized logit, since traditional maximum softmax OOD detection techniques fail as the probability mass is spread among several classes. \cite{Wang18} used a generative model for each class of a dataset. This way, they can use those models at test time to see from which instance an input is closest, and compare it to a threshold tuned on in-distribution data.

Other methods were proposed to tackle this issue, the two most well-known being ODIN \cite{Liang20} and Mahalanobis Distance \cite{Lee18}. One thing those two methods have in common is that they both pre-process the inputs by adding some gradient-based perturbation. The idea of using input perturbations to better distinguish OOD from in-distribution have been echoed in many methods; \cite{Hendrycks18} used noises as a way to train a classifier to distinguish noisy or not images based on a confidence score. \cite{Ren19} used them to disentangle what they call background (population statistics) and semantic (in distribution patterns) components which are two parts of the traditional likelihood\footnote{They argue that the background part is why OOD can be misinterpreted. Indeed, they observed both that several OOD can have similar background components as in-distribution data and that the background term can dominate the semantic term in the likelihood computation. By adding noise, they essentially mask the semantic term so they can train a model specifically on background components. This could explain why models such as PixelCNN can fail on OOD detection.}. An alternative to using proxies for OOD detections is to use a purely data-modeling approach such as the one in \cite{meinke2019towards}, where the full joint distribution of $x$ and $y$ is estimated with Gaussian Mixture modelings of in-distribution and out-of-distribution images. It is also possible to detect OOD by partitioning the input space with a decision tree, identify the leaves with low training point density, and reject any new instance that lands in those leaves \cite{gu2019towards}. \textcolor{black}{Other examples of OOD detection methods can be found in \cite{Gschossmann19} and \cite{Lust20}.}
 
OOD detectors must be evaluated on their ability to reject OOD inputs without rejecting too many in-distribution inputs, which would make the ML system too passive. \cite{Henriksson19} propose a set of evaluations for OOD detection for such problem. Moreover, such detectors can also show lower performance on Adversarial Examples, or even on Adversarial OOD (\ie{} OOD to which we applied adversarial perturbations) \cite{Sehwag19}. However, little work has been made to verify methods in real safety-critical settings, which are the settings in which OOD detection is the most pertinent and important. It's crucial that such methods generalize so as to account for all potential OOD samples the system might come across; in general, authors do not seem to consider that OOD samples follow a distribution (except for \cite{meinke2019towards}), but instead seem to use this term informally to refer to \enquote{any sample that is different enough from the training set or distribution}. In this context, the performance of a method depends heavily on the particular choice of OOD samples used for the evaluation and these scores must therefore be considered with caution. Moreover this performance comes with no theoretical guarantees, especially when it comes to the frequency of False Negatives, \ie{} OOD samples which are detected as safe for prediction, which is critical for certification. 

\begin{tcolorbox}[colback=blue!5,colframe=blue!40!black]
\begin{itemize}
    \item Uncertainty analysis is part of the ML certification process, as it deals with the ability of a model to detect unknown situations which are out of its normal range of operation, in order to be able to deal with it accordingly \textcolor{black}{(Section \ref{uncertainty:uncertainty})}.
    \item There is no universal metric to evaluate uncertainty so different uncertainty estimators must be compared on their usefulness in increasing the safety of ML systems by acting as proxies for prediction error, and by having frequentist interpretations on in-distribution data \eg{} being calibrated \textcolor{black}{(Section \ref{uncertainty:uncertainty})}.
    \item OOD detection is generally based on finding a mean to disentangle in from out distribution data, then applying a threshold cut that is correctly tuned. Although the state-of-the-art provides good results on traditional datasets/models, getting a better understanding of how NNs inner representations differ from in/out of distribution is necessary to improve on existing methods \textcolor{black}{(Section \ref{uncertainty:ood})}.
    \item Moreover, most techniques, while being pretty efficient, are evaluated on a limited number of OOD datasets. As such, there is at best only strong empirical evidence a technique can work and no guarantee it can generalize to every possible OOD the model might come across, which is not sufficient for safety-critical application. A better understanding of what constitutes all of OOD for a model (and a given task) could lead to formal guarantees such as it is the case for Adversarial Examples \textcolor{black}{(Section \ref{uncertainty:ood})}.
\end{itemize}
\end{tcolorbox}

\hypertarget{explainability}{\subsection{Explainability}}\label{explainability}

With the steady increase in complexity of ML models and their 
wide-spread use, a growing concern has emerged on the interpretability of their
decisions. Such apprehensions are especially present in
contexts where models have a direct impact on human
beings. To this end, the European Union has adopted in 2016 a 
set of regulations on the application of ML models, notably the
\enquote{right to explanation}, which forces any model
that directly impact humans to provide meaningful information about 
the logic behind their decisions 
\cite{goodman2017european}. Formally, given a model $f$, a data
point $x$, and a prediction $f(x)$, it is becoming increasingly
important to provide information about \enquote{why} or \enquote{how} the
decision $f(x)$ was made. These recent constraints have
led to the quick development of the field of
eXplainable Artificial Intelligence (XAI), a subfield
of AI interested in making ML models
more interpretable while keeping the same level
of generalisation performance \cite{arrieta2020explainable}. The fundamental objects studied in XAI are \enquote{explanations} \textcolor{black}{of several forms:
\begin{enumerate}
    \item \textbf{Textual Explanation:} Simplified text descriptions of the model behavior.
    \item \textbf{Visual Explanation:} Graphical visualisations of the model.
    \item \textbf{Feature Importance: } Each feature of the input vector $x$ is ranked with respect to its relevance toward the decision $f(x)$.
    \item \textbf{Counterfactual Examples:} Given $x$ and a decision $f(x)$, a counterfactual example $\widetilde{x}$ represents an actionable perturbation of $x$ that changes the decision $f(x)$. By actionable, we mean that the perturbations are done with respect to features an individual can act upon. 
\end{enumerate}
The main paradigms to compute these explanations are: the training of inherently interpretable models, and the use of post-hoc explanation, \ie{},  add-on techniques that allow to interpret any complex model after it has already been trained.
Our observation is that post-hoc explanations are considerably more popular in the literature than the development of interpretable models. For this reason, the former is now discussed in detail while the latter is elaborated on in the supplementary material\app.}
 
\hypertarget{explainability:model_agnostic}{\paragraph{Model-Agnostic}} 
These post-hoc explanation techniques provide information about the decision-making 
process \textcolor{black}{of any model}. In this context, the explainer is only able to query the black box $f$ at arbitrary input points $z$ in order to
provide an explanation. The most common model agnostic post-hoc explainers
are local surrogates, which are interpretable models trained to locally mimic the black box
around the instance $x$ at which one wishes to explain the decision $f(x)$. 
Formally, a sampling distribution $N_x$ is chosen to represent a neighborhood around the instance $x$ and the local surrogate $g_x$ is taught to mimic $f$ around $x$ by minimizing the neighborhood infidelity
\begin{equation}
g_x = \min_{g} \mathbb{E}_{z\sim N_x} \,\big[ (\,f(z) - g(z)\,)^2 \big].
\label{eq:local_fidelity}
\end{equation}
 
The surrogate model $g_x$ being interpretable, it can be used to provide a post-hoc
explanation of $f(x)$ in the form feature importance, textual explanation, or counterfactual examples. The most fundamental explainer of this type is called
LIME (Local Interpretable Model-agnostic Explanations), and fits a sparse linear model $g$ on $f$ in a locality $N_x$ that is specific to either tabular, textual, or image data \cite{ribeiro2016should}. The weights of the linear model are then used as a measure of feature importance. 
\textcolor{black}{Alternatively, LORE (LOcal
Rule-based Explanations) uses decision trees as surrogate
models $g$ and a genetic algorithm for the sampling
distribution \cite{guidotti2019factual,pedreschi2018open}. Decision trees are an interesting choice for local surrogates because they can provide textual descriptions (by writing the conjunction of all Boolean statements in the path from the tree's root to its leaf), and
counterfactual examples (by searching for leafs with similar paths but different predictions). Finally, DeepVID (Deep learning approach to Visually Interpret and Diagnose) \cite{wang2019deepvid} locally fits a linear regression $g$
using a Variational Auto-Encoder to sample points $z$ 
in the vicinity of $x$, while LEMNA (Local Explanation Method
using Nonlinear Approximation) fits a mixture regression and samples S
from $N_x$ by randomly nullifying components of the input vector $x$ 
\cite{guo2018lemna}.}
 
Alternative model-agnostic post-hoc explanations are diagnostic curves
such as Partial Dependency Plots (PDP) and Adaptive Dependency Plots (ADP)\footnote{Author's remark: The original ADP paper pushed on arXiv in 2019 has been improved and re-uploaded in 2020. In this review, we kept the 2019 reference, which was recovered by our methodology, but we invite reader to check the 2020 paper: \url{https://arxiv.org/abs/1912.01108}}
\cite{inouye2019diagnostic}, 
which aim
at visualising the black-box behavior by computing line charts along specific directions in input space.

\hypertarget{explainability:model_specific}{\paragraph{Model-Specific}}
Such post-hoc explanations are restricted to a specific model type, \ie{} tree-based models, NNs, etc.
\textcolor{black}{For image classification with CNNs, common explanations rely on saliency maps, \ie{} images 
emphasizing the pixels of the original image $x$ that contributed the most toward the prediction $f(x)$ (feature importance). In
\cite{wagner2019interpretable}, saliency maps were obtained by applying a binary mask 
over the original image.
The masks were computed via a so-called \textit{preservation game} where the
smallest amount of pixel is retained while keeping a high class-probability, or a \textit{deletion game} where the class-probability
is considerably reduced by masking off the least amount of pixels.
The optimization procedure used to compute the masks being very similar to adversarial attacks (see \textbf{Section \ref{robustness}}), \textcolor{black}{there is concern the saliency maps may contain artifacts that do not highlight meaningful features from the image.} To address this, the authors introduce clipping layers 
to ensure only a subset of the high level features identified on the original image (edges, corners, textures, etc.)} are used when optimizing the masks. This defense has the added beneficial effect of providing fine-grained saliency maps.
In \cite{nowak2019improve}, saliency maps were computed for charging post detectors using the specific Faster R-CNN network architecture. Their post-hoc method extracts information from a submodule of the Faster R-CNN network architecture called the Region Proposal Network.
For tabular data, \cite{amarasinghe2019explaining} combine a model-specific
post-hoc explanation called \textit{Layer-wise Relevance Propagation} and fuzzy set theory in order to provide textual explanations of a Multi-Layered Perceptron.
 
Other researchers have attempted to provide visual explanations of NNs work directly at the neuron-level by looking at neural activation patterns in supervised learning \cite{meyes2020under} and reinforcement learning \cite{meyes2020you}, as well as the high level features learned in deeper layers of a CNN \cite{kuwajima2019improving}.
 
\hypertarget{explainability:metrics}{\paragraph{Evaluation Metrics}}\label{explainability:metrics}
One major difficulty in evaluating and comparing post-hoc explanations is the lack of ground-truth on what is the true explanation of a decision made by a black box.
The amount by which an explanation is truly representative of the decision process of a model is called its \enquote{faithfulness}. When computing model-agnostic explanation extracted with local surrogate models, \ie{} simple models that attempt to mimic the complex model near the point $x$ of interest, faithfulness can be studied in terms of the error between the surrogate and the model, see Equation \ref{eq:local_fidelity}. Large errors suggest the local surrogates are unable to properly mimic the complex model and their explanations can safely be discarded. However, when several local surrogate models mimic the model well, it is hard to state which one provides the most meaningful explanations. \textcolor{black}{This is because the loss in Equation \ref{eq:local_fidelity} depends on a choice of neighborhood distribution $N_x$ that is different for every method}. 

\textcolor{black}{It was proposed to measure the faithfulness of CNN saliency maps by iteratively zeroing-out pixels in order of importance and reporting the corresponding decrease in class-probability output \cite{wagner2019interpretable}. The AUC of the resulting curve is a measure of unfaithfulness.}

A promising alternative to compare/evaluate explanations is to assess their usefulness in debugging the model and in the ML pipeline in general. For example, in \cite{nowak2019improve}, the saliency maps of an electric charger detector were used to understand where the model was putting its attention when it failed to detect the right objects, which helped engineering specific data augmentations. Notably, the network was found to put a lot of its attention on pedestrians in the background, which was tackled by introducing negative examples with only pedestrians and no electric charger.
 
\begin{tcolorbox}[colback=blue!5,colframe=blue!40!black]
\begin{itemize}
    \item Explainability is part of the ML certification process, as it deals with interpretability and traceability of a model, which is important in traditional settings, but even more so in ML because of its \enquote{black-box} nature.
    \item Explanations can take several forms : feature importance \cite{ribeiro2016should,wagner2019interpretable}, counterfactual examples \cite{guidotti2019factual,pedreschi2018open}, textual description \cite{amarasinghe2019explaining}, and model visualisation \cite{inouye2019diagnostic,meyes2020under}
    \item The two main paradigms of eXplainable Artificial Intelligence are the training of intrinsically interpretable yet accurate models, and the use of post-hoc explanations which aim at explaining any complex model after it is trained.
    \item There is currently no universal metric to assess whether or not a post-hoc explanation has truly captured the \enquote{reasons} behind specific decisions made by a black box. We are indeed very far from tackling the EU regulations on the \enquote{right to an explanation}. Therefore, we think that a more realistic short-term goal is to evaluate explanations on their usefulness in designing the ML pipeline and in debugging complex models, similar to the experiments in \cite{nowak2019improve}.
    \textcolor{black}{(Section \ref{explainability:metrics})}
\end{itemize}
\end{tcolorbox}
 
\hypertarget{verification}{\subsection{Verification}}\label{verification}
 
Verification is any type of mechanism that allows to check formally or not that a given model respects a certain set of specifications. As such, it differs from method seen in the \textbf{Section \ref{robustness}} in the sense that the methods are not used to improve a model resilience, but rather to evaluate a certain number of properties (which can be linked to robustness for instance) \textit{after} the model was refined and/or trained. Hence, it does not act on the model but rather aims at verifying it. Most of the approaches are based on test input generative approaches, that is techniques that aim to generate failure inducing test samples. The idea is that, through careful selection of such samples, it's possible to cover \enquote{corner-cases} of a dataset/model which are examples for which the model was not trained on or does not generalize well enough to the point it would fail to correctly predict on such inputs. In a sense, such examples can be considered a part of the distributional shift problem. We distinguish two main approaches: one that is based on formal methods, that is a complete exploration of the model space given a certain number of properties to check, and one that is based on empirically guided testing, using criteria or relations that can lead to error inducing inputs.
 
\hypertarget{verification:formal_methods}{\subsubsection{Formal Method}}\label{verification:formal_methods}
 
Formal methods provide guarantees on a model given some specifications through mathematical verification. The increasing use of ML on safety-critical applications has led researchers to derive formal guarantees on the safety of those applications. \textcolor{black}{In the following, we present per category some work that tackles the safety of machine learning based critical applications in terms of providing formal guarantees.}
 
 
\hypertarget{verification:formal_methods:reluplex}{\paragraph{Verification through Reluplex}}\label{verification:formal_methods:reluplex} 
 
\cite{katz2017reluplex} proposed Reluplex, an SMT (Satisfiability modulo theory)-based framework that verifies robustness of NNs  with RELU activations using a Simplex algorithm. In its most general form, Reluplex can verify logical statements of the form
\begin{equation}
    \exists\, x\in \mathcal{X} \,\,\text{ such that } \,\,f(x) \in \mathcal{Y},
\end{equation}
where $\mathcal{X}$ and $\mathcal{Y}$ are convex polytopes, \eg{} sets which can be expressed as conjunctions of linear inequalities. Reluplex can either assert that the formula holds (\texttt{SAT}) and yield the specific value of $x$ that satisfies it, or it can confirm that no such point $x$ exists (\texttt{UNSAT}). \textcolor{black}{Reluplex outperforms existing framework in the literature but is restricted to ReLU activations. As well as only verify robustness with respect to the $L_1$ and $L_{\infty}$ norms (because their open balls are polytopes)}

A framework similar to Reluplex is Reluval \cite{wang2018formal}, a system for formally checking security properties of Relu-based DNNs. ReluVal can verify a security property that Reluplex deemed inconclusive. But to improve its robustness verification, \cite{ren2019using} upgraded ReluVal by adding Quantifier Estimation to compute the range of activations of each neuron. Their method benefits from parallelization as neurons from one layer share the same inputs. Their experiments show they can verify robustness in small neighbourhoods. Using overapproximation and/or longer calculation time they can achieve this for bigger neighbourhoods. \cite{julian2019guaranteeing} present an approach for reachability analysis of DNN-based aircraft collision avoidance systems by employing Reluplex and Reluval verification tools to over-approximate DNNs. By bounding the network outputs, the reachability of near midair collisions (NMACs) is investigated.  Looseness of dynamic bounds, sensor errors and pilot delay were investigated, in the experiments addressing all issues with real-time costs. However, the approach needs to be tested on real aircraft collision avoidance systems, to better assess its effectiveness.
 
\hypertarget{verification:formal_methods:direct_verification}{\paragraph{SMT-based Model checking: Direct Verification}}\label{verification:formal_methods:direct_verification}
 
\textcolor{black}{SMT(Satisfiability Modulo Theories)-based verification process is applied on image classifiers to detect adversarial attacks in \cite{huang2017safety}. The process consists on discretizing the region around the input to search for adversarial examples in a finite grid. It proceeds by analyzing layers one by one. The results of the verification are either the NN is safe with respect to a manipulation or the NN can be falsified. The software Z3 is used to implement the verification algorithm} The approach gives promising results but suffers from its complexity, and the verification is exponential in the number of features. 
\cite{naseer2020fannet} propose FANNET, a formal analysis framework that uses model checking to evaluate the noise tolerance of a trained NN. During the verification process, in case of non-satisfiability of the input property, a counterexample is generated. The latter can serve to improve training parameters and workaround the sensitivity of individual nodes. The experiments successfully show the effectiveness of the approach, which is however limited to fully connected feed-forward NNs.

\hypertarget{verification:formal_methods:bounded}{\paragraph{Bounded and statistical Model Checking}}\label{verification:formal_methods:bounded} 
A sub-field of model checking is statistical model checking, SMC \eg{} a technique used to provide statistical values on the satisfiability of a property.  In \cite{gros2020deep}, the authors study Deep Statistical Model Checking. An MDP  describes a certain task and a NN is trained on that task. The trained NN is considered as a black-box oracle to resolve the non-determinism in the MDP whenever needed during the verification process. The approach is evaluated on an autonomous driving challenge where the objective is to reach the goal in a minimal number of steps without hitting a boundary wall. 
\cite{sena2019incremental} propose a verification framework based on incremental bounded model checking on NNs. 
To detect adversarial cases on NNs, two verification strategies are implemented: the first strategy is an SMT model checking with a model of the NN and some safety properties of the system. The second strategy is the verification of covering methods. A covering method can be seen as an assertion that measures how adversarial two images are. During the verification process, the properties can be verified or a counterexample is produced.

\hypertarget{verification:formal_methods:sets}{\paragraph{Reachable sets }}\label{verification:formal_methods:sets} 
 
Given an input set $I$ (possibly specified in implicit or abstract form, \eg{} polyhedron, star set, etc.), the corresponding reachable set is defined as the set $\{f(X) : X\in I\}$ of all outputs of the NN $f$ on inputs from $I$. Having access to reachable sets of a DNN for suitable input sets provides important information on the behavior of the model which can be leveraged to verify its safety. 
 In \cite{tran2019parallelizable}, the authors design three reachable computation schemes for trained neural nets, exact scheme, lazy-approximate scheme, and mixing scheme. On top of their techniques, they added parallel computing to reduce the run time of computing the exact schemes, which is performed by executing a sequence of StepRelu operations.The  evaluated their approach on safety verification and local adversarial robustness of feedforward NNs.  \cite{tran2020nnv} propose NNV (Neural Network Verification) a verification tool that computes reachable sets of DNNs to evaluate their safety. Under NNV, a DNN with RELU activation functions is considered to be safe if its reachable sets do not violate safety properties. Otherwise, a counterexample describing the set of all possible unsafe initial inputs and states can be generated \footnote{NNV benefits from parallel computing which makes it faster than Reluplex \cite{katz2017reluplex} and other existing DNN verification frameworks.}.  
 
\hypertarget{verification:formal_methods:abstract}{\paragraph{Abstract Interpretation}}\label{verification:formal_methods:abstract}
In an effort to verify desirable properties of DNNs, researchers have studied abstract interpretation to approximate the reachable set of a network. With abstract interpretation, input sets and over-approximations of their reachable sets are specified using abstract domains, \ie{} regions of space that are described by logical formulas.
The verification process leverages the DNN approximation to provide guarantees. \cite{gehr2018ai2} studied Abstract Interpretation to evaluate the robustness of feed-forward and CNNs. The goal is to propagate abstraction through layers to obtain the abstract output. Afterwards, the properties to be verified are checked on that abstract output. This approach suffers from the quality of properties to be verified since they are based on few random samples.  \cite{li2019analyzing} propose to enhance Abstract Interpretation to prove a larger range of properties. The main contribution of the paper is to use symbolic propagation through neurons of the DNN, to provide more precise results in the range of properties that can be verified through abstract interpretation. 

\hypertarget{verification:formal_methods:lp}{\paragraph{Linear Programming }}\label{verification:formal_methods:lp} 
 
Some verification techniques involve Linear Programming (LP). \cite{lin2019robustness} proposes to verify the robustness of a DNN classifier by leveraging linear programming. The idea of the proposed technique is to find, using nonlinear optimization, a suspicious point that is easier to be assigned a different label than other points and can mislead the classification process. In \cite{rubies2019fast}, a novel algorithm that computes more efficiently splits the NN inputs to reduce the cost of verifying deep feed-forward RELU. For the splitting process to be effective, they estimate lower and upper bounds of any given node by solving linear programs.LP is employed to provide certifiable robustness upper/lower bounds on NNs in \cite{lyu2020fastened}. \textcolor{black}{Given a classifier, the authors want to obtain a guarantee regarding the classification of an instance in case it is perturbed while being restrained to the $L_p$ ball.}
To do so, the authors compute a lower bound $L$ on the output of the network corresponding to the predicted class, as well as an upper bound 
$U$ on the output neurons of all other classes. If one has $U<L$, then the network is guaranteed to be robust to any small perturbations in $L_p$ norm. LP was also used by \cite{guidotti2019repairing} in order to automatically reduce the intent detection mismatch of a prosthetic hand.
 
\hypertarget{verification:formal_methods:milp}{\paragraph{Mixed-integer linear program}}\label{verification:formal_methods:milp}
 
Verification of NNs can also be modelled as a mixed-integer linear program (MILP). \cite{remeli2019towards} propose to verify the satisfiability of the negation of specification rules on a trained NN, both modelled as a MILP.  In \cite{dutta2017output}, the authors employed MILP to estimate the output ranges of NNs given constraints on the input.  \cite{cheng2017maximum} used a MILP solver to find the maximum perturbation bound an NNs can tolerate. This Maximum perturbation bound is defined as the norm of the largest perturbation which can be applied to an input that is strongly associated with a given class, while either maintaining the same predicted class, or keeping the probability of that class among the highest. For this approach, the perturbation bound does not apply when the input underwent an affine transformation. Authors of \cite{bunel2020branch} propose a general framework called Branch-and-Bound for linear and non-linear networks. This framework regroups several pre-existing verification techniques for NNs.  In particular, they improve Branch and bound to tackle ReLU non linearities. They demonstrate the good performance of their branching strategy over various verification methods.
 
\hypertarget{verification:formal_methods:GaussianP}{\paragraph{Gaussian processes verification }}\label{verification:formal_methods:GaussianP}
 
A Gaussian process is a stochastic process that can be used for regression and classification tasks. Their key feature is their ability to estimate their own uncertainty (aleatoric and epistemic), an information which can be leveraged for verification. In \cite{smith2019adversarial}, the discussed approach provides a lower bound on the number of dimensions of the input that must be changed to transform a confident correct classification into a confident miss-classification. However, the experiments only considered Gaussian Processes with Exponential Quadratic kernels. Other kernels types should be investigated. Similar to this approach,  \cite{cardelli2019robustness} provides a theoretical upper bound on the probability of existence of adversarial examples for Gaussian processes. They also argue that because of the convergence of wide networks to Gaussian Processes, their bound has some applicability to DNN. This statement would need to be assessed with more experiments.
 
\hypertarget{verification:formal_methods:other}{\paragraph{Formal verification: other methods\app }}\label{verification:formal_methods:other}
\textcolor{black}{In this section, we present other relevant studies that have been studied to verify DNN. Game theory approach\app which has been studied to evaluate the robustness of a DNN \cite{wu2020game}.}
A Bayesian approach is employed in \cite{ghosh2018verifying} to address the verification of a DNN. They provided a framework that uses Bayesian Optimization (BO) for actively testing and verifying closed-loop black-box systems in simulation. \textcolor{black}{\cite{salay2019safety} investigate errors made by a NN model to identifies relevant failure modes}. The authors use an abstraction of the perception-control linkage of the autonomous driving system. 

\hypertarget{verification:testing}{\subsubsection{Testing Method}}\label{verification:testing}
 
Testing activities are still a key part of traditional software systems. The most basic aspect of it is to input a value to a model for which we know the output that has to be returned (\emph{an oracle}). Many more advanced methods can be used to test a model whether it is based on criteria such as code coverage related metrics, on specifications such as functional testing, on an empirical comparison such as differential testing or more, with possible combinations of different techniques. And in that regard, safety critical systems are no exceptions. Standards such as DO-178C even introduced testing methods as an integral part of the process with the MC/DC criteria, which aim to test combinations of variables yielding different output with only one factor changing. Hence, testing will likely be an integrated part for ML certification. However, because of the paradigm shift that ML systems introduced, previously used methods need to be adapted and revised to take into considerations such differences and new emerging methods will have to be developed to tackle the arising challenges. 
 
\hypertarget{verification:testing:criteria}{\paragraph{Testing criteria}}\label{verification:testing:criteria} 
 
In the direct lineage of code coverage related criteria, a new set of ML related ones have been developed.  \cite{Pei17} was the first to introduce the notion of \textit{neuron coverage} (NC), directly inspired by traditional code coverage. Formally, given a set of neurons $N$, the neuron coverage of a test set $T$ of inputs was originally defined as
\begin{equation}
    \text{NC}(T) := \frac{\#\{n\in N\,|\,\text{out}(n, x)\geq \tau\,\,\,\forall x\in T\}}{\# N},
\end{equation}
where the symbol $\#$ refers to the number of elements in a set (cardinality) and $\text{out}(n, x)$ is the activation of the neuron $n$ when the input $x$ is fed to the network. Simply put, the NC designates the ratio of \enquote{fired} neurons (\ie, whose output is positive/past a given threshold) on a whole set of inputs. The main criticism associated with this measure is that it is fairly easy to reach a high coverage without actually showing good resilience, since it does not take into account relations between neurons as a pattern and discards fine grained considerations such as the level of activation by simply considering a boolean output. Following this first work, related criteria were developed to extend the definition and tackle those issues: KMNC (K-Multisection NC) \cite{Lei18}, Top-k NC \cite{Xiaofei19}, T-way combinations \cite{Ma19}\cite{Sekhon19} or Sign-Sign (and related) coverage \cite{Sun19} based on MC/DC criterion.
 
\hypertarget{verification:testing:coverage_based}{\paragraph{Test generation coverage based}}\label{verification:testing:coverage_based}
 
In general, criteria are not used as a plain testing metric like with traditional software, but rather as a way to incrementally generate test cases that maximize/minimize those given criteria. As most DNNs are trained on image datasets, it's fairly simple and straightforward to generate new images supporting an increasing coverage. To achieve this, techniques such as fuzzing process to randomly mutate base images sampled from a dataset \cite{Xiaofei19}\cite{Guo18}\cite{Demir19} or greedy search \cite{Tian18} and evolutionary algorithm \cite{BenBraiek19} coupled with transformation properties can be used. Those properties are common geometric and/or pixel based transformations. More complex methods can be used such as \emph{concolic} testing \cite{Sun19-2} which mixed symbolic execution (linear programming or Lipschitz based) with concrete input to generate new test cases helped by heuristic based on coverage with improved results compared to other methods. Some techniques leverage Generative Adversarial Network (GAN) \cite{Zhang19} or assimilated in order to generate more realistic test images through coverage optimization. Of course, if the image obtained tends to be more \enquote{natural} compared to fuzzy or metamorphic ones, these methods suffer from the traditional downsides specific to GANs and need extra data to work. Fairly few of those methods have been applied to transportation related datasets, DeepTest \cite{Tian18} being the most prominent example of application. Limiting factors of those techniques, aside from the choice of criteria, remain the time needed to obtain the generated test, the generalizability of the transformation and the effective quality (or validity) of the images. 
 
\hypertarget{verification:testing:other_generation}{\paragraph{Test generation other methods}}\label{verification:testing:other_generation}
 
If coverage based generation is the most widespread technique used, we identified techniques utilizing other methods to manage test cases generation for testing purposes. On image datasets, techniques based on Adaptive Random test \cite{Yan20} leveraging PCA decomposition of network's features, Low-discrepancy among sequences of images and Active Learning \cite{dreossi17}, with metamorphic testing either using entropy based technique with softmax predictions \cite{Udeshi20} or through the search of critical images following those transformations \cite{Pei17-2}. Note that metamorphic testing \cite{Chen20} is an effective proxy for image generation and testing which is also used by techniques mentioned previously such as DeepTest. \cite{Wicker18} used SIFT based algorithm to identify salient parts of the image and optimize for adversarial examples using a two-player game. \cite{Yaghoubi19} proposed a method to generate adversarial examples in a non-linear control system with a NN in the loop, by deriving a function from the interaction between the control and the NN. As for RL applications to transport related tasks, simulation environments are used for training models. In this context, testing methods consist of generating scenarios representing behavior of agents in the environment that are more likely to lead to failure of the learned policy. Meta-heuristic is the preferred method to generate procedural scenarios whether it be evolutionary \cite{Gambi19} or simulated annealing with covering arrays \cite{Truncali18}. Scenario configuration can also be tackled from the point of view of a grammar based descriptive system \cite{Wolschke17} which allows for a flexible comparison between cases.

\hypertarget{verification:testing:fault}{\paragraph{Fault localization/injection testing}}\label{verification:testing:fault}
 
Fault localization is a range of techniques that aim to precisely identify what leads to a failure. In the same vein as traditional fault localization, some papers investigated the root of prediction errors directly on neurons in order to find out which neurons are involved in it, opening the door for potential new test cases generation or resilience mechanisms. Note that fault localization in this context is at the crossroad between \textit{Verification} and \textit{Explainability}, illustrating that methods are not restricted to one domain. Deepfault \cite{Eniser19} was developed in order to identify exactly which pattern of neurons are more present in error inducing inputs, thus allowing a generation of failure inducing tests through the use of gradient of neuron activation on correctly classified examples. 

Fault injection denotes techniques that voluntarily (but in a controlled way) inject faults in order to assess how the system behaves in unforeseen situations. In that setting, TensorFI \cite{Chen20-1} is a specialized tool that aims to inject faults directly in the flow graph of Tensorflow based applications. Note they provide support both for software and hardware (bits) level of faults.
 
\hypertarget{verification:testing:differential}{\paragraph{Differential Testing}}\label{verification:testing:differential}
 
Classical testing methods for DNN classifiers require testing prediction of an input against a ground truth value. If this is possible for a labelled dataset, it's much more complicated when there are no known labels which arise especially during the operation phase, where the ML component is deployed in a live application. Here, we trust the performance of the network on the data distribution to predict in a relevant way. The existence of Adversarial (See \textbf{Section \ref{robustness}} \textit{Robustness}) or Out-of-Distribution (See \textbf{Section \ref{uncertainty:ood}} \textit{OOD detection}) examples demonstrate that it is not the best strategy. In particular, the absence of ground truth, or \enquote{oracle}, is known in software engineering as the \textit{oracle problem}. One way to circumvent this problem is to introduce \enquote{pseudo} oracle to test correctness of an input, which is the main idea behind differential testing. A simple implementation of this process takes the shape of \textit{$N$-versioning}, which consists in $N$ semantically equivalent models that will be used to test an input. $N$-versioning is strongly related to the notion of \textit{ensemble} learning, which uses the knowledge of multiple models. In particular, \cite{Grefenstette18} showed the advantage of ensemble learning against adversarial examples. In the case of pure N-versioning, NV-DNN \cite{Xu19} used this mechanism with a majority vote system in order to reject potential error. The combination of models possible for $N$-versioning was also studied in more detail in \cite{Machida19} where the diversity of different architecture can bring over input error rejection/acceptance is explored. D2Nn \cite{Yu19} uses neurons more likely to contribute to errors in order to build a secondary network. The primary and secondary network can then be used for comparison of predictions within a threshold. Yet, differential testing is not limited to semantically similar models, any semantic comparison allowing to build the \enquote{pseudo} oracle proxy is good. Hence, it's possible to use specificity of DNN through neurons activations \cite{Cheng19} to derive a \enquote{pseudo} oracle, by gathering activation patterns of the train data. Another technique investigated in \cite{Ramanagopal18} used pairs of spatially/temporally similar images for the comparison. The main limitation is determining the policy and mechanism in order to balance correctly between false negative and positive examples as well the semantic modifications. In particular, the change needs to induce diversity in the process while not being too different to preserve the similarity.
\hypertarget{verification:testing:quality}{\paragraph{Quality testing}}\label{verification:testing:quality}
While traditional testing methods aim at testing the model (and so the data specifications indirectly), another interesting idea would be to test the data \textit{directly}. Indeed, since the model is using this data to learn, if one could get a quality measure of the data used, it could add an extra layer to the certification process. We identify some techniques that focus on this approach:  \cite{Mani19} introduced quality criteria for CNN-based classifiers. Similarly, but for bias/confusion testing, \cite{Tian20} defined metrics to quantify bias or confusion for classes of a dataset; the idea here is to test if the model learned the data fairly. In \cite{Ameyaw19}, a Probability of Detection (POD) method on binary classifiers is proposed in order to quantify how well the test procedure can detect vital defects. In a different direction, some techniques deal with test prioritization. Indeed, with the ever growing size of the systems, testing them can become quite expensive. As such, being able to select test samples, which are most likely to induce errors, is important. DeepGini \cite{Feng20} took the approach to quantify the relevance of test inputs, for prioritization purposes, through the likelihood of miss-classification of the model. In \cite{Alagoz17} test instances were instead clustered based on their semantic and test history and some statistical measures, established on similar test failures to prioritize the most important test samples, are used. While in \cite{Gladisch20}, datasets were analyzed to extract relevant features and combinatorial testing is used in order to cover as many cases as possible with a small number of tests. 
 
\begin{tcolorbox}[colback=blue!5,colframe=blue!40!black]
\begin{itemize}
    \item Verification is part of the ML certification process, as it encompasses testing methods that can probe the systems for potential short-comings or infringement of mandatory desirable properties, in order to expose them, which strengthen certification procedures.
    \item Formal verification for ML  can suffer from combinatorial explosion regarding the size and the complexity of the model to be verified. \textcolor{black}{(Section \ref{verification:formal_methods})}
    \item The verification process requires a formal representation of the system, which is subject to interpretation as to which best describes the system. \textcolor{black}{(Section \ref{verification:formal_methods})}
    \item Testing methods for ML certification generally reuse traditional methods such as coverage based testing \textcolor{black}{(Section \ref{verification:testing:coverage_based})} or differential testing \textcolor{black}{(Section \ref{verification:testing:differential})}, while taking into account ML specificity to adapt them.
    \item Most of those methods however are based on empirical considerations or ad-hoc observations \textcolor{black}{(Section \ref{verification:testing})}, hence they would benefit from either a more theoretical approach to derive criteria or a combination with more formal methods \textcolor{black}{(Section \ref{verification:formal_methods})} presented in the previous section to bolster their effectiveness.
\end{itemize}
\end{tcolorbox}
 
\begin{figure}
\begin{center}
\includegraphics[width=0.55\linewidth]{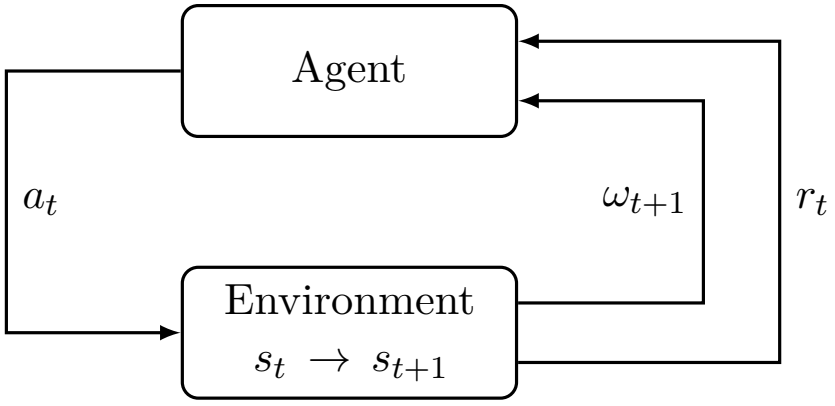}
\caption{Agent interacting with its environment \cite{LavetDRL2018}.}
\label{fig:agent-E}
\end{center}
\end{figure}

\hypertarget{safe_rl}{\subsection{Safe Reinforcement Learning}}\label{safe_rl}
Suppose that an agent interacts with an environment by perceiving the environment, performing actions and then receiving a reward signal from the environment. The main task here consists of learning how to perform sequences of actions in the environment to maximize the long-term return which is based on the real-valued reward. Formally, the RL problem is formulated as a discrete-time stochastic control process in the following way: At each time step $t$, the agent has to select and perform an action $a_t \in A$. Upon taking the action, (1) the agent is rewarded by $r_t \in R$, (2) the state of environment is changed to $s_{t+1} \in S$, and (3) the agent perceives the next observation of $\omega_{t+1} \in \Omega$. Fig.~\ref{fig:agent-E} illustrates such agent-environment interaction. An RL agent aims at finding a policy $\pi \in \Pi $ that maximizes the expected cumulative reward or \textit{return}:
\begin{equation}
V^\pi(s) = \mathbb{E}[\,R_t | s_t = s\,], \text{with}\,\, R_t = \sum_{k=0}^{\infty} \gamma^k r_{t+k+1},
\end{equation}
where $\gamma \in [0,1]$ is a discount factor that applies to the future rewards. A deterministic policy function indicates the agent's action given a state, $\pi(s): S \rightarrow A$. In the case of stochastic policies, $\pi(s, a)$ indicates the probability of choosing action $a$ in state $s$ by the agent.\\
Recently, researchers have successfully integrated DL methods in RL to solve some challenging sequential decision-making problems \cite{Goodfellow-et-al-2016}. This combination of RL and DL is known as deep RL, and benefits from the advantages of DL in learning multiple levels of representations of the data to address large state-action spaces with low prior knowledge. For example, a deep RL agent has successfully learned by raw visual perceptual inputs including thousands of pixels \cite{mnih2015human}. Deep RL algorithms, unlike traditional RL, are capable of dealing with very large input spaces, and indicating actions that optimize the reward (e.g., maximizing the game score). As a consequence, imitating some human-level problem solving capabilities becomes possible \cite{gandhi2017learning, moravvcik2017deepstack}.

\paragraph{Value-based approaches}
The value-based algorithms in RL aim at learning a value function, which subsequently makes it possible to define a policy. The value function for a state is defined as the total amount of discounted reward that an agent expects to accumulate over the future, starting from that state. The Q-learning algorithm \cite{watkins1992q} is the simplest and most popular value-based algorithm. In its basic version, $Q(s, a)$ with one entry for every state-action pair is used to approximate the value function. To learn the optimal Q-value function, the Q-learning algorithm uses a recursive approach which consists in updating Q-values based on Bellman equation:
\begin{equation}
Q(s_t,a_t) \leftarrow Q(s_{t},a_{t}) + \alpha[\,r_{t+1} + \gamma\max_{a'\in A} Q(s_{t+1},a') - Q(s_{t},a_{t})\,],
\end{equation}
where $\alpha$ is a scalar step size called the learning rate. Given Q-values, the optimal policy is obtained via:
\[
\pi(s) := \argmax_{a \in A} Q(s,a).
\]
The idea of \textit{value-based deep RL} is to approximate the value function by DNNs. This is the main principle behind deep Q-networks (DQN), which were shown to obtain human-level performance on ATARI games \cite{mnih2015human}.
 
\paragraph{Policy gradient approaches}
Policy gradient methods maximize a performance objective (typically the expected cumulative reward $V^\pi(s)$) by discovering a good policy. Basically, the policy function is directly approximated by a DNN meaning that the network output would be (probability of) actions instead of action values. It is acknowledged that policy-based approaches converge and train much faster specially for problems with high-dimensional or continuous action spaces \cite{agostinelli2018reinforcement}. The direct representation of a policy to extend DQN algorithms for addressing continuous actions was introduced by Deep Deterministic Policy Gradient (DDPG) \cite{lillicrap2015continuous}. This algorithm updates the policy in the direction of the gradient of Q which is a computationally efficient idea. Another approach is using an actor-critic architecture which benefits from two NN function approximators: an actor and a critic. The actor denotes the policy and the critic is estimating a value function, e.g. the Q-value function.\\
\\
\textbf{Safety in RL}\\
In safety-critical environments, researchers concentrate not only on the long-term reward maximization, but also on damage avoidance since the agent does not always perform safe actions that may lead to hazards, risks, and accidents in the environment. For example, in autonomous driving, an unsafe action may lead to a collision between the ego vehicle and other vehicles, pedestrians, or another object in the environment. From a theoretical perspective, a high-level argument on safety of RL-based systems was reported in \cite{bragg2018acceptably} by analyzing technical and socio-technical factors that could be used as the basis for the safety case of RL. They stated that the traditional approach to safety which assumes a deterministic (predictable) system would not work for ML in general and RL in particular, suggesting an \enquote{adaptive} approach to safety. Experience-Based Heuristic Search (EBHS) \cite{bernhard2018experience} extended the idea of combining RL with search-based planning to continuous state spaces by improving heuristic path planner using learned experiences of deep RL. They have applied Deep Q-learning from Demonstrations (DQfD) for learning from demonstration which is based on pretraining from an expert policy. They have shown computational advantages and reliability of such an approach for a standard parking scenario. However, the evaluated scenarios are too simple compared to real world settings.\\
In this paper, we identify three main categories concerning approaches to safe RL: 
\begin{itemize}
\item \textbf{Post-optimization}: Adding an additional safety layer after the RL to exclude unsafe actions like safe lane merging in autonomous driving \cite{hart2019lane},
\item \textbf{Uncertainty estimation}: Estimating what the agent does not know in order to avoid performing certain actions, making the agent's behaviour robust to unseen observations like collision avoidance for pedestrian \cite{lutjens2019safe},
\item \textbf{\textcolor{black}{Certification of RL-based controllers}}: Providing theoretical guarantees for RL-based controllers like \cite{nguyen2020towards}.
\end{itemize}
 
\hypertarget{safe_rl:post_opt}{\subsubsection{Post-optimization}}\label{safe_rl:post_opt}
The idea behind these approaches is to add an additional safety layer after RL to exclude unsafe actions. Authors in \cite{hart2019lane} have employed a policy-based RL algorithm, i.e. soft actor-critic, to solve complex scenarios in autonomous car driving, such as merging lanes with multiple vehicles. To make it applicable in safety-critical applications, they have used a non-linear post-optimization to optimize the RL policy. After the post-optimization, an additional collision-check is performed to check for a high level of safety. Experiments were performed using Divine-RL, an autonomous driving simulation environment and results showed that the collision rate dropped to almost zero for the given time-horizon. Comparison to alternative approaches and evaluation using real road traffic have remained untouched. A dynamically-learned safety module has been proposed for RL agents in \cite{baheri2019deep}. The module is a recurrent NN trained to predict whether future states in the environment can lead to an accident or a collision. Hand-crafted rules and the learned behavior (from the history of the system) were combined in this approach. Experiments were conducted in a simulation environment with an autonomous driving car on a highway that can be surrounded by other traffic vehicles. Results showed that the proposed approach significantly reduces the number of collisions. In \cite{isele2018safe}, authors have used a prediction mechanism for safe intersection handling in a RL-based autonomous vehicle. The prediction mechanism aims both at minimizing disruption to traffic (as measured by traffic braking), while avoiding collisions, and at the same time, at maximizing distance to other vehicles, while still getting through the intersection in a fixed time window. 

Distinguishing task failures has been proposed in \cite{o2020dependable}. The approach consists in training a NN in an RL environment: adjusting the velocity of a robot to avoid moving obstacles. Two failures were identified: 1) misclassifications that do not violate the safety constraints, and 2) harmful failures. Finally, they added a safety function to prevent harmful failures and show through experiments that it did almost nullify harmful failures. However, the probability distribution of testing and operating conditions was assumed to be known. Yet this assumption seems restrictive for real world problems (they can be calculated, for instance, by studying weather patterns in the environment).
 
\hypertarget{safe_rl:uncertainty}{\subsubsection{Uncertainty estimation}} \label{safe_rl:uncertainty}
Similar to estimating the uncertainty in Section \ref{uncertainty}, estimating what the agent does not know is found helpful to avoid unsafe actions in the literature. These approaches attempt to make the agent behaviour robust and then safe to unseen observations. Model uncertainty estimation is employed in \cite{lutjens2019safe} to develop a safe RL framework for collision avoidance of pedestrians. The main component is an ensemble of LSTM networks that was trained to estimate collision probabilities. These estimations have been used as predictive uncertainty, to cautiously avoid dynamic obstacles. The results showed that the model knows what it does not know; the predictive controller employs the increased regional uncertainty in the direction of novel obstacle observations, to act more robustly and cautiously in some novel scenarios. Cautious Adaptation in RL (CARL) \cite{zhang2020cautious} is a general safety-critical adaptation task setting to transfer knowledge (or skills) learned from a set of non-safety-critical source environments (e.g., simulation environment in which failures do not have heavy cost) to a safety-critical target environment. A probabilistic model is trained using model-based RL to capture uncertainty dynamics and catastrophic states of source environments. The results are promising and one of the tested environments is Duckietown car driving. Worst Cases Policy Gradients (WCPG) \cite{tang2019worst}, a novel actor-critic framework, has been proposed to model uncertainty of the future in environments. In fact, WCPG estimates distribution of the future reward for any state and action and then learns a policy based on the uncertainty model (optimized for different levels of conditional Value-at-Risk). Therefore, the obtained policy is sensitive to risks and avoids catastrophic actions. While WCPG can be adjusted dynamically to select risk-sensitive actions, it performed better compared to other state-of-the-art RLs in terms of collision rate in two scenarios of unprotected left turn and merge into highway in a driving simulator.

Some other researchers attempted to estimate the possibility of agent’s failure to prevent hazardous actions leading to such failure. For example, reliable evaluation of the risk of failure in RL agents has been studied in \cite{uesato2018rigorous}. Authors proposed a continuation approach for learning a failure probability predictor to estimate the probability of agent failures given some initial conditions. The main idea is employing data gathered from some less robust agents which fail often to improve the learning. The proposed approach has been successfully evaluated on two RL domains for failure search and risk estimation: 1) Driving domain: the agent (on-policy actor-critic) drives a car in the TORCS simulator and rewarded for driving forward without crashing, and 2) Humanoid domain: the agent (off-policy distributed distributional deterministic policy gradient-D4PG) runs a 21-DoF humanoid body in the MuJoCo simulator and rewarded for standing without falling.

\hypertarget{safe_rl:stability}{\subsubsection{\textcolor{black}{Certification of RL-based controllers}}}\label{safe_rl:stability}
In \cite{nguyen2020towards}, authors have provided quantitative stability conditions for a special kind of DL-based controllers in a closed-loop manner. The controller is a non-autonomous input-output stable deep NN (NAISNet), which consists of a residual NN composed of several blocks where the input and a bias is fed to each layer within a single block. They employed Lyapunov functions, a classic approach for analyzing stability of dynamical systems in control theory and evaluated their approach on a simple controller, namely a continuously stirred tank reactor. Lyapunov functions are leveraged also in \cite{berkenkamp2017safe}, to provide safety in terms of stability in a continuous action space environment. The idea is to extend Lyapunov stability verification to statistical models of the dynamics for obtaining high-performance control policies with provable stability certificates. Moreover, they showed that one can effectively and safely explore the environment in order to learn about the dynamics of the system and then employ such information to improve control performance and expand the safe region of the state space. A probabilistic model predictive safety certification (PMPSC) scheme was proposed for learning-based controllers in \cite{wabersich2021probabilistic}. This approach is designed to equip any controller with some probabilistic constraint satisfaction guarantees. They have combined Model Predictive Control (MPC) with RL to achieve safe and high performance closed-loop system operation. They have successfully tested their approach to learn how to safely drive a simulated autonomous car along a desired trajectory, without leaving a narrow road. For car simulation, they have used an a priori unknown nonlinear, time-invariant discrete-time dynamical systems described by equations of state-space representation. A modified version of the classical policy iteration algorithm has been presented in \cite{chakrabarty2019approximate}, to preserve safety by constraint satisfaction. The key idea is to compute control policies and associated constraint admissible invariant sets, that both ensure the system states as well as control that inputs never violate design constraints. The preliminary results revealed that asymptotic convergence of the sequence of policies to the optimal constraint-satisfying policy is guaranteed.
 
In another work, authors have proposed to use projection to ensure that an RL process is safe without disrupting the learning process \cite{gros2020safe}. It is based on direct minimization of the learning equation under some safety constraints. MPC techniques are used by the authors to compute the safe set of states. The idea of Bayesian MPC has been introduced in \cite{wabersich2020bayesian}. Then authors have extended this idea to a theoretical framework for learning-based MPC controllers and have proposed a modified version that introduces cautiousness \cite{wabersich2020performance}. This cautious Bayesian MPC formulation uses a simple state constraint tightening that relates the expected number of unsafe learning episodes, which could be defined particularly for each system or scenario, to the cumulative performance regret bound. They have successfully tested their approach on a generic drone search task, where goal is defined as collecting information about an a priori unknown position using a quadrotor drone. The idea of verification-preserving model updates has been introduced in \cite{fulton2019verifiably}. The paper aims at obtaining formal proofs for RL in settings where multiple environmental models must be considered. For this purpose, authors presented an approach using a mix of design-time model updates and runtime model falsification for updating an existing model while preserving the safety constraint. 
 
Some approaches guarantee safety of RL in specific types of problems. For example, constrained cross-entropy \cite{wen2020constrained} addressed safe RL problem with constraints that are defined as the expected cost over finite-length trajectories. The constrained cross-entropy generalizes the cross entropy method for unconstrained optimization by maximizing an objective function, while satisfying safety requirements on systems with continuous states and actions. Logically-Constrained Reinforcement Learning (LCRL) \cite{hasanbeig2020towards} is a general framework that guarantees the satisfaction of given requirements and guides the learning process within safe configurations in high-performance model-free RL-based controllers. Authors have used Linear Temporal Logic (LTL) to specify complex tasks encompassing safety requirements. LCRL has been successfully evaluated on a set of numerical examples and benchmarks, including NASA Opportunity Mars-rover.
 
Besides online RL, it is possible to learn from pre-collected data since usually large amounts of data have been collected with existing policies. However, certifying constraint satisfaction, including safety constraints, during off-policy evaluation in sequential decision making, is not straightforward and could be challenging. Batch policy learning under constraints \cite{le2019batch} adapted abundant (non-optimal) behavior data to a new policy, with provable guarantees on constraint satisfaction. Authors have proposed an algorithmic framework for learning policies from off-policy data respecting both primary objective and constraint satisfaction. An algorithm for measuring the safety of RL agents has been proposed in \cite{bacci2020probabilistic}. A controller is modelled to characterize the actions taken by the agent and the possibilities that result from taking them. The controller is modelled as a MDP and probabilistic model checking techniques are leveraged to produce probability guarantees on the behaviour of the agent. The verification of the controller’s model aims at finding the probability of reaching a failure state given a particular initial state.  

\textcolor{black}{Barrier functions have been widely used to design safe RL-based controllers\app \cite{marvi2020safe,cheng2019end,yang2020safe,deshmukh2019learning}. Based on its definition, the value of a barrier function on a point is increased to infinity, as the point gets close to boundaries of the feasible region of an optimization problem \cite{nesterov2018lectures}. Such functions are alternatives for inequality constraints: a penalizing term is added to the objective function.} 

\begin{tcolorbox}[colback=blue!5,colframe=blue!40!black]
\begin{itemize}
    \item In safety-critical environments, it is not enough to focus on reward maximization since the agent’s actions may lead to hazards, risks, or damages in the environment. Safe RL helps to certify agent’s actions by determining a safe set of actions or forbidding potential risky or uncertain actions.
    \item Adaptive approaches to safety should be considered for RL-based systems: Complying with pre-defined constraints is not enough and the system should be able to deal with situations that cannot be predicted. Similarly, assuming a prior set of safe states in many real-world problems is not realistic. \textcolor{black}{(Section \ref{safe_rl:post_opt})}
    \item Although existing results of certifying stability of RL-based controllers are interesting, the application scope is still non-practical: certification usually is performed for toy problems or simple control tasks, so they should be extended to address realistic tasks. \textcolor{black}{(Section \ref{safe_rl:stability})}
    \item Generalizability of uncertainty estimation can be challenging: an approach may successfully estimate failure distribution for a particular situation but not for others. Therefore, various safety critical scenarios/environments should be tested to assess the effectiveness of such methods. \textcolor{black}{(Section \ref{safe_rl:uncertainty})}
    \item Using other ML approaches to make safe decisions in RL is not safe: outcomes of any ML-based systems should be investigated to be error-free and safe to apply, therefore one can not rely on their prediction for safety of RL agents.
\end{itemize}
\end{tcolorbox}

\hypertarget{direct_certif}{\subsection{Direct Certification}} \label{direct_certif}

By \enquote{Direct Certification} we refer to any paper that aims at a general framework for ML certification through a certain number of methods, considerations or steps. Although no ISO 26262 type standard has been created yet for ML-based systems, some papers indeed tackle this challenge. As such, those papers differ from papers described in previous sections, as they do not use a specific method to cover a certain aspect of certification. Instead, they try to address higher level considerations and attempt to describe how all those methods could fit together to reach a viable certification process. 

Most papers rely on already established concepts such as assurance cases type, graph structured notation (GSN) or even adaptation of standards. These concepts are viewed as a foundational starting point to be adapted to the specificities of ML. The GSN was used to build an approach similar to assurance cases and hazard analysis \cite{Fujino19}\cite{Rudolph18}\cite{Gauerhof18}. Authors defined multiple safety layers for verification, with for instance a layer about different states of the system, \eg{} normal or emergency, with specification of what to do \cite{Fujino19}. If those methods guarantee a better comprehension of what is desired, they remain very high-level. 
 
\cite{Salay18} discuss how ISO 26262 could be adapted to ML by listing the main obstacles for adaptation; notably the lack of specification that is not covered strictly by train data and the lack of interpretability of models. The authors also advocate for coverage metrics/safety envelopes to investigate training data and recommend avoiding end-to-end approaches, that is a system solely based on ML. They argue that such a system would be incompatible with the ISO 26262 framework assumptions about stability of components, since the ML model weights change depending on the training and training set. Moreover, to circumvent the actual problem, it is recommended to use techniques based on intent/maturity rather than clear specification, which echoes the \enquote{Overarching Goals} we mentioned in Introduction. Note that some methods described in previous sections highlight the necessity to take into consideration aspects of ISO 26262, and even propose some techniques to serve as a basis for further inquiries of specific points of the standards. For instance, \cite{Heinzmann19} covered safety constraints built through expert knowledge and/or statistical data with a given severity level, which can then be assessed for safety violation in simulation. This could serve as a good basis for further analysis or evaluation of safety requirements which is one point made by ISO 26262. While ISO 26262 is the most studied proxy for ML certifications standard, it is not the only one. As discussed in the introduction, ISO21448 (SOTIF) intended functionality is an interesting standard for ML, as it focuses on unexpected behavior, such as prediction on OOD and Adversarial Examples. Aside from techniques mentioned in previous sections, \cite{Henriksson19-2} developed on the OOD problem and how it relates to SOTIF. \cite{Pedroza19} proposed an overall iterative generic (OGI) method for developing safe-by-design AI-based systems. For air transportation, \cite{Cofar20} discussed a run-time assurance based on the ASTM F3269-17 for bounded behavior of complex systems. \textcolor{black}{They mainly rely on monitoring ML components with backup functionality, inside a larger non-ML systems}. 

In \cite{Biondi20} a different approach for the actual certification of ML was suggested; it relies on an architecture to control the ML component, with the fault recovery mechanism being the cornerstone. They claim that, using this architecture, DNN no longer needs to be certified, only the fault recovery system needs to be. This observation is similar to techniques using monitors to control the ML actuators such as \cite{Cofar20} that was discussed previously or \cite{Kaprocki19} which made use of ASIL (Automotive Safety Integrated Level). However, to be effective, this requires that the monitor can be more easily certified than the actual ML system. 

Other papers \cite{Burton19}\cite{Aravantinos19}\cite{Biondi20}\cite{Gauerhof20}\cite{Rahimi19} raised similar observations about ML limits or issues, which we mentioned in the previous sections. Those issues represent a direct threat to certification; lack of specification, distributional shift, adversarial examples, out-of-distribution problems, lack of interpretability, lack of testing/verification approaches. Methods presented earlier are cited as being a way to tackle precise problems, but they also make the case for more traditional techniques such as; redundancy and fault tolerance subsystems, look ahead components, backup systems, coverage criteria or traceability through collection of NNs related artifacts such as weights or versions.

In all of those studies, little to none practical case study were presented with detailed use cases. Described techniques do not necessarily make use of all the considerations we elaborated on in previous sections, showing there is still ground for a unified process. As such, there is still no concrete standard or draft of standard. We have nonetheless illustrated that this preoccupation is currently understood by the scientific community.
 
\begin{tcolorbox}[colback=blue!5,colframe=blue!40!black]
\begin{itemize}
    \item Direct Certification encompasses all high-level discussion about the ML certification process, in particular it also includes a draft of standards or process in order to tackle ML certification.
    \item Some efforts have been made to adapt existing standards or considerations to specifics of ML, in particular ISO26262 seems a promising basis for car related systems.
    \item However, there is no clear process established and most studies offer incomplete approaches with no practical real case studies. 
    \item Considering the plethora of methods developed within the various certification categories (Robustness, Uncertainty, Explainability, Verification), there are countless opportunities to apply them simultaneously in use case studies, inching closer to universal ML standards.
\end{itemize}
\end{tcolorbox}

\section{Future Direction and Research Opportunities in ML Certification}\label{sec:discussion} 
 
The low-level, technical considerations we derived from our paper reviews allowed us to present state-of-the-art techniques as well as existing challenges and limitations. \textcolor{black}{It also allowed us to highlight that certification is of great importance for ML software systems, even more so than for traditional ones because of the paradigm shift introduced. In this section, we discuss at a higher level, the future possibilities for ML certifications in accordance with our previous discussions.}
 
\paragraph{Increasing diversity of safety-critical use cases}
 
Safety-critical considerations in ML have been gaining a lot of attention over the past years; since 2017, the number of papers dealing with this topic has been steadily rising, \textcolor{black}{showing} 
the growing preoccupation of the community on this topic. Concerning \enquote{certification} considerations, we find 
most research \textcolor{black}{studies 
applications involving image data, popular for being
easily accessible and visually interpretable}. As such, the majority of the datasets used in experiments were: MNIST, CIFAR and ImageNet. While these datasets are interesting to introduce new concepts and theory, in particular with regards to understanding models behaviors, their relevance seems quite limited, when it comes to empirically evaluating techniques for safety-critical systems in aviation or automotive. Only about a fifth of all datasets used are directly related to such fields. Moreover, while aviation seems to have a standard dataset with ACAS-XU, there is no clear \enquote{driving} dataset that is widespread in the case of automotive. 
\textcolor{black}{Moreover, although our objective is to ensure critical-safety of general ML, DNNs (especially CNNs) are overwhelmingly represented. Given the importance of structured (tabular) datasets, and that alternative models such as Random Forests and Gradient Boosted Trees can work as well (or better) than DNNs on this data, we recommend to extend the safety-considerations previously presented to these models.}
 
\paragraph{Bolstering partnership between Academia \& Industries}
 
While academia research represents the majority of the papers 
screened in our study (around 60\%), we believe that collaborations between academia and industry (only around 32\% of screened papers) represent a great opportunity for the research on ML certification. Indeed, as explained in the review, assessing the critical-safety of ML systems requires more practical datasets, which could be accessed through industrial partnership. A deeper collaboration between industry and academia would be mutually beneficial as it would encourage researchers to adapt the current ML models (or develop new ones) to meet the specific constraints of a given industrial partner. In the short term, we think 
adapting ML methods to respect industrial constraints on a partner-by-partner basis is more realistic than aiming at developing uniform safety standards. We therefore trust that initiatives such as DEEL\footnote{\url{https://www.deel.ai}} (DEpendable and Explainable Learning) which allows for cooperation between academia and industry are to be encouraged, in order to foster the development of suitable concepts and techniques.
 
\paragraph{Adapting proven techniques to ML specificities}
 
ML certification research should not be restricted to devising new techniques, as already established methods can be useful, while already benefiting from a solid background. The verification techniques we presented are perfect examples. Whether it is using Formal Solver such as Linear Programming or more classical Software Engineering techniques such as MC/DC, there are already plenty of existing techniques that could potentially offer extra safety to ML. Those techniques would only require adaptation to the \textcolor{black}{specificities of ML}. For instance, \cite{Sun19} devised a criteria adapted to DNNs that is analogous to the traditional MC/DC, which is used in classical software testing.
 
\paragraph{Deriving formal guarantees}
Currently, \textcolor{black}{only robustness properties of ML models appear to have been rigorously formalized to yield strong guarantees that supplement empirical evidence}. Indeed, empirical evidence is limited by the fact that it can only assess the safety of the model on the finite datasets used in experiments. Other certification sub-fields such as OOD detection, Uncertainty, Explainability, and Testing currently lack formalism and guarantees. For example, in OOD detection, no formal definition of what constitute the set of all OOD inputs is used and OOD detectors are currently evaluated by simply measuring their ability to discriminate between instances from the dataset used for training (MNIST for example), and instances taken from a completely different OOD dataset (notMNIST for example). Although this type of experiment is helpful to compare OOD detectors, it cannot provide true guarantees that the model \textcolor{black}{will perform safely in production as the OOD dataset may not represent of all possible OOD inputs}. Moreover, we believe that the applicability of these techniques requires guarantees on False Negative rates, \ie{} how often an unsafe prediction on a OOD sample will be performed in deployment, without raising an alarm, which are yet to be developed.
Those considerations could lead to a \enquote{safeguard} system, \ie{} letting the ML part acts until it strays away from a \enquote{safe zone}, which would trigger a fail-safe option. For instance, in \cite{Biondi20}, authors advocate for a monitoring system that can watch over the ML component and trigger a fail-safe. They argue that only this monitoring system needs to be certified. Such monitoring system could potentially labels the states of the ML components as safe/unsafe, based on theory-grounded guarantees.
 
\paragraph{Finding new avenues to complement formal guarantees}
 
\textcolor{black}{It is possible formal guarantees might not be achievable for certain parts of the ML framework, especially for certification sub-fields where one does not have access to ground-truth values.} For uncertainty, 
it is not clear whether or not specific computations of aleatoric and epistemic uncertainties are meaningful. \textcolor{black}{At a high level, it is expected that good uncertainties should correlate with model error, although there is no universal way to evaluate the adequacy of the uncertainty estimates as a proxy for the prediction failure.} In post-hoc explainability, it is not always possible to know what is the ground-truth for explanations, because the models that are studied are black box by nature and because it is hard to extract human-understandable summaries of the reasoning behind their decisions. When testing data quality, it is not clear what qualifies as \enquote{good data} and what metrics can encode the right notions of quality. This specific issue faced in testing is in a sense similar to the non-testable program paradigm introduced by \cite{Weyuker82}. Even formal verification is limited by the formalization of the property it aims to verify. If a property cannot be properly formulated, it cannot be verified. Taking inspiration from differential testing, the lack of ground truth could be tackled by introducing novel pseudo-oracles. This would in turn possibly lead to other forms of guarantees.

\paragraph{Studying cross sub-fields of certification}
 
\textcolor{black}{Most papers focus on a single category in search of interesting results. However, in practice, all sub-fields are expected to work hand in hand. As such, more studies should investigate the connections between sub-fields and the trade-offs involved.}
For instance, \cite{Sehwag19} demonstrated that adversarial OOD can fool both OOD and AE dedicated detectors. Possible connections we identified in our review are the following:
 
\begin{itemize}
    \item Robustness is linked to OOD detection as the right OOD detector should reject all samples on which the system cannot be trusted while Robustness measures in a sense \enquote{how much} samples we can safely predict on. More generally, the notion of distributional shift, that encompasses those two sub-fields, could benefit from a unified treatment.
    \item We suspect Robust training and Explainability of DNNs are deeply connected. On the one hand, we could expect adversarially robust NNs to have more interpretable saliency maps because their decisions cannot be significantly altered by spurious perturbations of the input. On the other hand, as noted in \cite{madry2017towards}, to reliably defend against adversarial attacks, DNNs require more capacity, making them less interpretable in the process. Hence, connections between model capacity, robustness, and explainability \textcolor{black}{are non-trivial and should be thoroughly explored}.
    \item Uncertainty and Explainability share 
    similarities. Indeed, they both help machines mimic how human beings make decisions, therefore increasing trust users have in models. For this reason, we think that further studying the relation between model uncertainty and explainability could foster understanding in the two respective domains. For instance, we suggest adapting some of the metrics used in uncertainty to explainability. As stated previously, uncertainties are currently used as proxies for prediction confidence, and are expected to be high on instances where the model fails and low on instances where it predicts correctly. Similar notions could be extended to post-hoc explanations, \eg{} when studying an instance on which the model fails to make the right prediction, we would expect the explanation to be aberrant or misleading. 
    \item \textcolor{black}{There are also possible connections between Adversarial Robustness and Uncertainty. As discussed earlier, some techniques from both categories modify the standard training procedure by including adversarial loss (Robustness), loss attenuation (Aleatoric Uncertainty), and MCDropout (Epistemic Uncertainty). Still, it is unclear how these different training objectives interact when used together. Although loss attenuation and MCDropout have previously been used in-tandem during training \cite{gruber2018uncertainties}, the further addition of adversarial loss remains to be investigated.}
    \item Verification and Explainability/Uncertainty. Most work on verification, whether it is through formal methods or test-based generation, focuses on generating adversarial examples or verifying robustness properties. However, we would like to point out that such approaches could also be useful to verify equivalent properties for other concepts related to explainability or uncertainty through formal checking (although a more formal framework for both would be required), or generation of \enquote{corner-case}, as it is currently done for robustness to adversarial examples. 
    \item Data quality and all other fields. Many techniques we have reviewed focus on certifying properties of a fixed model. While there exists a relation between data and models (since the former is used to train the latter), it seems to us that there are some inherent limitations in focusing only on models. In particular, a model can behave correctly regarding one of the properties, while the data itself does not represent the full picture, hence possibly leading to error when the model is put in operation conditions. As such, methods which focus on studying how such considerations are present \emph{directly} in the data are important and why recent efforts have been focused on bettering their collection, processing and analysis \cite{Roh21}.
    \item Finally, we observed studies in the literature that apply Robustness and/or Uncertainty techniques to increase the safety of Reinforcement Learning agents. It would be interesting to go a step further and add explainability to the picture. Indeed, getting insight on the decision-making process of an agent would be an important step in making sure that no future decisions are unsafe.
\end{itemize}
 
Overall, we believe that studying certification of ML from multiple aspects at the same time would not only bring more knowledge into each given sub-field, but would help make significant steps toward a global certification approach.
 
\paragraph{Unifying all sub-fields for a complete standard}
 
As already mentioned, there is currently no standard for ML certification. The attempts that were discussed in \textbf{Section \ref{direct_certif}} generally stick to high-level considerations, do not cover all aspects of a system certification, and/or do not take into account possible trade-off of using different certifications aspects at the same time. However, it is clear that there is some basis that can be adapted, using previously defined standards such as ISO 26262. We trust that a true standard can only be defined if the expertise from all the discussed sub-fields is brought together. In this perspective, a long term goal for the research community would be to devise a framework that provides clear-cut criteria, to generate models that are robust to adversarial perturbations and distributional shift (Robustness), that know what they don't know (Uncertainty and OOD detection), that can provide insight on how their decisions are made (Explainability) and whose properties can be verified/tested (Verification).

\section{\textcolor{black}{Related Works}}\label{sec:rel}

\textcolor{black}{Certification standards in classical software engineering have been widely discussed in the past; Authors in \cite{Kornecki08} reported on certification in safety-critical standards focusing on aviation ones, two standards used in aviation were compared in \cite{Youn14}, while authors analyzed techniques compliance with ISO 26262 in \cite{Manoj15}. However, all those standard analysis focused on non-ML adapted standards, as such they did not consider all the problems or challenges of certifying ML-based systems we discussed in this paper.}

\textcolor{black}{Regarding ML oriented certification, multiple papers tackled the issue, but, to the best of our knowledge, none to the extent we did; \cite{Rajabli21} focused on validation/verification (\ie{} testing, adversarial examples, software cages, formal methods, fault injection) in autonomous car specifically, with a discussion on ISO26262. However, they did not tackle uncertainty/explainability and focused solely on autonomous car. \cite{Zhang20} proposed a SLR similar to ours, that tackled testing and verification of DNN in safety-critical control systems, with many car/aerial systems represented, by taking the IEC 61508-3 standard and discussing how techniques could validate each point of the standard. While they discussed testing completeness, verification, robustness and interpretability, the did not consider uncertainty and OOD in their study, which are an important part of the certification of ML systems as we have seen. Moreover, their study considered papers from 2011 to 2018, which mean papers from 2019-2020 were not considered. According to our search results in Section \ref{sec:results:stats}, these papers constitute the majority of the papers we gathered in our study, meaning many new improvements were brought about recently and not discussed in their study. Robustness, monitoring and explainability were briefly reviewed in \cite{Hendrycks21} but many details were left out of their short discussion. While they considered cyber-attack issues (out-of-scope in our study), they did not considered RL or direct certification applications. \cite{Vidot21} focused solely on robustness and explainability with some points on verification. \cite{Zhang20-2} focused on ML and robustness testing (testing properties such as robustness, testing components, testing workflow and application scenarios) extensively, with some points on fairness and interpretability, yet they did not tackle uncertainty, OOD or direct certification.}

\section{\textcolor{black}{Threats to Validity}}\label{sec:threats}
\paragraph{\textcolor{black}{Construction validity}}
\textcolor{black}{
Potential limitations to the process of our SLR can come from 1) the time range and 2) the keywords used for the search. We chose to limit the study to a period of 5 years from 2015 to September 2020. As the 5 years period is a typical time range in SLRs, most of the papers were found in 2019-2020 (Figure \ref{fig:yearly_count}) indicating a recent field of study. Because we did our search by September 2020, we might have missed some of the latest developments. However, the number (over 200 which is much more than the average number of papers in SLR) and variety of papers gathered ensured a good coverage of most of the sub-field involved in the discussion.
}

\textcolor{black}{
Regarding keywords, it is hard to strike a good balance between gathering relevant papers for the certification process while not being too restrictive, missing potentially interesting certification-oriented ideas about. We tried to encompass the two aspects we deemed important for the study; the \enquote{certification} and the \enquote{machine learning}. For \enquote{machine learning}, we tried to list down as many synonyms as possible to remain general. For \enquote{certification}, we noticed after testing some queries that \enquote{safety-critical} or \enquote{safety assurance} were much more related on their own to the notion of certification, than \enquote{certification} itself. This could lead to some False Negatives, since the introduction/abstract of papers present concepts, where more general vocabulary is used, which is the case with the word \enquote{certification} or its derivative form. We found that transportation vocabulary leads to more relevant papers. This, coupled with the fact that most software engineering certification processes of safety-critical systems are transportation based, led us to take into account the necessity of including transportation in our keywords. Our goal was to give an overview as complete as possible, but being exhaustive was not possible because of the wide number of sub-fields and definitions used.}

\paragraph{\textcolor{black}{Internal validity}}
\textcolor{black}{
Internal limitations can be related to the extraction of data from the papers as well as the criteria used (quality and inclusion/exclusion). We mitigated this issue by ensuring at least two reviewers covered each paper to make sure correct information was understood. Moreover, we ask authors of some papers we deemed important for our study to screen the summary we have made of their paper to be sure not to miss anything.
}

\textcolor{black}{The other limitation is with regard to our choice of criteria to prune papers. We based on criteria that were previously used in other SLR and we only added necessary ones that are specific to our question to further prune papers. In all steps, we choose to be conservative; hence we probably accepted more papers than we should have which can explain the huge number of papers in the end. This was done to ensure all nuances of information were properly gathered. This also means that some papers might not directly relate to certification in \emph{the strict sense}, but are still relevant for the discussion.}
\paragraph{\textcolor{black}{External validity}}
\textcolor{black}{
External limitations are related to the generalization of the results presented. We tried to cover as much as possible the different aspects of the certification problem for ML, by adding as much detail as possible. If some precise concepts or techniques were not tackled (because of the search patterns, criteria exclusion, ...), the general topics covered are relevant and can be generalized, even if some adjustment needs to be made.}

\paragraph{\textcolor{black}{Conclusion validity}}
\textcolor{black}{
Conclusion limitations are based on potential wrong classified or missing papers as well as the replicability of the study. The number of papers and the variety made sure we can still generalize our conclusion even if some concepts or aspects are missing because of the chosen process. For instance there are no Natural Language Processing (NLP) specific papers, which can partly be explained by the scarcity of \enquote{safety-critical} angle in this domain. However, we are confident the snapshot and considerations provided would have led to the same conclusions and discussion would those missing domains be included. At least two reviewers discuss the main contributions of each paper to reduce the possibility of wrongly assigned papers to a category. Some papers might be considered tackling multiple categories, but in that case we either included it in different categories or we focused on the more important contribution. Finally, we provided a replication package\footnote{\url{https://github.com/FlowSs/How-to-Certify-Machine-Learning-BasedSafety-critical-Systems-A-Systematic-Literature-Review}} to allow for reproducibility of our results as well as to allow other researchers to build on our study.}

\section{Conclusion}\label{sec:conclusion}
This paper provides a comprehensive overview of certification challenges for ML based safety-critical systems. We conducted a systematic review of the literature pertaining to 
\emph{Robustness}, \emph{Uncertainty}, \emph{Explainability}, \emph{Verification}, \emph{Safe Reinforcement Learning} and \emph{Direct Certification}. We identified gaps in this literature and discussed about current limitations and future research opportunities. 
With this paper, we hope to provide the research community with a full view of certification challenges and stimulate more collaborations between academia and industry.

\begin{acknowledgements}
We would like to thank the following authors (in no particular order) who kindly provided us feedback about our review of their work: 
Mahum Naseer, Hoang-Dung Tran, Jie Ren, David Isele, Jesse Zhang, Michaela Klauck, Guy Katz,  Patrick Hart, Guy Amit, Yu Li, Anurag Arnab, Tiago Marques, Taylor T. Johnson, Molly O'Brien, Kimin Lee, Lukas Heinzmann, Björn Lütjens, Brendon G. Anderson, Marta Kwiatkowska, Patricia Pauli, Anna Monreale, Alexander Amini, Joerg Wagner, Adrian Schwaiger, Aman Sinha, Joel Dapello, Kim Peter Wabersich. Many thanks also goes to Freddy Lécué from Thalès, who provided us feedback on an early version of this manuscript. They all contributed to improving this SLR.
\end{acknowledgements}

%
\section*{Conflict of interest}
The authors declare that they have no conflict of interest. 
\section*{Appendices}
We provide the list of the papers used in each section. Note that some can only be found in the complementary material, in order to provide readers detailed information while keeping the main review concise.

\small
\begin{xltabular}{\textwidth}{|>{\centering\arraybackslash}m{2em}|c|>{\centering\arraybackslash}X|>{\centering\arraybackslash}X|}
\caption{Papers reference for each section. Note that "Others" and "Explainability/Interpretable Model" categories are not presented in main development, only in Complementary Material, in order to keep the paper concise.\label{tab:my_label}}\\
         \hline
         \textbf{Sec.} & \textbf{Sub-sections} & \textbf{Sub-categories} &\textbf{References}\\
         \toprule
         \hline
         \hline
         \multirow{2}{*}{\raisebox{-1.8\height}{\rotatebox[origin=c]{90}{\textbf{Robustness}}}} & \multirow{2}{*}{\emph{Robust Training}} & Empirically Justified &  \cite{summers2019improved}, \cite{li2018learning}, \cite{bakhti2019ddsa}, \cite{bar2019robustness}, \cite{bar2020robust}, \cite{suri2018hardening}, \cite{lecuyer2018connection}, \cite{rakin2018parametric},  \cite{duddu2019adversarial,liu2019affine}, \cite{laidlaw2019playing}, \cite{mirman2018differentiable}, \cite{gopfert2018mitigating} \\\cline{3-4}
         & & Formally Justified & \cite{pauli2020training},\cite{croce2019provable}, \cite{croce2019provable2},  \cite{hein2017formal}, \cite{sinha2017certifying}, \cite{kandel2020safe},  \cite{sinha2017certifying}, \cite{liu2020invariant}, \cite{reeb2018learning},  \cite{everett2020certified}, \cite{richards2018lyapunov}, \cite{revay2020convex},  \cite{dean2020robust}, \cite{varghese2020unsupervised} \\\cline{2-4}
         & \multirow{5}{*}{\parbox{2.2 cm}{\emph{Post-Training Robustness Analysis}}} & Pruning & \cite{Sehwag20}, \cite{Consentino19}, \cite{Yushuang19} \\\cline{3-4}
         & & CNN Considerations & \cite{Arnab2018}, \cite{zhang2019neuron}, \cite{liu2018analyzing} \\\cline{3-4}
         & & Other models considerations & \cite{Wang19-2}, \cite{Grefenstette18}, \cite{Mani19-2},\cite{Wei20}, \cite{Pan19} \\\cline{3-4}
         & & Non-Adversarial Perturbations & \cite{Hendrycks19}, \cite{Arcaini20}, \cite{Muller15}, \cite{Colangelo19}, \cite{Jeddi20}, \cite{rakin2018parametric}, \cite{lecuyer2018connection} \\\cline{3-4}
         & & Nature Inspired & \cite{dapello2020simulating}, \cite{rusak2020increasing}, \cite{Ye19} \\
         \midrule
         \hline
         \multirow{2}{*}{\raisebox{-1.15\height}{\textbf{\rotatebox{90}{Uncertainty}}}} & \multirow{3}{*}{\emph{Uncertainty}} & Aleatoric Uncertainty & \cite{gruber2018uncertainties}, \cite{le2018uncertainty}, \cite{taha2019unsupervised}, \cite{segu2019general} \\\cline{3-4}
         & & Epistemtic Uncertainty & \cite{gruber2018uncertainties}, \cite{sheikholeslami2020minimum}, \cite{lee2019ensemble},  \cite{toubeh2019risk}, \cite{peng2019bayesian}, \cite{postels2019sampling}, \cite{park2018sampling}, \cite{jain2020decision}, \cite{amini2019deep}\\\cline{3-4}
         & & Uncertainty Evaluation \& Calibration &  \cite{klas2019uncertainty}, \cite{henne2020benchmarking}, \cite{michelmore2018evaluating}, \cite{feng2019can}, \cite{levi2019evaluating},
         \cite{kuppers2020multivariate}, 
         \cite{turchetta2016safe}, \cite{fisac2018general}, \cite{Fan20},\cite{zhan2017safe}\\\cline{2-4}
         & \emph{OOD Detection} & & \cite{Hendrycks18}, \cite{Gal16}, \cite{Liang20}, \cite{Lee18}, \cite{Ren19}, \cite{Hein19},  \cite{Amit20}, \cite{Hendrycks20}, \cite{Wang18}, \cite{Lust20}, \cite{Gschossmann19}, \cite{meinke2019towards}, \cite{gu2019towards}, \cite{Henriksson19}, \cite{Sehwag19}\\
         \midrule
         \hline
         \multirow{3}{*}{\raisebox{-1\height}{\textbf{\rotatebox{90}{Explain.}}}} & & Model-Agnostic & \cite{ribeiro2016should}, \cite{guidotti2019factual}, \cite{pedreschi2018open}, \cite{wang2019deepvid}, \cite{guo2018lemna}, \cite{inouye2019diagnostic} \\\cline{3-4}
          & & Model-Specific & \cite{wagner2019interpretable}, \cite{nowak2019improve}, \cite{amarasinghe2019explaining}, \cite{meyes2020under}, \cite{meyes2020you}, \cite{kuwajima2019improving}\\\cline{3-4}
          & & Evaluation Metrics & \cite{wagner2019interpretable}, \cite{nowak2019improve}\\\cline{3-4}
          & & Interpretable Models & \cite{ignatiev2018sat}, \cite{daniels2018scenarionet}\\
          \midrule
          \pagebreak \\
          \hline
          \multirow{2}{*}{\raisebox{-2.75\height}{\textbf{\rotatebox{90}{Verification}}}} & \multirow{9}{*}{\emph{Formal Method}} & Verification through Reluplex &  \cite{katz2017reluplex}, \cite{gopinath2017deepsafe}, \cite{julian2019verifying}, \cite{keyfinding}, \cite{wang2019verification}, \cite{Julian20}, \cite{wang2018formal}, \cite{ren2019using}, \cite{julian2019guaranteeing}\\\cline{3-4}
          & & SMT-based Model checking: Direct Verification & \cite{huang2017safety}, \cite{naseer2020fannet}\\\cline{3-4}
          & & Bounded and statistical Model Checking & \cite{baluta2019quantitative}, \cite{gros2020deep}, \cite{sena2019incremental}\\\cline{3-4}
          & & Reachable sets & \cite{tran2020nnv}, \cite{katz2017reluplex}, \cite{tran2019parallelizable}, \cite{xiang2019reachable}\\\cline{3-4}
          & & Abstract Interpretation & \cite{gehr2018ai2}, \cite{li2019analyzing}\\\cline{3-4}
          & & Linear Programming & \cite{lin2019robustness}, \cite{rubies2019fast}, \cite{anderson2020tightened}, \cite{lyu2020fastened}, \cite{guidotti2019repairing}\\\cline{3-4}
          & & Mixed-integer linear program & \cite{remeli2019towards}, \cite{dutta2017output}, \cite{cheng2017maximum}, \cite{singh2018boosting}, \cite{bunel2020branch}\\\cline{3-4}
          & & Gaussian processes verification & \cite{smith2019adversarial}, \cite{cardelli2019robustness}\\\cline{3-4}
          & & Formal verification: other methods & \cite{tornblom2020formal}, \cite{wu2020game}, \cite{ruan2019global}, \cite{wang2018towards}, \cite{katz2017reluplex}, \cite{ghosh2018verifying}, \cite{salay2019safety}, \cite{fremont2020formal}, \cite{wang2018efficient}\\\cline{2-4}
          & \multirow{9}{*}{\emph{Testing Method}} & Testing criteria & \cite{Pei17}, \cite{Lei18}, \cite{Xiaofei19}, \cite{Ma19}, \cite{Sekhon19}, \cite{Sun19} \\\cline{3-4}
          & & Test generation coverage based & \cite{Xiaofei19}, \cite{Guo18}, \cite{Demir19}, \cite{Tian18}, \cite{BenBraiek19}, \cite{Sun19-2}, \cite{Zhang19}, \cite{Tian18}\\\cline{3-4}
          & & Test generation other methods & \cite{Yan20}, \cite{dreossi17}, \cite{Udeshi20}, \cite{Pei17-2}, \cite{Chen20}, \cite{Wicker18}, \cite{Yaghoubi19}, \cite{Gambi19}, \cite{Truncali18}, \cite{Wolschke17}\\\cline{3-4}
          & & Fault localization/injection testing & \cite{Eniser19}, \cite{Chen20-1} \\\cline{3-4}
          & & Differential Testing & \cite{Grefenstette18}, \cite{Xu19}, \cite{Machida19}, \cite{Yu19},  \cite{Cheng19}, \cite{Ramanagopal18}\\\cline{3-4}
          & & Quality testing & \cite{Mani19}, \cite{Tian20}, \cite{Ameyaw19}, \cite{Feng20}, \cite{Alagoz17}, \cite{Gladisch20}\\
          \midrule
          \hline
          \multirow{3}{*}{\raisebox{-1.2\height}{\textbf{\rotatebox{90}{Safe-RL}}}} &  & Post-Optimization &  \cite{hart2019lane}, \cite{baheri2019deep}, \cite{isele2018safe}, \cite{o2020dependable},  \\\cline{3-4}
          &  & Uncertainty estimation &  \cite{lutjens2019safe}, \cite{uesato2018rigorous}, \cite{zhang2020cautious}, \cite{tang2019worst}\\\cline{3-4}
          &  & Certification of RL-based controllers & \cite{nguyen2020towards}, \cite{wabersich2021probabilistic}, \cite{berkenkamp2017safe}, \cite{chakrabarty2019approximate}, \cite{gros2020safe}, \cite{wabersich2020bayesian}, \cite{wabersich2020performance}, \cite{fulton2019verifiably}, \cite{wen2020constrained}, \cite{hasanbeig2020towards}, \cite{le2019batch}, \cite{bacci2020probabilistic}, \cite{marvi2020safe}, \cite{cheng2019end}, \cite{yang2020safe}, \cite{deshmukh2019learning}  \\
          \midrule
          \hline
          \textbf{\raisebox{-0.6\height}{\rotatebox{90}{\parbox[t]{1.8cm}{Direct\\ Certification}}}} &  &  &  \cite{Fujino19}, \cite{Rudolph18}, \cite{Gauerhof18}, \cite{Salay18}, \cite{Heinzmann19}, \cite{Henriksson19-2}, \cite{Pedroza19}, \cite{Aravantinos19}, \cite{Biondi20}, \cite{Cofar20}, \cite{Kaprocki19}, \cite{Burton19}, \cite{Gauerhof20}, \cite{Rahimi19}\\
          \midrule
          \hline
          \textbf{\raisebox{0.\height}{\rotatebox{90}{\parbox[t]{1.2cm}{Others}}}} &  &  &  \cite{Cheng19-3}, \cite{Sohn19}, \cite{Ghosh18}, \cite{Cheng20-2}, \cite{Dey20}, \cite{Scheel20}, \cite{Steinhardt17}, \cite{Aslansefat20}, \cite{Lee19}, \cite{Kuutti19}, \cite{ayers2020parot}\\
          \bottomrule
\end{xltabular}
\normalsize
\bibliographystyle{spbasic}      
\bibliography{anle,florian,gabriel,amin,stevia}   

\clearpage

\begin{appendices}\label{appendix}

\section{Robust Dynamic Systems Modeling}
In the main text, robustness of ML systems refered to the resilience of the model when being fed inputs that differ from the training data \eg{} Distributional Shift and Adversarial Attacks.
However, in the context of dynamic systems modeling, the term robustness is used interchangeably with the term stability, which refers to the property that the modeled system should not diverge as time updates are applied. Recent ML-based modelings of dynamical systems with formal robustness guarantees include Lyapunov networks \cite{richards2018lyapunov}, a convex reparametrization of Recurrent NNs \cite{revay2020convex}, and linear dynamical systems with partial state information extracted from images \cite{dean2020robust}.
A notion very similar to stability of dynamical modeling is the 
\enquote{temporal consistency} of semantic segmentation of videos, \ie{} the constraint that an object should not appear/disappear in consecutive frames \cite{varghese2020unsupervised}. Combining the notions of stability and temporal consistency, we can state more generally that systems evolving over time are robust if they behave appropriately at any time $t$. This definition simultaneously includes behavior over infinitesimal time steps and behavior as time goes to infinity.

\section{Uncertainty estimation and OOD detection}

\subsection{Uncertainty to guide Exploration}
It was argued in the main text that uncertainty estimates are of high quality if they act as accurate proxies of model confidence \ie{} they are higher on instances where the model is likely to fail and lower on instance were the predictions are correct. For this reason, good uncertainty estimates are essential tools increase the safety of ML models at prediction-time, seeing that large values of uncertainty could force the system to ignore ML decisions and fall back to safer programs.

However, some papers in the literature were found to use uncertainty in a different manner. For example, we found studies where uncertainty estimates were applied to Markov Decision Processes \cite{turchetta2016safe}, and dynamical systems \cite{fisac2018general,Fan20}, allowing for safer autonomous control. Note that the definition of \enquote{safe} is application dependent and must be derived from domain knowledge. For example, in autonomous driving, criteria for safe control could be to have a low path curvature and being far away from lateral obstacles \cite{zhan2017safe}. On these tasks, uncertainty measurements are not used as proxies for prediction confidence, but more as tools that allow autonomous agents to dynamically update the amount of information they know about their environment.

\subsection{Calibration in Regression Tasks}

Although the concept of calibration is intuitive for classification, it is far less straight-forward to define in regression. Indeed, in classification, if 100 predictions with calibrated certainty 0.9 are made, we expect around 10 of those predictions to be wrong. In regression tasks, the difficulty of defining calibration stems
from the fact that uncertainty estimates yield a probability density function of the target given the input $p(y|x)$. What is the frequentist way to interpret this distribution across multiple inputs $x$? 
A first definition of calibration advanced in the literature is to measure how well the quantiles of the conditional distribution $p(y|x)$  match their empirical counterparts over the whole dataset \cite{feng2019can}. This formulation has however recently been criticized and redefined in terms of the ability of uncertainty estimators to predict the
Mean Square Error of the model at any datapoint \cite{levi2019evaluating}. Therefore, as we see it, finding the right definition of calibration is still an open research question.

\section{Explainability via Interpretable Models}
As stated in the main paper, the two main schools of thought in eXplainable Artificial Intelligence (XAI) are post-hoc explanations of black boxes and the training of interpretable yet accurate models. This duality is induced by the interpretability-accuracy trade-off \ie{} the
observation that black boxes tend to out-perform  interpretable models in practice. The first paradigm (post-hoc explanations) tackles the trade-off by leaving the current state-of-the-art models intact and develop add-on methods to interpret their decisions. Since the performance is kept intact, these methods have quickly gained in popularity.

The second paradigm addresses the trade-off by developing new models that have a high performance but retain interpretability. For example, \cite{ignatiev2018sat} proposed new SAT-based solutions 
to learn Decision Sets, a task that is known to be NP-hard.
Moreover, a novel CNN architecture called ScenarioNet introduces the notion of
scenarios: sparse representations of data encoding sets of co-occurring objects in images \cite{daniels2018scenarionet}.

\section{Formal verification: Other methods} 
The VoTE tool \cite{tornblom2020formal} has 2 components: VoTE core and VoTE Property Checker. VoTE core will compute all equivalence classes in the prediction function related to the tree ensemble. VoTE Property Checker takes as input the equivalence classes and the property to be verified, and checks if the input-output mappings from each equivalence class are valid. The experiment conducted by the authors consists of evaluating the robustness, scalability and node selection strategy of VoTE. The latter shows successful performance. 

The authors in \cite{wu2020game} describe their approach as follow: Given an input, Player I selects by turning a feature to perturb and Player II chooses a perturbation. While Player II aims at minimizing the distance to adversarial examples, the game is cooperative for the maximum safe radius approximation and competitive for the feature robustness. Moreover, they use Lipschitz continuity to bound the maximum
variation on outputs depending on inputs in order to provide guarantee bounds on all possible inputs. To address the intractable nature of the computation space, the authors use an approximation relying on Monte-Carlo Tree search for the upper bound estimation and the path finding $A^{*}$ algorithm with pruning for the lower bound. Unfortunately, the lower/upper bound approximation gap widens as the number of features increases. 

\cite{ghosh2018verifying}, used Bayesian Optimization (BO) to predict the environment scenarios and the counterexamples that are most likely to cause failures in the designed controllers. They have tested their approach on some simple functions and then on some benchmark environments of OpenAI Gym. Regarding the application domain, it is unclear what kind of specification could be covered by the approach.

In \cite{ruan2019global}, the authors provide global robustness approximation sequences for lower/upper bounds using Hamming Distance, since the classical safe radius with $L_0$ norm is NP-hard. Their approach offers provable guarantees and an effective and scalable way of computing them. However, the experiment section of the paper would appear to lack coherence, which makes the approach hard to evaluate.

Relying on a \emph{classification hierarchy} , \cite{salay2019safety} identify four possible outcomes in a classification task, based on whether the input is correctly classified to the \enquote{best} label: correct classification (\eg{} correctly predict a \enquote{car}), under-classification (\eg{} correctly predict a \enquote{vehicle}), misclassification (\eg{} incorrectly predict a \enquote{truck}), and under-misclassification (\eg{} incorrectly predict \enquote{other}). The authors introduced equations to calculate the risk and control policy, as well as the action severity based on these two metrics. The result and the analysis process could be very useful for autonomous car producers to improve their ML systems. During experiments, the authors also observed that even a simple classification architecture can lead to a large number of classification cases because of the under-classification. Thus a complexity reduction strategy needs to be introduced in the future. However, This work can be improved in some aspects:
\begin{itemize}
    \item The choice of non-leaf classes can lead to different results in terms of safety analysis. Especially, the complexity of the classification cases and the progress (time to complete a task) can vary much based on if we merge two leaf classes or not.
    \item The proposed safety analysis framework is very interesting but still needs to be verified in a more realistic scenario to validate how it can be generalized and whether/how it can handle a real-time data stream.
\end{itemize}

A hybrid approach to verifying a DNN has been studied in \cite{wang2018towards}. The authors propose to estimate the sensitivity of NNs to measure their robustness against adversarial attacks. The sensitivity metric is computed as the volume of a box over-approximation of the output reachable set of the NN given an input set. They applied two methods to compute the approximations, a dual objective function representing the sensitivity computed using the dual formulation and the sensitivity computed via Reluplex \cite{katz2017reluplex}. However, it is not clear what kind of adversarial attacks the robustness criteria might prevent from.
To assess the safety of aircraft systems, the authors of \cite{fremont2020formal} study the verification of Boeing's NN-based autonomous aircraft taxiing system. They defined a safety requirement as: in 10 seconds, the plane must reach within 1.5 m of the centerline and then stay there for the remainder of the operations. Failing to satisfy this requirement will be considered as a counterexample. In their experiment with the X-Plane flight simulator, they found only 55.2 \% of the runs satisfying the requirement while 9.1 \% of the runs completely left away from the centerline. 
They analyzed the failed cases and observed that cloud and shadow can misguide the plane's taxiing. In addition, the NN of the TaxiNet system poorly handled intersections. Using this diagnosis information, the authors retrained the NN and obtained 86\% of successful runs and only observed 0.5\% of runs leaving the runway. This approach is very promising to be used by other avionic companies to verify their autonomous AI or ML-based taxiing systems. It can also be used to verify other autonomous systems, such as landing, collision avoidance, or deicing systems. As the author mentioned, there are still 14\% of the runs that failed to satisfy the safety requirement. To certify this taxiing system for real aircraft, we need to further improve the performance of the NN. Another line of research aims to check the satisfiability of safety properties of NNs by estimating tight output bounds given an input range \cite{wang2018efficient}. This approach relies on symbolic linear relaxation to provide those tighter bounds on the network output. Then a directed constraint refinement process follows to minimize errors due to the relaxation process. The experiment's results show that the approach outperforms state-of-the-art analysis systems and can help improve the explainability of NNs. 
 
\section{Certification of RL-based controllers using barrier function}

The value of a barrier function on a point is defined to be increased to infinity, as the point approaches boundaries of the feasible region of an optimization problem \cite{nesterov2018lectures}. Control Barrier Functions (CBF) is consequently defined to be positive within the predefined safe set and it reaches infinity at the boundary of that set. Therefore, CBFs plays a equivalent role to Lyapunov functions (for stability guarantee) in the study of liveness properties of dynamical systems; if one finds a CBF for a given system, it becomes possible to define the set of admissible initial states and a feedback strategy that ensures safety for that system by indicating unsafe states. For example, \cite{marvi2020safe} proposed a safe learning-based controller using CBF and actor-critic architecture. The authors have augmented the original performance function with a CBF candidate, to penalize actions that violate safety constraints. CBFs, with the initial condition belonging to a predefined set, guarantee that the states of the system will stay within that set by imposing proper condition on the trajectories of a nonlinear system. The control policy has been evaluated on a lane changing scenario as a challenging task in vehicle autonomy showing its effectiveness. However, a known invariant set of safe states is required for employing CBFs, which is not the case for many real-world problems. Similarly, RL-CBF framework \cite{cheng2019end} is a combination of a model-free RL-based controller, model-based controllers utilizing CBFs and real-time learning of the unknown system dynamics. Since CBF-based controllers can guarantee safety and also conduct RL learning by restricting the set of explorable policies, the goal is to ensure the safety during learning while benefiting from high performance of RL-based controllers. The safe RL-CBF has been successfully evaluated on an autonomous car-following scenario with wireless vehicle-to-vehicle communication performing more efficiently than other state-of-the-art algorithms while staying safe. A new actor-critic-barrier structure for multiplayer safety-critical games (systems) has been introduced in \cite{yang2020safe}. Assuming that a system is represented by a non-zero-sum (NZS) game with full-state constraints, a barrier function was used to transform the game to an unconstrained NZS. An actor-critic model with deep architecture is then utilized to learn the Nash equilibrium in an online manner. Authors showed that the proposed structure, \ie{} actor-critic-barrier, does not violate the constraints during learning, given that the initial state is in a predefined bound. A nonlinear system with two-players has been successfully used to evaluate the proposed structure along with analyzing its boundedness and stability. Authors have proposed an approach to train safe DNN controllers for cyber-physical systems so that they can satisfy given safety properties in \cite{deshmukh2019learning}. The key idea of the approach is to embed safety properties into the RL. The properties are checked through an SMT-based verification process to impose penalties on undesired behaviors. To do so, if the verification results in failure, the RL cost function is modified using the generated counter-example to search towards safe policies. This process is repeated until achieving a proof of safety. As a proof of concept, three nonlinear dynamical system examples have been successfully tested.

\section{Others} 
This section regroups papers that could not fit in other categories, and as such couldn't be included in the main development. However, they present interesting approach that are within the scope of our study and thus we give a brief overview of what they elaborate on.
 
\begin{itemize}
    \item NN-dependability-kit \cite{Cheng19-3} is a data driven toolbox that aims to provide directives to ensure uncertainty reduction in all the life cycle steps, notably robustness analysis with perturbation metrics and t-way coverage. This technique was used by \cite{Gauerhof20} mentioned in the \enquote{Direct Certification} with its requirements-driven safety assurance approach for ML. Building upon NN-dependability-kit, "specific requirements that are explicitly and traceably linked to system-level safety analysis" were tackled. While NN-dependability-kit did not focus on this aspect, it is an important one to comply with safety issues.
    \item Instead of trying to improve the model resilience through loss regularization or other mechanisms, some techniques focused on post training reparation of the model; the common idea is to search through the DNN for neurons that could lead to unexpected behavior and patch them. \cite{Sohn19} used Particle Swarm Optimization (PSO) to modify weights of layers to correct faulty behavior. \cite{Ghosh18} instead used Markov Decision Process (MDP) to repair a model through safety properties that can directly be exploited to modify weights during training. They also show that with this technique they can \enquote{repair} data, by screening noisy data that would be outside of a safety envelope and remove those outliers.
    \item Generally used in object recognition, \emph{bounding boxes} are a technique to increase safety. \cite{Cheng20-2} applied it on 3d points cloud NN PIXOR whose task is to predict the final position of an object. They split the decision part of the algorithm into non-critical and critical area detection, critical area symbolizing where the model should process carefully. For the latter part, they use a non-max-inclusion algorithm, which enlarges the prediction area of the same object by taking into account boxes with lower probability, in order to remain conservative as ground truth is not known in operation. 
    \item DNN have been quite developed in our paper, as they represent a huge portion of the ML model used. However, there are some efforts to bring certification processes also for other models’ types. \cite{Dey20} proposed a method for augmenting decision trees through expert knowledge with refinement to reduce the number of variables. They further leverage the information gain metric to \enquote{prune} the decision tree to retain an optimal decision tree. However, they assume attributes are independent, so a combination can exist in the tree without really existing in actual possibilities which can induce biases.
    \item Another interesting addition to safety-critical certification could come from transfer learning. Transfer learning is a learning technique which aims at reusing a part or an entire pre-trained model on task A, in order to adapt it to task B. This allows for decreased training time as well as potentially increased accuracy. This approach could be useful in safety-critical applications when environment modifications happen. Indeed, every time a sensor configuration is modified, one would like to avoid retraining the network from scratch to account for this modification. Transferring already established knowledge could help solve this issue. \cite{Scheel20} proposed an approach based on NNs to calculate a transformation matrix mapping each input from an existing domain to the new one, helping at the same time to understand how data are mapped between domains. They mainly tested it on a lane-changing driving problem. However, further studies would be required to take into consideration harmful consequences  that transferred knowledge could bring.
    \item \cite{Steinhardt17} tackles data poisoning attack, \ie{} a game between a defender and an attacker. The goal of the defender is to learn a good model, while the attacker's goal is for the defender to fail. $N$ data points are selected, the attacker choose $M$ poisoned data points such that $M :=\lfloor \epsilon N \rfloor \leq N$ (depending the budget $\epsilon$ of attacker), the defender learns on the $M + N$ data points and aim to minimize the loss. The poisoned data $D_p$ consists only of new data, that is the attacker cannot modify existing sane data $D_c$. The authors use a modified version of data sanitized defense, consisting of finding the poisoned data and removing it. They upper bound the worst test-case loss under attack with $\displaystyle \max_{D_p \subseteq \mathcal{F}} \min_\theta \mathcal{L}(D_p \cup D_c)$, where $\mathcal{F}$ is called the feasible set, and can be for instance a neighborhood sphere of a given radius. The main assumption of the method is that all clean data points are feasible \eg{} $D_c \subset \mathcal{F}$.
    \item Statistical metrics can be leveraged to improve safety of ML-based systems. \cite{Aslansefat20} does so by first training an offline model with a trusted dataset, gathering information such as cumulative distribution of classes data, accuracy of model, etc. The system can then be put online and can compare new influx of data through the same statistics using confidence level. If a high divergence is measured, the system can report the error.
    \item \cite{Lee19} proposes an approach for a safe visual-based navigation system by exploiting perceptual control policies. To that end, a model predictive network, which itself relies on a model predictive controller, is used to provide information about the vehicle and select regions of interest on the visual input. This information is considered as expert trajectories and is used through imitation learning to learn a perceptual controller of the navigation system. The perceptual controller, which is the main contribution of this approach outperforms baselines, by quickly detecting unsafe conditions that the navigation system might encounter through uncertainty quantification.
    \item Finally, \cite{Kuutti19} describes the use of safety cages to control actions in an autonomous vehicle. Safety cages are generally used on black box systems where we do not have full understanding of how the system works. They limit unsafe actions the system can take. Their approach is based on Imitation Learning, which learns from a simulation based autonomous vehicle. From the information collected via imitation learning, the safety process leads the autonomous vehicle to avoid collisions.
    \item We suspect that, as modules implementing techniques from all the sub-fields become readily available in popular Deep Learning frameworks such as TensorFlow and Pytorch, we will see an increase in cross sub-fields studies. As an example, robustness modules for Tensorflow/Keras are now available \cite{ayers2020parot}.
\end{itemize}

\section{ML vs standard SE: a simplified example}

This section addresses the main differences between traditional Software Engineering and the Machine Learning methodology in a manner that naturally highlights the ML challenges identified in the paper (Robustness, Uncertainty, and Explanability). The comparison is done through the lens of a simple task: estimating a sine function.

In this experiment, practitioner A will employ standard SE programming while developer B will apply the ML methodology. We shall see that, even in this simple setting, the ML challenges restrict programmer B satisfying basic requirements, henceforth making certification impossible. Note that we use this mock example only for comprehension purpose and we shall exaggerate some aspects to better distinguish the SE and ML paradigms, therefore it is not exactly representative of what would be done in practice.

\subsection{Building the System}
To accomplish the task, coder A must have a mathematical formulation of the sine in terms of standard computer operations (additions, multiplications, logic, etc.), and then implement it.
Assuming that all derivatives of a sine function exist and are continuous, doing a Taylor expansion of the function around $x=0$ yields the formula
\begin{equation}
    \sin(x)=x-\frac{x^3}{3!} + \frac{x^5}{5!}-
    \frac{x^7}{7!} +\frac{x^9}{9!}+\ldots
\end{equation}
The programmer A may therefore start with the very simple algorithm.

\begin{algorithm}
\caption{Simple approximation of $\sin(x)$}
\begin{algorithmic}
\Procedure{naive\_sine}{x}
\State $y \gets x$;   \hspace{0.25in}\green{//output}
\State $t \gets x$;   \hspace{0.25in}\green{//next term}
\For{$k=1, 2, 3 ,4$}
    \State $t \gets \big[\frac{-x^2}{2k(2k+1)}\big]\times t$;
    \State $y \gets y+t$;
\EndFor
\State \Return y
\EndProcedure
\end{algorithmic}
\label{alg:sine}
\end{algorithm}
On the other hand, developer B who follows the ML methodology requires a dataset to
represent prior knowledge about the task at hand
$S=\{(x^{(1)}, y^{(1)}), (x^{(2)}, y^{(2)}),\ldots, (x^{(N)}, y^{(N)})\}$ whose inputs are (for example) sampled uniformly
in the interval $[-4\pi, 4\pi]$ and labeled with the sine function. Using the additional information that the sine function is continuous, it is decided to model it using a simple Multi-Layered Perceptron (MLP)
\begin{equation}
    \text{MLP}(x)=\sum_{\ell=1}^nw^{[2]}_{\ell}\bigg(\sum_{k=1}^d w^{[1]}_{\ell, k}x_k + b^{[1]}_\ell\bigg) + b^{[2]}.
\end{equation}
whose weights and biases are \enquote{learned} by minimizing the mean squared error on the training data.

\subsection{Interpretability}

To ensure that the programs work according to task specifications, both practitioners start by evaluating their function on various inputs.

During testing, developer A observes that, for $x=6$,  the output is $7.029$, which violates the requirement that sine is between $-1$ and $1$. The practitioners then starts looking for the problem and, since Algorithm \ref{alg:sine} is traceable and interpretable, the developer can backtrack through the operations and observe that the output was high because the function computed a large positive contribution $x^9/(9!)$ compared to other terms in the Taylor expansion. Based on this, the programmer develops the intuition that the model will not work well on large magnitude inputs ($|x|\gg 1$) seeing that this ninth-power term will explode.

On the other hand, practitionner B observes that the MLP sometimes yields values with magnitude larger than one, but unlike Algorithm \ref{alg:sine}, the MLP is opaque and it not immediately clear to the developer what exactly causes these aberrant outputs simply by looking at the weights and biases. We have therefore identified a first ML challenge.

\begin{center}
    \textbf{Challenge $\#1$ : Interpretability/Explainability}. The processes that yield decisions of ML models are often opaque so when something goes wrong in the system is it not immediately clear what parameters induced the fault, nor what corrections should be applied to the model.
\end{center}

\subsection{Uncertainty}
The two methods implemented by programmers A and B are still approximations of the true sine function, and therefore the coders are both required to provide assessments of the quality of their approximation at any given $x$.

Looking deeper at the mathematics behind
Taylor expansions, developer A discovers that the error of the program at any given $x$ is theoretically bounded by $|x|^{11}/(11!)$\ (if we ignore floating point arithmetic errors), and decides to report this measure along with the output of Algorithm \ref{alg:sine}.

\begin{figure}[t]
    \centering
    \begin{minipage}{.478\textwidth}
        \centering
        \includegraphics[width=\linewidth]
        {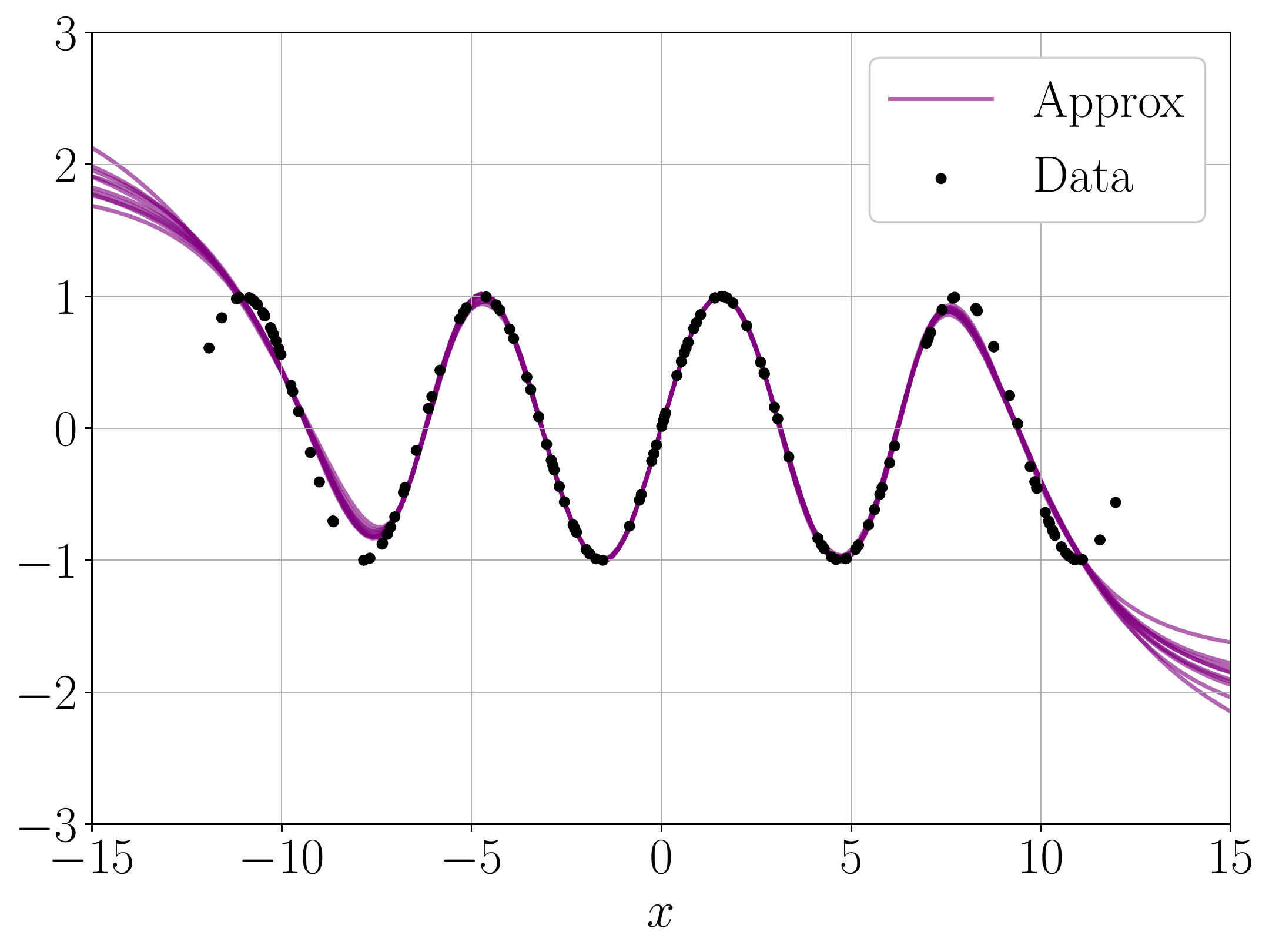}
        \caption{Sine approximation with ten independently trained MLPs.}
    \label{fig:sine_ML}
    \end{minipage}\hfill
    \begin{minipage}{0.504\textwidth}
        \centering
        \includegraphics[width=\linewidth]
        {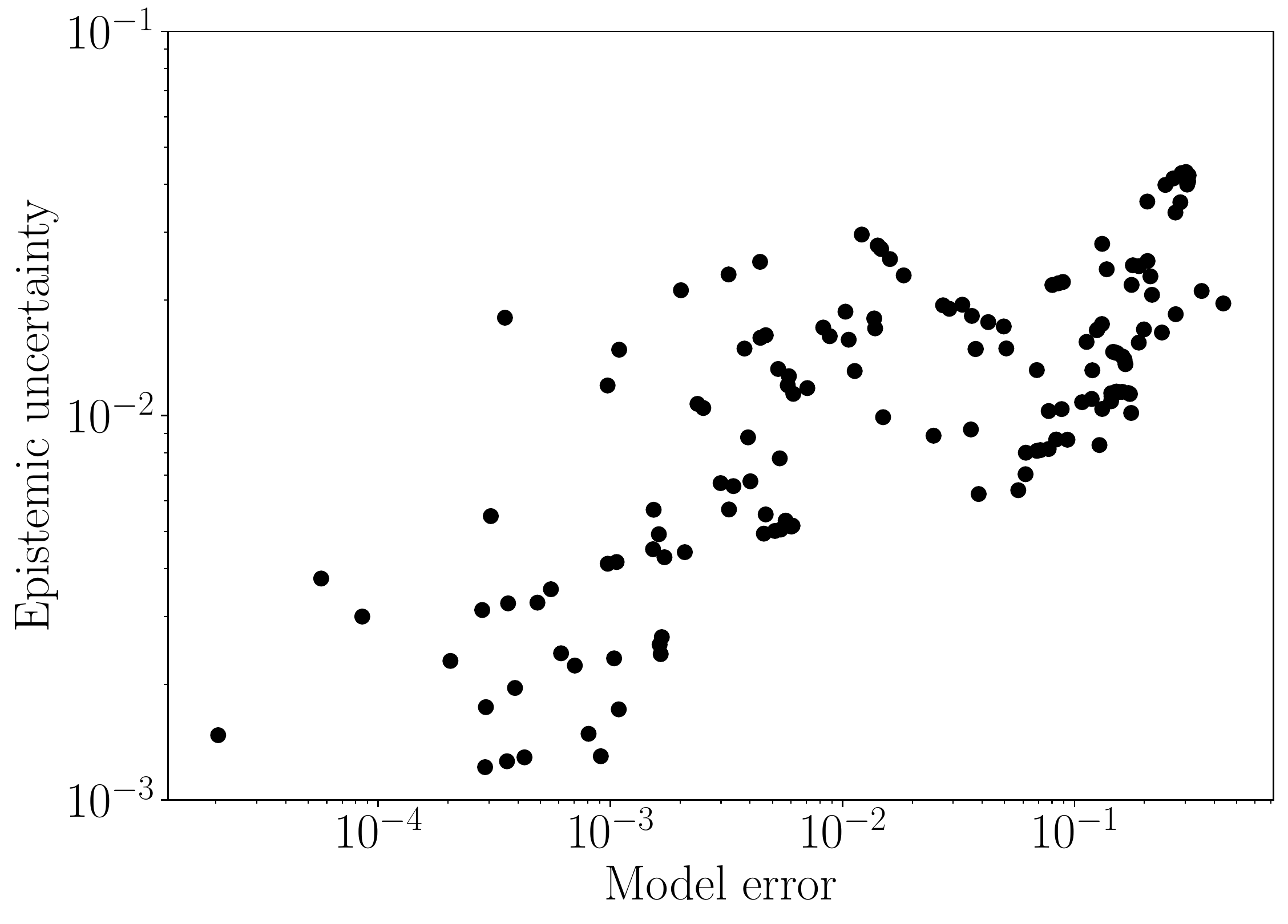}
        \caption{Using uncertainty as a proxy for error.}
        \label{fig:sine_proxy}
    \end{minipage}
\end{figure}

Most theoretical bounds in ML are probabilistic in essence and do not provide guarantees of the model error at any specific $x$, but only on the average error. For this reason, practitioners B opts to employ uncertainty estimates as proxies for model error at any specified input. Considering that, for the sine computation task, the relationship between the input and the target is deterministic, only epistemic uncertainty needs to be considered. To obtain estimates of the epistemic uncertainty, developer B uses the Ensemble method: trains ten MLPs independently on the training set, and uses the input-wise disagreement (variance) between the ten models predictions as an uncertainty measure, see Figure \ref{fig:sine_ML}. The uncertainty is finally computed on held-out data points and compared to the error of the average MLP on these same points (Figure \ref{fig:sine_proxy}). Based on this plot, developer B states that the uncertainty is a proxy of model error (seeing that they are positively correlated) and that predictions with a \enquote{large enough} uncertainties should be ignored. Still, there are no theoretically grounded ways to choose the appropriate threshold of uncertainty over which predictions are considered untrustworthy, nor any guarantee that all predictions made with a low uncertainty will be accurate. We have just identified a second ML challenge.

\begin{center}
    \textbf{Challenge $\#2$ : Uncertainty}. As processes in ML systems are opaque and also subject to stochasticity, it is primordial to develop reliable error proxies in the form of uncertainty estimates, which can be used to reject model decisions on inputs where the system is likely to fail.
\end{center}

\subsection{Robustness}

To certify a program, one must provide guarantees that
it will work properly in deployment. In the case of a sine approximation, the output of the program is expected be accurate given any input $x\in[-\infty, \infty]$.
Sadly, through testing, both developers see that the accuracy of their method highly depends on the localisation of the inputs and therefore they cannot pass certification as is.

Nonetheless, because Algorithm \ref{alg:sine} is interpretable and because the error upper bound scales as $|x|^{11}$, practitioner A knows that the program is likely to fail on inputs with large magnitudes. If we assume that the programmer has access to the background knowledge that the sine function is $2\pi-$periodic, the code can be made more robust by translating large inputs so they land in the $[-\pi, \pi]$ interval where the naive model is known to work quite well. This leads to the new Algorithm \ref{alg:robust_sine}, which is more likely to pass certification.

\begin{algorithm}
\caption{Robust approximation of $\sin(x)$}
\begin{algorithmic}
\Procedure{robust\_sine}{x}
\State \green{// Translate large inputs}
\If{$x>\pi$ or $x<-\pi$}
\State $x\gets x - 2\pi\times \texttt{round}(\frac{x}{2\pi})$
\EndIf
\State \green{// Sine approximation}
\State \Return \Call{naive\_sine}{$x$};
\EndProcedure
\end{algorithmic}
\label{alg:robust_sine}
\end{algorithm}

We are going to assume that developer $B$ does not have access to the knowledge that the target is periodic, and that the only information known about the sine is provided by the dataset. This is reasonable considering that a major appeal of ML is that it makes loose assumptions about the task, making is a very versatile tool.
Since the inputs from the training data where uniformly sampled from the interval $[-4\pi, 4\pi]$, coder B has no mean to guarantee that the models is trustworthy when being fed inputs that land outside this interval. This distributional shift between training and deployment domains is the third identified challenge.
\begin{center}
    \textbf{Challenge $\#3$ : Robustness}. A robust ML system should operate safely when put into deployment, even when a distributional shift occurs with the training data.
\end{center}
In this simple example, a trivial way for programmer B to increase robustness is to ignore predictions on inputs that fall outside the $[-4\pi, 4\pi]$ interval on which the model was trained. However, in most settings where ML is used, the data is high dimensional and determining the \enquote{safe zones} is not as simple as rejecting all inputs that fall outside some interval. Another property of high dimensional spaces is that data points sparsely populate them \ie{} there is a lot of space between data points. Highly non-linear models such as Neural Networks can have unpredictable behaviors in the large empty spaces between data points, which may explain the existence of the so-called Adversarial Examples that currently plague modern Neural Networks. Plus, in the mock example we make the
hypothesis that the data lives in on smooth manifold which may not always be the case in for many practical datasets.

\subsection{Beyond the sine}
As shown in the simple task of computing a sine function, the ML method introduces new challenges: lack of interpretability, requirement for reliable error proxies via uncertainty measures, and robustness issues induced by distributional shift between training and deployment domains.
Simply put, even on the simple task of computing a sine function, the combination of these challenges prevents coder B from certifying the program. 

This use-case can make ML seem less appealing than standard SE methods, but this is not the desired message. Indeed the main point of this example is to illustrate how ML challenges appear on basic tasks. Moreover, the reason that SE works so well is that this example is low-dimensional (the input has one dimension) and that strong assumptions were made on the target function, such that it is $2\pi-$periodic and all of its derivatives exist and are continuous. In such settings, standard methods are preferable because they come with strong theoretical guarantees and are more traceable/interpretable. 

Nonetheless, ML shines on tasks where the data is high-dimensional, and where very few assumptions can be made about the target (image classification for instance), which is also why testing is an major ML challenge.
\begin{center}
    \textbf{Challenge $\#4$ : Verification/Testing}. ML shines in settings where 1) the data is high-dimensional and 2) few assumptions about the ground-truth are available (no oracle).
    However, the high dimensionality of the input space increases of burden of generating test cases, while the lack of oracle
    induces an ambiguity as to what properties should be verified to assess model correctness.
\end{center}

\end{appendices}

\end{document}